%% file: root.tex
\documentclass{ieeetj}

\IEEEoverridecommandlockouts                              

\makeatletter
\let\NAT@parse\undefined
\makeatother
\usepackage[numbers,sort&compress]{natbib}

\usepackage[pdftex]{graphicx}
\usepackage{amsmath}
\usepackage{amssymb}  
\usepackage{subcaption}
\usepackage{multirow}
\usepackage{array,booktabs}
\usepackage{diagbox}
\usepackage{balance}
\usepackage{xspace}
\usepackage{algorithm}
\usepackage{algpseudocode}
\usepackage{adjustbox}
\usepackage{romannum}

\usepackage[final]{hyperref}
\hypersetup{
 colorlinks=true,
 linkcolor=blue,
 filecolor=magenta,
 urlcolor=blue,
 citecolor=black
}
\usepackage{cleveref}
\crefname{figure}{Fig.}{Figs.}
\Crefname{figure}{Fig.}{Figs.}
\usepackage{rpm_SIunits}
\usepackage{rpm_acronyms}
\usepackage{rpm_math}
\usepackage{rpm_misc}
\usepackage{bbm}
\usepackage{placeins}
\usepackage{textcomp}

\usepackage{soul,color}
\usepackage{lipsum}
\usepackage{dsfont}

\usepackage{xcolor}
\usepackage{colortbl}

\newcommand{\bl}[1]{{\textcolor{blue}{#1}}}

\usepackage{amssymb}
\usepackage{pifont}
\newcommand{\xmark}{\ding{55}}%
\usepackage{tikz} 

\hypersetup{hidelinks=true}
\def\BibTeX{{\rm B\kern-.05em{\sc i\kern-.025em b}\kern-.08em
    T\kern-.1667em\lower.7ex\hbox{E}\kern-.125emX}}
\AtBeginDocument{\definecolor{tmlcncolor}{cmyk}{0.93,0.59,0.15,0.02}\definecolor{NavyBlue}{RGB}{0,86,125}}

\def\OJlogo{\vspace{-4pt}$<$Society logo(s) and publication title will appear here.$>$}
\def\seclogo{\vspace{10pt}$<$Society logo(s) and publication title will appear here.$>$}

\DeclareCaptionFont{ieeetjcap}{%
\fontfamily{\sfdefault}\fontsize{7}{9}\selectfont\bfseries
}
\DeclareCaptionFont{ieeetjblue}{\color{tmlcncolor}}

\def\authorrefmark#1{\ensuremath{^{\textbf{#1}}}}

\NewDocumentCommand{\LCindent}{O{1}m}{%
  \Statex 
  \hspace*{\dimexpr #1\algorithmicindent\relax}%
  {\color{blue}\texttt{//}~#2}\strut%
}

\algrenewcommand\algorithmicrequire{\textbf{Input:}}
\algrenewcommand\algorithmicensure{\textbf{Output:}}

\begin{document}
\receiveddate{XX Month, XXXX}
\reviseddate{XX Month, XXXX}
\accepteddate{XX Month, XXXX}
\publisheddate{XX Month, XXXX}
\currentdate{XX Month, XXXX}
\doiinfo{XXXX.2022.1234567}

\markboth{}{Author {et al.}}

\title{TreeLoc++: Robust 6-DoF LiDAR Localization in Forests with a Compact Digital Forest Inventory}

\author{Minwoo Jung\authorrefmark{1}, Dongjae Lee\authorrefmark{1}, Nived Chebrolu\authorrefmark{2},\\ Haedam Oh\authorrefmark{3}, Maurice Fallon\authorrefmark{3} and Ayoung Kim\authorrefmark{1}}
\affil{Department of Mechanical Engineering, Seoul National University, Seoul, South Korea}
\affil{Department of Computer Science and Engineering, Indian Institute of Technology Bombay, India}
\affil{Oxford Robotics Institute, Department of Engineering Science, University of Oxford, Oxford, UK}
\corresp{Corresponding author: Ayoung Kim (email: ayoungk@snu.ac.kr).}
\authornote{This work was supported by the National Research Foundation of Korea (NRF) grant funded by the Korea government (MSIT) (No. RS-2024-00461409); the Horizon Europe project DigiForest (101070405); and EPSRC Project Mobile Robotic Inspector (EP/Z531212/1). For the purpose of open access, the authors have applied a Creative Commons Attribution (CC BY) license to any Author Accepted Manuscript version arising.}

\input{0_abstract}


\maketitle

\input{1_introduction}
\input{2_relatedwork}

\input{3_method}

\input{4_experiment}

\input{5_limitation}

\input{6_conclusion}


\balance
\small
\bibliographystyle{IEEEtranN} 
\bibliography{string-short,references}

\vfill\pagebreak

\newpage
\appendix
\nobalance

\input{appendix}
\end{document}

%% file: 0_abstract.tex
\begin{abstract}
Reliable localization is essential for sustainable forest management, as it allows robots or sensor systems to revisit and monitor the status of individual trees over long periods. In modern forestry, this management is structured around Digital Forest Inventories (DFIs), which encode stems using compact geometric attributes rather than raw data. Despite their central role, DFIs have been overlooked in localization research, and most methods still rely on dense gigabyte-sized point clouds that are costly to store and maintain.
To improve upon this, we propose TreeLoc++, a global localization framework that operates directly on DFIs as a discriminative representation, eliminating the need to use the raw point clouds. TreeLoc++ substantially strengthens the existing TreeLoc framework by reducing false matches in structurally ambiguous forests and improving the reliability of full 6-DoF pose estimation. Building upon TreeLoc, it augments coarse retrieval with a pairwise distance histogram that encodes local tree-layout context, subsequently refining candidates via {tree diameter at breast height (DBH)}-based filtering and yaw-consistent inlier selection to reduce mismatches. Furthermore, a constrained optimization leveraging tree geometry jointly estimates roll, pitch, and height, enhancing pose stability and enabling accurate localization without reliance on dense 3D point cloud data.
Evaluations on 27 sequences recorded in forests across three datasets and four countries show that TreeLoc++ achieves precise localization with centimeter-level accuracy. We further demonstrate robustness to long-term change by localizing data recorded in 2025 against inventories built from 2023 data, spanning a two-year interval. The system represents 15 sessions spanning \unit{7.98}{km} of trajectories using only \unit{250}{KB} of map data and outperforms both hand-crafted and learning-based baselines that rely on dense point cloud maps. This demonstrates the scalability of TreeLoc++ for long-term deployment.
TreeLoc++ is open-sourced at \bl{https://github.com/minwoo0611/TreeLoc-plusplus}.
\end{abstract}

%% file: 1_introduction.tex
\section{Introduction}
\begin{figure*}[!t]
    \centering
    \includegraphics[width=.97\textwidth]{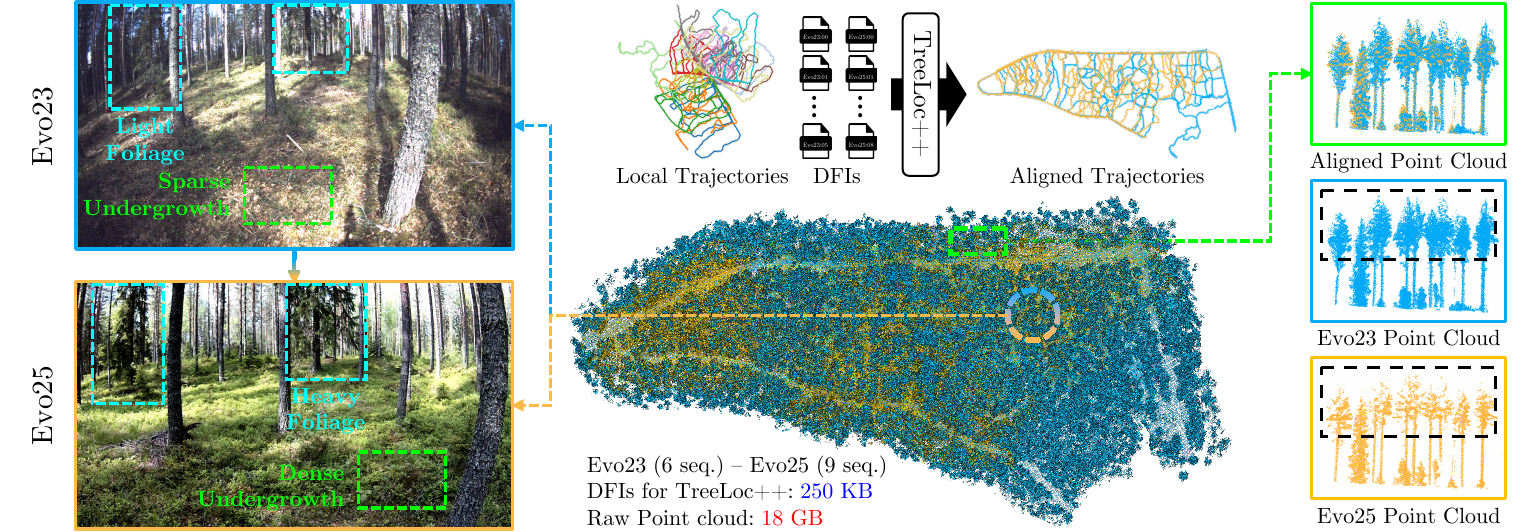}
\caption{(Left) Images captured from the same location in the \texttt{Evo} forest two years apart, showing vegetation growth {(light green)} and denser foliage (cyan), which alter scene appearance and increase localization difficulty. (Right) TreeLoc++ achieves accurate multi-session alignment using only lightweight DFIs by leveraging tree attributes, avoiding raw point clouds that can vary across sessions due to different LiDAR sensors (black).}
    \vspace{-3mm}
    \label{fig:main}
\end{figure*}

\label{sec:intro}
\IEEEPARstart{R}
{}eliable and accurate localization is essential in forestry and environmental monitoring, where systems operate in complex under-canopy environments~\cite{yin2024survey, miettinen2007simultaneous, mattamala2025wild}. Localization enables key long-term tasks such as repeated site access, consistent map maintenance under dynamic canopy conditions, and digital inventory updates for sustainable forest management~\cite{mattamala2025building, liang2016terrestrial, pierzchala2018mapping}. For practical deployment, localization systems must be robust to challenges specific to forests such as structural repetition and seasonal variation, provide centimeter-level accuracy, and rely on input representations that minimize storage and computational overhead in long-term, resource-constrained missions~\cite{chen2020sloam, liu2022large}.

While satellite-based localization systems such as GNSS are widely used in outdoor environments, they struggle in forests due to signal degradation under dense canopies~\cite{valbuena2010accuracy}. Even high-precision RTK services often fail to maintain consistent performance~\cite{lee2006performance}, resulting in meter-level errors, which are insufficient for tasks requiring tree-level correspondence such as growth estimation~\cite{dassot2011use}. These limitations underscore the need for onboard, geometry-based alternatives that do not rely on external positioning infrastructure; LiDAR is a promising solution due to its ability to capture precise 3D geometric information~\cite{cho2022application, wulder2008role}.

The standard LiDAR global localization pipeline typically involves place recognition for candidate retrieval~\cite{kim2018scan, kim2021scan, kim2024narrowing} followed by pose estimation through {local descriptor and feature matching}~\cite{yuan2022std, yuan2024btc, gupta2024effectively} or scan registration~\cite{yang2020teaser, lim2022single, lim2025kiss}. Although effective in structured environments, they incur high memory usage due to dense point cloud storage, and their performance degrades in forests, where repetitive geometric layouts cause perceptual aliasing and seasonal changes further alter scene appearance and consistency~\cite{oh2024evaluation}. This leads to more frequent retrieval errors and unstable pose estimates.

To mitigate reliance on dense point clouds and reduced discriminative power in forests, recent research has proposed compact, high-level representations as alternatives to raw point clouds~\cite{li2021ssc, dube2017segmatch}. Segmentation-based methods reduce storage by clustering geometrically or semantically similar points, but their discriminative power remains limited in forests due to low semantic diversity, where only a few semantic categories recur~\cite{vidanapathirana2025wildscenes}. Other methods such as NSM~\cite{tinchev2018seeing} and ESM~\cite{tinchev2019learning} extract descriptors from individual tree instances, but still rely on point clouds and overlook inter-tree relationships, making them prone to perceptual aliasing in environments where tree geometries are indistinguishable. These limitations motivate the use of more compact and interpretable representations that are better suited to the structural characteristics of forest environments.

A more compact alternative would be to use a \ac{DFI} directly for localization. A \ac{DFI} represents each tree by attributes such as stem axis and \ac{DBH} without requiring any raw point clouds. TreeLoc~\cite{jung2025treeloc} introduced a DFI-based framework that leverages inter-tree spatial relationships for matching. While efficient, it remains sensitive to structural aliasing, where locally similar tree layouts repeat across different parts of the forest, leading to incorrect matches and localization failures.

{To overcome these limitations, we propose TreeLoc++, a forest-specific localization system that builds on TreeLoc to address three key limitations of the original pipeline: ambiguous coarse retrieval in repetitive layouts, false correspondences caused by triangle-hash collisions, and unstable full 6-DoF refinement. Compared with TreeLoc, TreeLoc++ strengthens the localization pipeline at three stages. First, it augments the \ac{TDH} descriptor with a complementary \ac{PDH} that captures pairwise geometric context for more reliable candidate retrieval. Second, it refines tentative triangle matches through lightweight \ac{DBH} filtering and yaw-consistent inlier voting, reducing attribute- and pose-inconsistent correspondences. Third, it improves geometric verification and pose estimation through an overlap score with a spatial proximity prior and a constrained optimization that jointly estimates roll, pitch, and height. In particular, the overlap score discourages associations between geometrically similar but spatially distant scenes, while the joint refinement yields more stable 6-DoF localization than estimating these components independently.}

{These additions improve both downstream metric localization and retrieval robustness while preserving the lightweight design of \ac{DFI}-based localization. As shown in \figref{fig:main}, TreeLoc++ offers a lightweight and scalable alternative to conventional point cloud-based localization systems. By using only a compact \ac{DFI} as its prior map, it supports efficient multi-session alignment without retaining or revisiting raw scans. Its stem-based representation also enables accurate localization under seasonal change. This lightweight and stable localization pipeline supports continuous \ac{DFI} updates across sessions, enabling long-term forest monitoring and management. Our contributions are summarized as follows:}

\begin{itemize}

\item An advanced global localization framework for forests that resolves geometric ambiguity in tree-based configurations. 
Our framework also improves 6-DoF pose estimation by leveraging tree geometry and optimizing roll, pitch, and height simultaneously, enabling accurate localization in complex under-canopy conditions.

\item A retrieval pipeline combining \ac{TDH} and \ac{PDH}, which jointly encodes individual tree attributes and inter-tree connectivity. 
This combination strengthens coarse retrieval, ensuring high recall and increasing the fraction of true positives among the top-ranked candidates.

\item Robust outlier rejection through DBH-based filtering and yaw-consistent inlier voting.
These methods reduce ambiguous matches caused by hash collisions, enabling more consistent candidate retrievals for precise place recognition and pose estimation, particularly in scenarios with repeated or symmetric tree layouts.

\item Practical applicability across diverse localization scenarios such as inter-session localization, global map alignment, and continuous \ac{DFI} updates. This is enabled by an overlap score that identifies true positives across sessions without environment-specific tuning, ensuring reliable performance in real-world deployments.

\item Extensive benchmarking on three datasets with structural diversity and seasonal variations shows that TreeLoc++ consistently outperforms existing methods. 
Compared to the learning-based method LoGG3D-Net, it improves Recall@1 on \texttt{Evo} from 0.521 to {0.956} and reduces pose estimation error from decimeter-level to centimeter-level, while requiring only kilobytes of map storage with query times under \unit{10}{ms}.

\end{itemize}


%% file: 2_relatedwork.tex
\section{Related Work}
\label{sec:relatedwork}
\subsection{Digital Forest Inventories from Tree Segmentation}
LiDAR plays a central role in forestry, supporting biodiversity monitoring, biomass estimation, and sustainable forest management~\cite{pierzchala2018mapping, gleason2012forest}. Both static terrestrial LiDAR systems and mobile LiDAR platforms provide high-resolution 3D data for detailed forest analysis~\cite{liang2016terrestrial}. A key application is individual tree segmentation~\cite{li2012new, wallace2014evaluating, itakura2021estimating, malladi2024tree, freissmuth2024online}, addressed through traditional clustering and deep learning methods such as PointNet++~\cite{qi2017pointnet++} and RandLA-Net~\cite{hu2020randla}. Public forest datasets \cite{malladi2025digiforests, knights2022wild, vidanapathirana2025wildscenes, oh2024evaluation} have further accelerated research progress.

From segmented point clouds, tree attributes such as height, crown structure, stem position, orientation, and \ac{DBH} can be extracted \cite{bailey2018semi, freissmuth2024online}. These parameters form the basis of the \ac{DFI}, a compact, point cloud-independent representation that supports applications in precision forestry and long-term monitoring. RealtimeTrees~\cite{freissmuth2024online} introduced a learning-free pipeline for real-time DFI construction for mobile platforms, eliminating the need for post-processing. While DFI construction has become efficient, applying DFIs across multiple sessions introduces new challenges. Effective forest monitoring requires not only generating local inventories, but also localizing new observations against prior maps to detect revisits and to maintain correspondence between individual trees observed at different times. These capabilities are critical for enabling inventory updates, long-term monitoring, and consistent map integration, highlighting the need for robust global localization in forest environments~\cite{hussein2015global, valbuena2010accuracy}.

\begin{figure*}[!t]
    \centering
    \includegraphics[width=\textwidth]{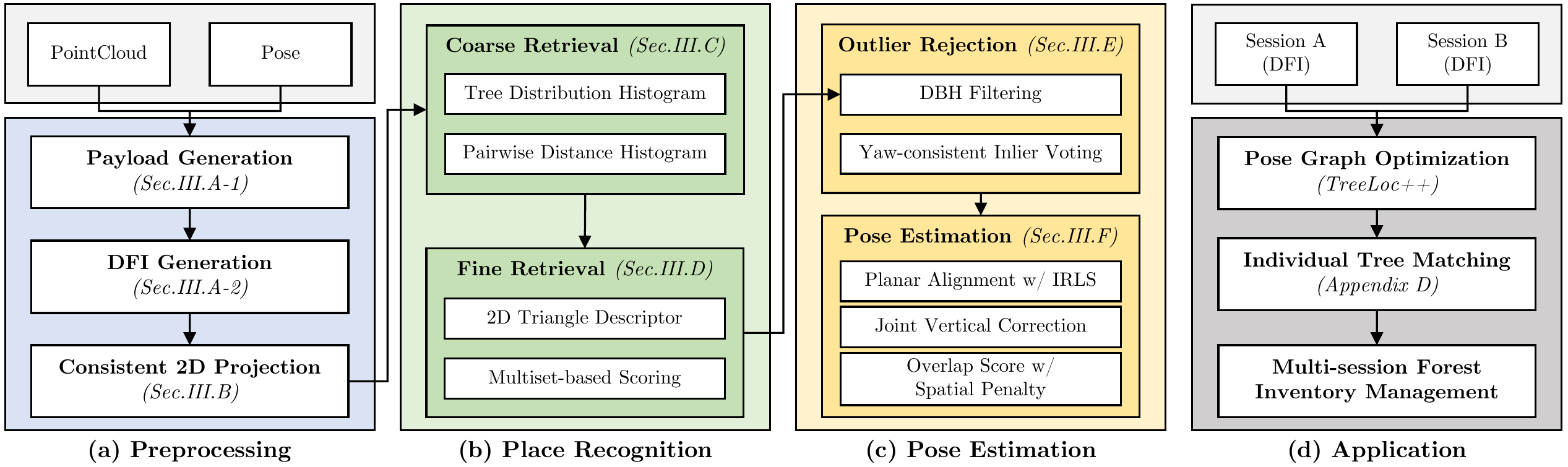}
    \caption{{(a) TreeLoc++ extracts tree-level traits using RealtimeTrees \cite{freissmuth2024online} and projects them onto a consistent 2D plane. (b) From the projected DFI, two histogram descriptors and a 2D triangle descriptor are generated for place recognition. (c) After fine retrieval, outliers from incorrect triangle matches are rejected using DBH consistency and yaw estimation, followed by full 6-DoF pose estimation and verification with an overlap score augmented by a spatial penalty. (d) TreeLoc++ enables tree association across multiple sessions, supporting consistent matching and incremental trait updates.}}
    \vspace{-3mm}
    \label{fig:pipeline}
\end{figure*}

\subsection{LiDAR-based Global Localization}
Accurate forest localization is challenging due to GNSS degradation under dense canopies, making LiDAR-based global localization a viable alternative.
This section reviews representative approaches based on hand-crafted descriptors and deep learning, and introduces inventory-based localization as a compact alternative. We highlight their limitations in forests to motivate our inventory-based formulation.

{\noindent\textbf{Hand-crafted Descriptors-based Localization:}
Conventional hand-crafted LiDAR place recognition methods are typically divided into global and local approaches, with global descriptors compressing a scan or submap into a compact representation for efficient nearest-neighbor retrieval~\cite{kim2018scan, kim2021scan, xu2023ring++, kim2024narrowing, he2016m2dp}. Representative examples include Scan Context++~\cite{kim2018scan, kim2021scan}, which encode scans on polar grids, RING++~\cite{lu2022one, xu2023ring++} with Radon transform-based yaw-translation decoupling, and SOLiD~\cite{kim2024narrowing}, formulated in azimuth-elevation space. While efficient in urban environments, these descriptors perform poorly in forests due to structural similarity and sensor variability, such as differences in angular resolution and mounting height, which lead to inconsistent projections and degraded robustness.}

{Local descriptors~\cite{rusu2009fast, salti2014shot, bosse2013place} aim to improve distinctiveness through salient geometric regions, but in forests, dense and structureless vegetation produces many indistinct keypoints, increasing computational cost and false positives. Projection-based methods such as MapClosure~\cite{gupta2024effectively, gupta2025efficiently} improve efficiency through ground-projected BEV matching, although their reliance on an observable ground plane often fails in natural environments. Triangle-based methods~\cite{yuan2022std, yuan2024btc, zhang2024intensity, park2025re} use hash-based indexing for efficient matching, but still degrade in forests due to unstable feature extraction from foliage and the lack of distinctive geometric structures.}


{\noindent\textbf{Learning-based Localization:}
To address the limitations of hand-crafted descriptors, learning-based methods exploit deep networks for localization. Point-based methods~\cite{xu2021transloc3d, vidanapathirana2022logg3d} such as TransLoc3D and LoGG3D-Net aggregate point-wise features into global descriptors, while sparse convolution methods~\cite{komorowski2022improving, jung2025helios} such as MinkLoc3Dv2 improve scalability. Transformer-based forest localization models~\cite{griffiths2025hotformerloc, shen2025forestlpr}, including HOTFormerLoc and ForestLPR, further capture long-range or vertical tree structure.}
{In parallel, projection-based approaches such as BEVPlace++~\cite{luo2025bevplace++} leverage 2D representations including BEV or range images~\cite{luo2021bvmatch, jung2025imlpr}, and some methods further integrate verification or relative pose estimation into the localization pipeline~\cite{vidanapathirana2023spectral, cattaneo2022lcdnet, komorowski2021egonn}. However, in forest environments, these approaches remain challenged by domain shift and limited generalization across forest types.}

{\noindent\textbf{Segment-based Localization:}
Segment-based methods represent scenes at the object level to improve robustness. Early approaches~\cite{dube2017segmatch,dube2020segmap} perform place recognition by extracting and matching 3D segments, later incorporating data-driven descriptors. Extending this segment-level perspective to natural environments, NSM~\cite{tinchev2018seeing} uses hand-crafted shape descriptors for segment matching, whereas ESM~\cite{tinchev2019learning} replaces the descriptor with a learned deep representation. Both have shown promising results in forests, but segment-level descriptors can remain ambiguous for geometrically similar tree trunks, especially without broader inter-segment spatial relationships. Although semantic graph-based methods incorporate object-level relationships for structural reasoning~\cite{pramatarov2022boxgraph, ma2024tripletloc}, their benefit in forests is limited by low semantic diversity, as only a few semantic classes repeatedly appear.}

{\noindent\textbf{Inventory-based Localization:}
These limitations motivate the use of more compact and forest-specific representations for localization.
To address these issues, recent work explores \acp{DFI} as compact, interpretable priors encoding tree attributes such as position, orientation, and \ac{DBH}. Early inventory-based approaches rely on stem-level geometric matching. For example, \citeauthor{tremblay2018towards}~\cite{tremblay2018towards} perform marker-free registration via stem-triplet matching, but because the method relies mainly on inter-stem distances and \ac{DBH} consistency, it remains vulnerable to structural ambiguity in repetitive forests. \citeauthor{liang2026ground}~\cite{liang2026ground} also exploit stem centers and triangle relationships, but their method is limited to planar alignment and does not incorporate richer tree attributes. Their method further relies on brute-force triangle matching and repeatedly applies the transformation estimated from each triangle correspondence to all stem centers for verification, resulting in high computational cost.}

{TreeLoc~\cite{jung2025treeloc} advances this direction by operating directly on \acp{DFI}, using \ac{TDH} and 2D triangle descriptors in a coarse-to-fine matching pipeline. However, it remains vulnerable to repetitive stem layouts, where \ac{TDH}-only coarse retrieval can miss pairwise layout cues and triangle-hash collisions can induce ambiguous correspondences. Compared with TreeLoc, TreeLoc++ strengthens the pipeline at three stages: it augments coarse retrieval with a complementary \ac{PDH}, refines correspondences through lightweight \ac{DBH} filtering and yaw-consistent inlier voting, and improves metric localization through overlap-based verification and joint refinement of roll, pitch, and height. These additions improve robustness in repetitive forests and long-term multi-session localization.}

%% file: 3_method.tex
\section{System of TreeLoc++}
\figref{fig:pipeline} shows the overall pipeline of TreeLoc++. The system begins with tree reconstruction from LiDAR scans and associated poses, generating a \ac{DFI}. Based on this representation, localization is performed in three stages: 
(i) preprocessing, which projects tree locations onto a consistent 2D plane estimated from stem axes;  
(ii) place recognition, which retrieves candidate revisits through a coarse-to-fine matching strategy; and  
(iii) pose estimation, which performs geometric verification and refinement using robust filtering methods.

\subsection{Digital Forest Inventory Construction}
\label{sec:reconstruction}
\subsubsection{Tree Extraction}
To ensure sufficient point density for tree segmentation and modeling, TreeLoc++ aggregates $k$ consecutive LiDAR scans into a local submap called a payload, where each payload shares $v$ scans with the previous one to maintain spatial continuity. Payloads are generated only when the system is moving, and their poses are obtained from a local trajectory provided by a LiDAR-based SLAM system.

\begin{figure}[!t]
    \centering    \includegraphics[width=1.0\columnwidth]{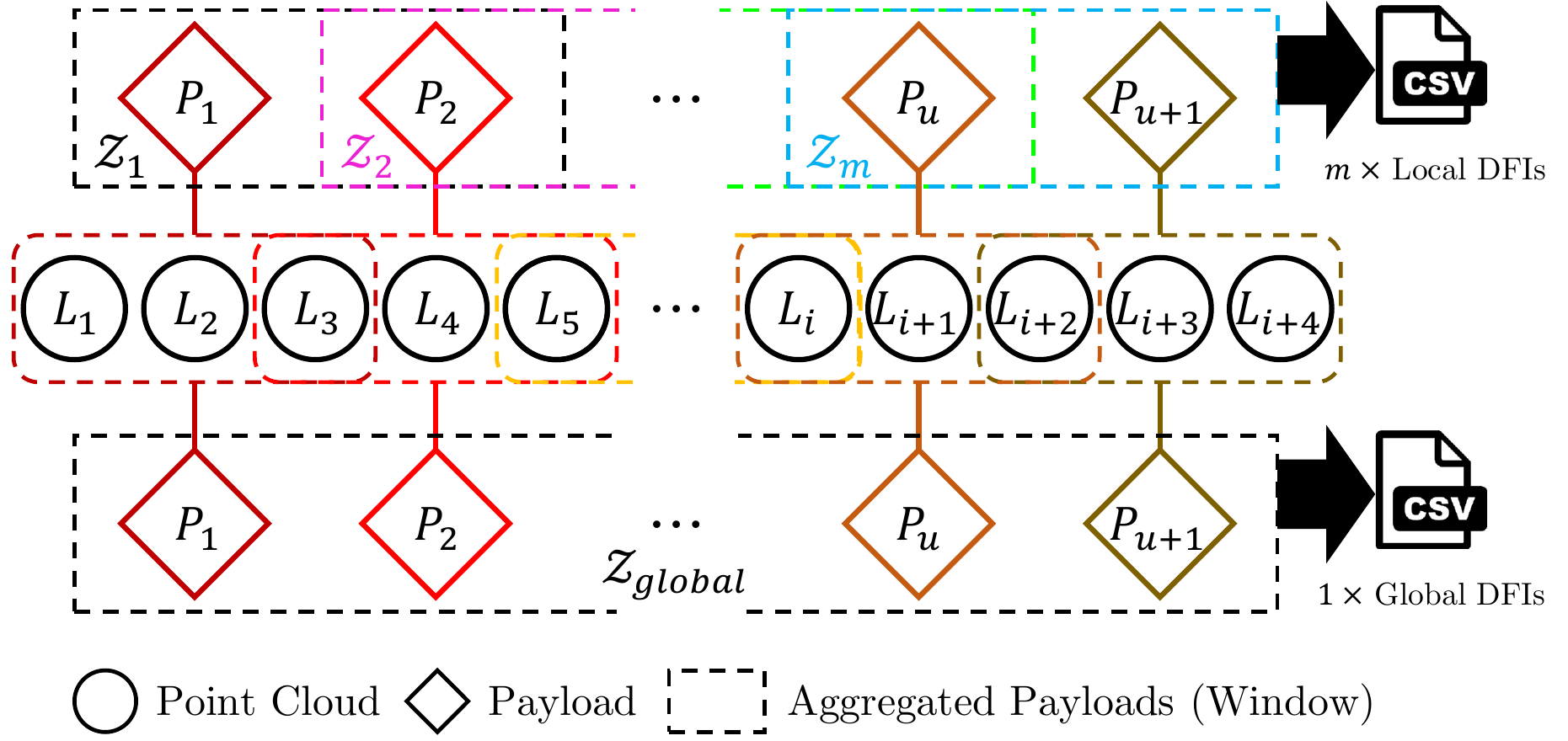}
    \caption{Payload-based forest inventory generation. (Top) Payloads are aggregated over temporal windows to form local inventories along the trajectory. (Bottom) All payloads are merged into a global inventory, which is queried at sampled poses to generate local inventories for localization.
}
    \label{fig:payload}
    \vspace{-3mm}
\end{figure}

Let $\mathcal{P}_u$ denote the payload indexed by $u$. For a given index $t$, we define a window of $s$ payloads as $\mathcal{W}_t = \{ t - \lfloor \frac{s-1}{2} \rfloor, \dots, t + \lfloor \frac{s-1}{2} \rfloor \}$. These payloads are transformed into a common local frame and aggregated into a submap $\mathcal{Z}_t = \bigcup_{u \in \mathcal{W}_t} \mathbf{T}_{t \leftarrow u} \cdot \mathcal{P}_u$, where $\mathbf{T}_{t \leftarrow u}$ transforms each payload into the reference frame of $\mathcal{P}_t$. This submap-based approach, illustrated in \figref{fig:payload}, provides the dense geometric input required for subsequent tree modeling.


Tree instances are extracted from $\mathcal{Z}_t$ using RealtimeTrees~\cite{freissmuth2024online}. {We refer the reader to~\cite{freissmuth2024online,jung2025treeloc} for the underlying extraction details. The extracted instances are then classified based on geometric completeness, where those satisfying minimum observation range and angular span requirements are regarded as successfully reconstructed trees, while the remainder are retained as candidate stems.} The resulting collection of trees and their geometric attributes constitutes the \ac{DFI}, which serves as the fundamental data structure for subsequent localization.

\subsubsection{Digital Forest Inventory for Localization}

TreeLoc++ operates in two inventory modes: local and global. As illustrated in the upper part of \figref{fig:payload}, the local mode aggregates payloads within a temporal window $\mathcal{W}_t$ to support incremental tasks such as SLAM loop closure or multi-robot localization. {By sliding this temporal window along the trajectory, a total of $m$ local inventories are generated.} Conversely, the global mode merges all mission payloads into a single window $\mathcal{W}_{\mathrm{global}}$, where trees are extracted once over the entire dataset. For localization, the global inventory is queried at selected poses, such as trajectory points or spatially sampled locations, to extract nearby trees within a fixed radius. Whether derived from local windows or queried from the global mode, the resulting inventory is transformed into the local coordinate frame of the corresponding pose to ensure consistent descriptor generation. This design enables the reuse of globally consistent tree data for large-scale, multi-session registration and long-term mapping applications.

For each reconstructed tree, we extract its 3D stem orientation $\mathbf{A}_j \in \mathrm{SO}(3)$ and \ac{DBH} $d_j$. The stem position $\mathbf{p}'_j \in \mathbb{R}^3$ is defined by its horizontal stem center $\mathbf{c}'_j$ and base height $\mathbf{b}'_j$, determined by the terrain elevation at the stem center. An inventory at time $t$ is represented as $\mathbf{M}_t = (\mathbf{T}_t, \mathcal{I}_t)$, where $\mathbf{T}_t$ is the system pose and $\mathcal{I}_t = \{ (\mathbf{A}_j, \mathbf{p}'_j, d_j) \}_{j=1}^{n_t}$ is the set of $n_t$ tree attributes. These attributes are computed from the aggregated payloads $\mathcal{Z}_t$ using RealtimeTrees.

{Unlike TreeLoc, which relies only on the current local inventory, TreeLoc++ hierarchically augments sparse local inventories using preceding observations when the reconstructed-tree count falls below a required threshold.} The augmentation integrates reconstructed trees from preceding inventories ($\mathcal{I}_{t-1}$ to $\mathcal{I}_{t-5}$), while filtering duplicates using the local trajectory. It compensates for limited local visibility and segmentation inconsistencies, and is omitted in global mode, where the comprehensive aggregation already provides sufficient coverage. If the tree count remains insufficient, candidate stems are incorporated based on their observation frequency to stabilize descriptor generation.

\subsection{Axis-Based Alignment and 2D Projection}
\label{sec:axis_alignment}

To generate compact descriptors from \ac{DFI}, TreeLoc++ projects stem centers onto a common 2D plane derived from the forest inventory $\mathcal{I}_t$. This removes height ambiguity while preserving inter-tree geometry for place recognition, but roll and pitch variations across viewpoints can distort the local $xy$-plane and lead to inconsistent projections.

\begin{figure}[!t]
    \centering    \includegraphics[width=.99\columnwidth]{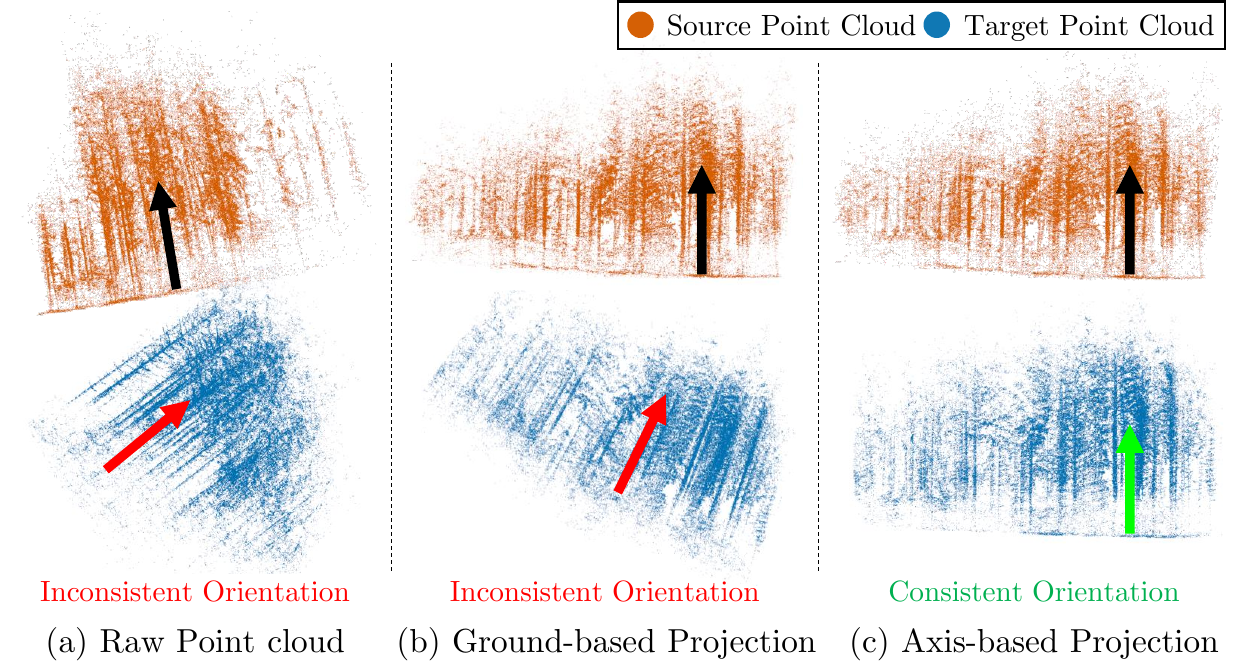}
    \caption{Comparison of alignment strategies for 2D projection. {Arrows indicate the vertical direction of the ground plane after transformation. (a) Raw point clouds show inconsistent directions, with the target (red) misaligned with the source (black). (b) Ground-based alignment remains inconsistent across views. (c) Axis-based alignment using tree stems yields consistent directions, with the source and target aligned (green).}}
    \label{fig:align}
    \vspace{-3mm}
\end{figure}

Ground-plane fitting is often used to correct such distortion but is unreliable in forests with uneven or partially observed terrain. As shown in \figref{fig:align}, ground-based alignment may work in some views but fail in others due to inconsistent terrain estimation. In contrast, alignment using stem axes provides a consistent frame across varying viewpoints.
We therefore align each frame using the stem axes of reconstructed trees. For each tree $j$, we extract a unit vector $\mathbf{a}_j$ from the third column of its orientation matrix $\mathbf{A}_j \in \mathrm{SO}(3)$. A rotation $\mathbf{R}^A_t \in \mathrm{SO}(3)$ is then estimated to align the axes $\{\mathbf{a}_j\}$ to a reference direction $\mathbf{v} \in \mathbb{S}^2$ by solving
\begin{equation}
\label{eq:axis_align_general}
\mathbf{R}^A_t =
\underset{\mathbf{R} \in \mathrm{SO}(3)}{\arg\min}
\sum_j \left( 1 - \left| \mathbf{v}^\top \mathbf{R}\mathbf{a}_j \right| \right)^2.
\end{equation}

We set $\mathbf{v} = \mathbf{e}_z = (0, 0, 1)^\top$ in practice, although other reference directions can also be used. {As in TreeLoc, this alignment corrects roll and pitch while leaving yaw unconstrained, thereby preserving stable horizontal inter-tree geometry.} After estimating $\mathbf{R}^A_t$, each stem center $\mathbf{p}'_j$ is transformed to the aligned frame as $\mathbf{p}_j = \mathbf{R}^A_t \mathbf{p}'_j$.

To obtain 2D coordinates for descriptor generation, the aligned stem $\mathbf{p}_j$ are projected onto the plane orthogonal to $\mathbf{v}$ using an orthonormal basis $\{\mathbf{u}_1, \mathbf{u}_2\}$ satisfying $\mathbf{u}_i^\top \mathbf{v} = 0$, resulting in projected coordinates $\mathbf{c}_j = \begin{bmatrix} \mathbf{u}_1^\top;\ \mathbf{u}_2^\top \end{bmatrix} \mathbf{p}_j \in \mathbb{R}^2$.
In the special case where $\mathbf{v} = \mathbf{e}_z$, this simplifies to:
\begin{equation}
\label{eq:projection_xy}
\mathbf{c}_j =
\begin{bmatrix}
1 & 0 & 0 \\
0 & 1 & 0
\end{bmatrix}
\left( \mathbf{p}_j - (\mathbf{e}_z^\top \mathbf{p}_j)\mathbf{e}_z \right),
\end{equation}
yielding a roll- and pitch-invariant 2D representation.

\subsection{Coarse Retrieval via Tree-Based Histograms}
\begin{figure}[!t]
    \centering
    \begin{subfigure}[t]{\columnwidth}
        \centering
        \includegraphics[width=\linewidth]{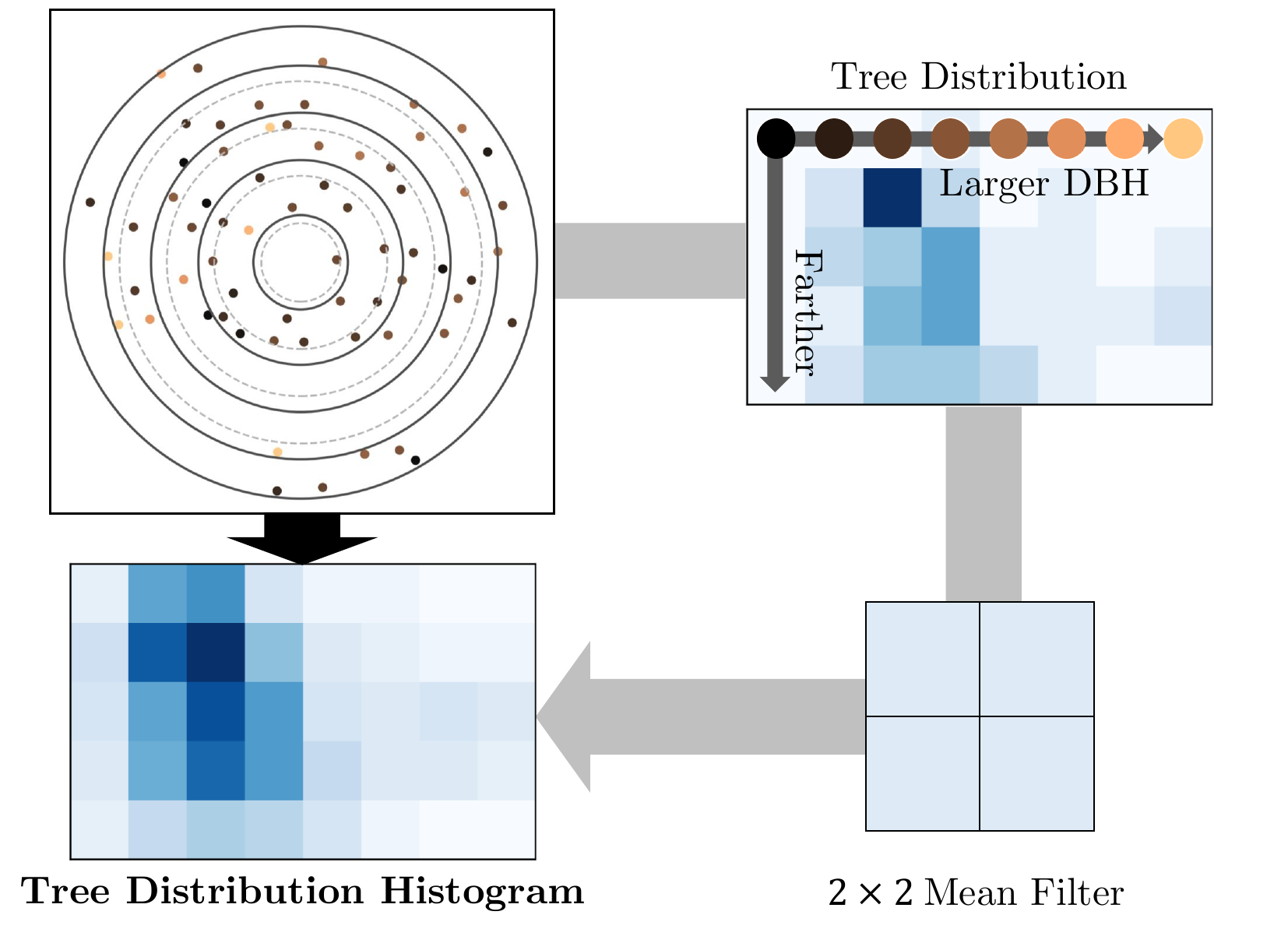}
        \vspace{-6mm}
        \caption{Tree Distribution Histogram (TDH) Generation}
        \label{fig:histogram_a}
    \end{subfigure}\hfill
    \begin{subfigure}[t]{\columnwidth}
        \centering
        \includegraphics[width=\linewidth]{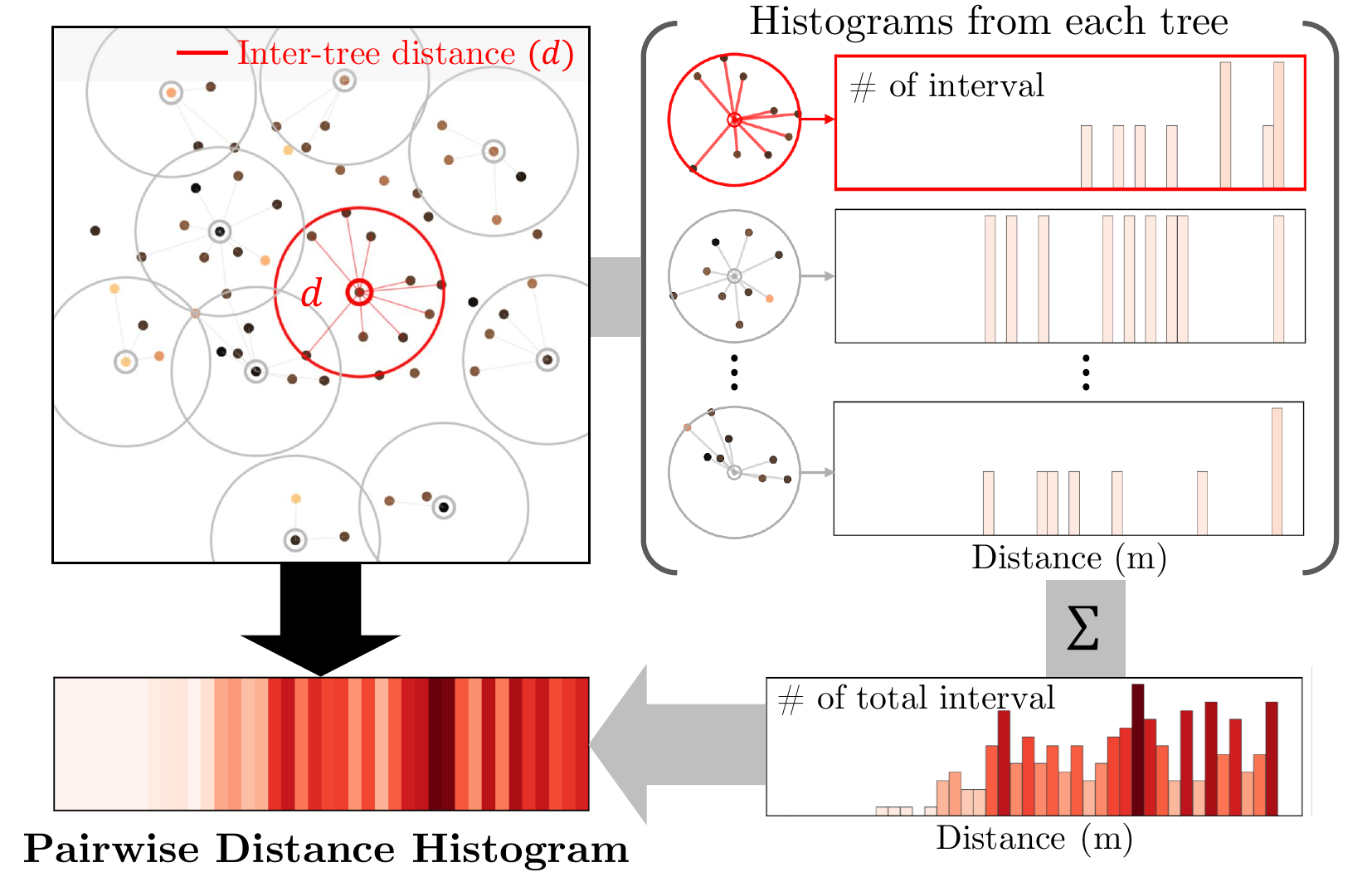}
        \vspace{-6mm}
        \caption{Pairwise Distance Histogram (PDH) Generation}
        \label{fig:histogram_b}
    \end{subfigure}
    \vspace{-2mm}
    \caption{Generation of histogram descriptors. 
    (a) TDH encodes tree distributions using overlapping radial and DBH bins. Each radial bin spans from a dashed circle to the second subsequent solid circle, illustrating the overlapping construction. (b) PDH aggregates inter-tree distance histograms computed for each tree into a 1D scene-level representation.}
    \label{fig:histogram}
    \vspace{-3mm}
\end{figure}

To enable efficient and robust place recognition, we summarize the spatial structure of each inventory using two complementary histogram descriptors, \ac{TDH} and \ac{PDH}, computed from the reconstructed trees $\mathcal{I}_t$, as illustrated in \figref{fig:histogram}.

Firstly, \ac{TDH} encodes 2D spatial structure by assigning each tree to overlapping radial and DBH bins based on its stem center $\{\mathbf{c}_j\}$ and diameter $\{d_j\}$. Radial bins partition $[r_{\min}, r_{\max}]$ into $n_r$ intervals with overlap $w_r$, while DBH bins partition $[d_{\min}, d_{\max}]$ into $n_d$ intervals with overlap $w_d$. The resulting histogram $\mathbf{H}_t \in \mathbb{R}^{n_r \times n_d}$ is smoothed with a $2 \times 2$ uniform kernel and flattened into a 40-dimensional descriptor. Overlap parameters and smoothing improve robustness to noise and pose variation.

While TDH captures tree attributes via spatial binning, it lacks inter-tree relationships and is sensitive to the choice of reference center, varying with missing trees or slight shifts in their positions. To address this, we introduce PDH, a 1D histogram that encodes local structure by aggregating pairwise Euclidean distances between stem centers. Distances are binned over $[l_{\min}, l_{\max}]$ into $n_{\text{bins}}$ intervals, yielding a 40-dimensional descriptor invariant to rotation and translation. Although structural perturbations may affect some distances, the aggregation distributes their impact across many bins, reducing sensitivity to local changes. {Compared with TDH, PDH is generally more stable under slight translation deviations and missing-tree perturbations because it depends on pairwise distances rather than a reference-centered layout. However, since it does not explicitly encode the spatial distribution of trees around the local center, PDH is not always more discriminative than TDH as a standalone descriptor. We therefore use it to complement, rather than replace TDH.}

For each query inventory at index $t$, both descriptors are compared against the database. Let $\mathbf{h}^{\text{TDH}}_t$ and $\mathbf{h}^{\text{PDH}}_t$ denote the query descriptors, and $\mathbf{h}^{\text{TDH}}_i$, $\mathbf{h}^{\text{PDH}}_i$ those of a candidate $i$. Descriptor distances are computed using the chi-square metric $\chi^2$ as $d^{\text{TDH}}_i = \chi^2(\mathbf{h}^{\text{TDH}}_t,\ \mathbf{h}^{\text{TDH}}_i)$ and $d^{\text{PDH}}_i = \chi^2(\mathbf{h}^{\text{PDH}}_t,\ \mathbf{h}^{\text{PDH}}_i)$. Each distance is min-max normalized to balance the influence of TDH and PDH, with the result denoted as $\hat{d}$. The retrieval score is defined as $\text{score}_i = -(\hat{d}^{\text{TDH}}_i + \hat{d}^{\text{PDH}}_i)$, such that lower descriptor distances yield higher scores. The top 100 candidates are retained for fine-grained retrieval.

\subsection{2D Triangle Descriptor Using Tree Centers}

To refine candidates from histogram-based coarse retrieval, we introduce a 2D triangle descriptor that is rotation- and translation-invariant, capturing local inter-tree geometry. Following geometric hashing principles~\cite{yuan2024btc}, the descriptor is constructed from the aligned 2D stem centers.

For each center $\mathbf{c}_i$, we select its $m$ nearest neighbors and form triangles by combining $\mathbf{c}_i$ with each unordered neighbor pair $(\mathbf{c}_j, \mathbf{c}_k)$. Each triangle is represented by its side lengths $(\ell_{ij}, \ell_{jk}, \ell_{ik})$, sorted in ascending order as $(\ell_1, \ell_2, \ell_3)$ for permutation invariance, together with its area $A$. All geometric quantities are quantized with resolution $\delta_\ell$ as $\ell_i^{\mathrm{q}} = \lfloor \ell_i / \delta_\ell \rfloor$ and $A^{\mathrm{q}} = \lfloor A / \delta_\ell \rfloor$. The final triangle hash is computed by:
\begin{equation}
\begin{aligned}
h'_{ijk} &= \left( \left( \ell_3^{\mathrm{q}} \cdot \rho + \ell_2^{\mathrm{q}} \right) \bmod U \cdot \rho + \ell_1^{\mathrm{q}} \right) \bmod U, \\
h_{ijk} &= \left( h'_{ijk} \cdot \rho + A^{\mathrm{q}} \right) \bmod U.
\end{aligned}
\end{equation}
where $\rho$ is a large prime and $U$ is the maximum hash range. These hashes index triangles for efficient matching.

Metadata $\mathbf{D}_{ijk} = \{\mathcal{A}_i,\ \mathcal{A}_j,\ \mathcal{A}_k,\ \mathbf{q}_{ijk},\ s \}$ is stored in each triangle, where $\mathcal{A}$ contains tree attributes, $\mathbf{q}_{ijk}$ is the triangle centroid, and $s$ is the scene index. {This metadata is directly used in the later TreeLoc++ stages for outlier rejection and geometric verification.}

For each scene $s$, we store the multiset of triangle hashes $\mathcal{H}_s=\{h_{ijk}\}$. Let $\mathcal{K}_s$ denote the set of unique hash keys in $\mathcal{H}_s$. During fine retrieval, a query scene $Q$ is compared with a candidate scene $C$ by counting shared triangle hashes:
\begin{equation}
\label{eq:multiset}
S(Q,C) =
\sum_{h \in \mathcal{K}_Q \cap \mathcal{K}_C}
\min \bigl( \mathrm{freq}_Q(h),\ \mathrm{freq}_C(h) \bigr),
\end{equation}
where $\mathrm{freq}_Q(h)$ and $\mathrm{freq}_C(h)$ are the occurrences of hash $h$ in $\mathcal{H}_Q$ and $\mathcal{H}_C$, respectively. Unlike simple set intersection {used in TreeLoc}, multiset-based scoring accounts for repeated geometric patterns, improving similarity discrimination. The top 10 candidates ranked by $S(Q,C)$ proceed to geometric verification.

\subsection{Outlier Rejection via DBH Filtering and Yaw Voting}
\label{sec:inlier_sets}

To extract an inlier set for geometric verification, we refine the triangle matches. Ambiguous hash collisions can produce false correspondences, which often manifest as attribute- or pose-inconsistent matches. We suppress these outliers through DBH filtering and yaw voting.

\subsubsection{Outlier Rejection via DBH Filtering}
When a common hash value appears multiple times, it is unclear which triangles should be matched, as shown in \figref{fig:intra_triangle}. Let $m$ and $n$ be the numbers of triangles in the query and candidate scenes sharing a given hash. In such cases, at most $\min(m,n)$ valid correspondences can exist.

To resolve this ambiguity, we use the triangle DBH from metadata $\mathbf{D}$. Let $\mathbf{d}_Q^{i} = (d_{Q1}^{i}, d_{Q2}^{i}, d_{Q3}^{i})$ and $\mathbf{d}_C^{j} = (d_{C1}^{j}, d_{C2}^{j}, d_{C3}^{j})$ be the DBH vectors of the $i$-th and $j$-th triangles. Since DBH vectors follow the sorted side length order, we can directly compare them and define the matching cost as
\begin{equation}
\Delta_{ij} = \sum_{v=1}^3 \left| d_{Qv}^{i} - d_{Cv}^{j} \right|.
\end{equation}
We select up to $\min(m,n)$ one-to-one correspondences with minimum total cost and reject pairs violating $\tau_{\mathrm{DBH}}$:
\begin{equation}
\max_{v \in \{1,2,3\}} \left| d_{Qv}^{i} - d_{Cv}^{j} \right| < \tau_{\mathrm{DBH}}.
\end{equation}

\subsubsection{Yaw-Consistent Inlier Voting}
\begin{figure}[!t]
    \centering
    \begin{subfigure}[t]{\columnwidth}
        \centering
        \includegraphics[width=\linewidth]{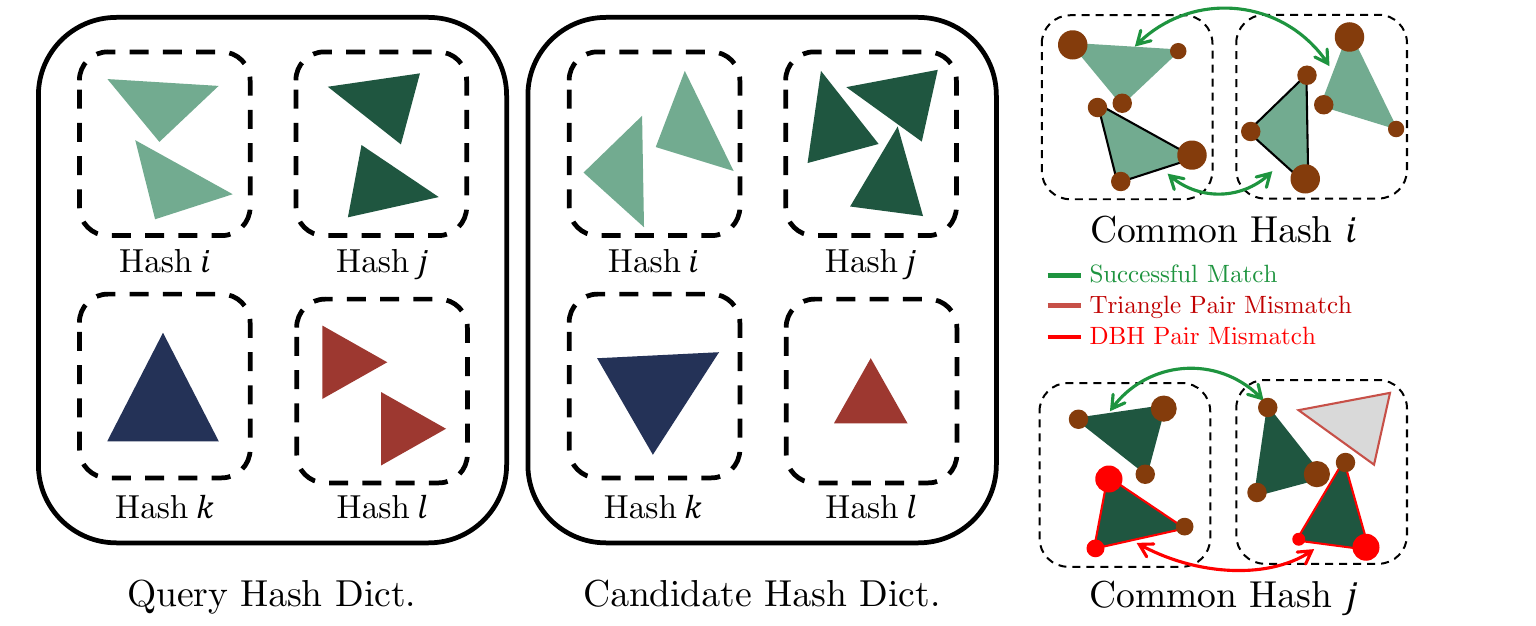}
        \caption{DBH Filtering for Hash Collision}
        \label{fig:intra_triangle}
    \end{subfigure}\hfill
    \begin{subfigure}[t]{\columnwidth}
        \centering
        \includegraphics[width=\linewidth]{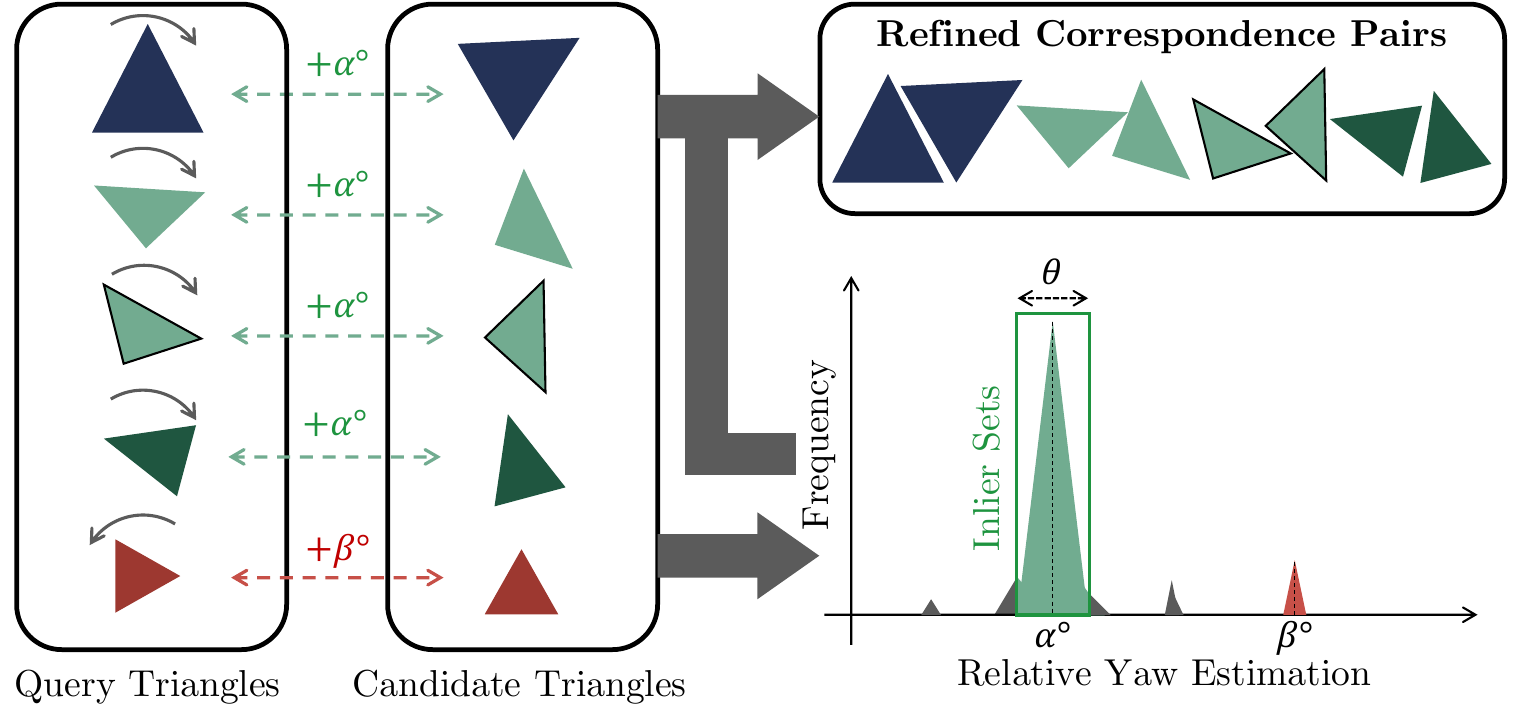}
        \caption{Yaw Voting for Outlier Suppression}
        \label{fig:inter_triangle}
    \end{subfigure}
    \vspace{-3mm}
    \caption{Inlier set selection via DBH filtering and yaw voting. (a) Ambiguous triangle correspondences sharing the same hash are disambiguated using DBH consistency. (b) Yaw angles from remaining correspondences are aggregated, and only those consistent with the dominant yaw are retained.}
    \label{fig:Fig2}
    \vspace{-3mm}
\end{figure}

Even after DBH filtering, repetitive local structures may still include incorrect correspondences that induce erroneous pose estimates. To enforce global consistency, we analyze the relative yaw rotation implied by each remaining triangle match as shown in \figref{fig:inter_triangle}.

For each matched triangle pair, a 2D rigid transformation is estimated using \ac{SVD}-based alignment, from which the relative yaw $\theta_k$ is extracted.
We exclude matches that yield a reflection with $\det(\mathbf{R}_k)<0$. The remaining angles are accumulated into a histogram to identify the dominant yaw $\theta^*$. The final inlier set is defined as:
\begin{equation}
\mathcal{I} =
\left\{ k \;\middle|\; |\theta_k - \theta^*| < \tau_{\mathrm{yaw}} \right\},
\end{equation}
where $\tau_{\mathrm{yaw}}$ is a predefined angular threshold.

This two-step process, DBH-based disambiguation followed by yaw voting, yields correspondences consistent in both local attributes and global pose. The inlier set is then used for geometric verification and 6-DoF pose estimation.

\subsection{Geometric Verification and 6-DoF Pose Estimation}
\label{sec:geom_verif}

We estimate the 6-DoF relative transformation between a query scene $\mathcal{I}_Q$ and a candidate scene $\mathcal{I}_C$. {Following the shared geometric verification backbone of TreeLoc, TreeLoc++ strengthens the refinement with IRLS-based planar refinement, tree axis-based roll-pitch estimation, joint vertical correction, and a spatial penalty-aware overlap score.} This process is applied to the top 10 candidates retrieved via triangle hash matching, and the best match is selected through geometric verification. An overview of the pipeline is shown in \figref{fig:pose_estimation_pipeline}.

\subsubsection{Planar Alignment via Inlier Triangles and IRLS Refinement}
Using the inlier triangle correspondences, we extract the matched centroids $\{(\mathbf{q}_u^Q, \mathbf{q}_u^C)\}_{u=1}^N$ from metadata $\mathbf{D}$ and estimate an initial planar transformation $\mathbf{T}_{C \leftarrow Q}^{\mathrm{init}} \in \mathrm{SE}(2)$, parameterized by $(\mathbf{R}^{\mathrm{init}}_{2\text{D}}, \mathbf{t}^{\mathrm{init}}_{2\text{D}})$, via least-squares alignment:
\begin{align}
\label{eq:init_svd}
\mathbf{R}_{2\text{D}}^{\mathrm{init}}, \mathbf{t}_{2\text{D}}^{\mathrm{init}}
&= \arg\min_{\mathbf{R}, \mathbf{t}}
\sum_{u=1}^{N}
\left\| \mathbf{q}_u^C - (\mathbf{R} \mathbf{q}_u^Q + \mathbf{t}) \right\|^2.
\end{align}
{TreeLoc++ then further refines this initial planar estimate using iteratively reweighted least squares (IRLS) to suppress residual outliers before subsequent 6-DoF refinement.} IRLS is initialized with $(\mathbf{R}_{2\text{D}}^{\mathrm{init}}, \mathbf{t}_{2\text{D}}^{\mathrm{init}})$, and we refine the planar transformation over all vertex-level matches $(\mathbf{c}_i^Q, \mathbf{c}_i^C)$ in corresponding triangles:

\begin{figure}[!t]
    \centering
    \includegraphics[width=\columnwidth]{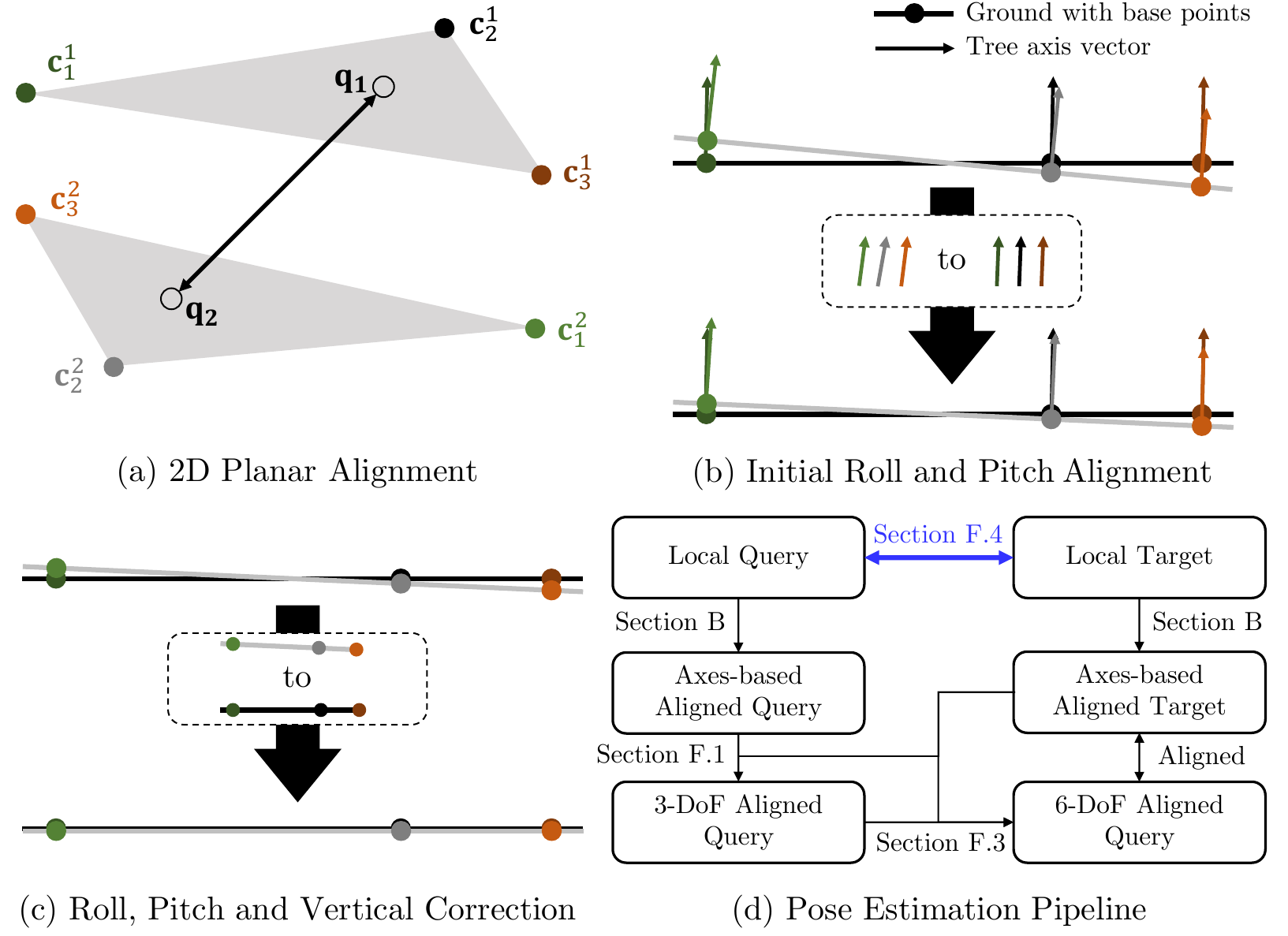}    
    \caption{Pose estimation pipeline. (a) Initial 2D planar alignment is estimated using triangle correspondences. (b) Roll and pitch are aligned using the axes of matched trees. (c) Vertical correction is performed by jointly refining height together with roll and pitch. (d) The final 6-DoF transformation is obtained, summarizing the overall pose estimation process.}
    \label{fig:pose_estimation_pipeline}
    \vspace{-3mm}
\end{figure}

\begin{align}
\label{eq:irls}
\mathbf{R}_{2\text{D}}, \mathbf{t}_{2\text{D}}
&= \arg\min_{\mathbf{R}, \mathbf{t}} \sum_i w_i
\left\| \mathbf{c}_i^C - (\mathbf{R} \mathbf{c}_i^Q + \mathbf{t}) \right\|^2 .
\end{align}
where weights $w_i$ follow a Huber kernel based on residuals.

\subsubsection{Tree-Level Correspondences and Roll-Pitch Estimation}

Applying the planar alignment, each query $\mathbf{c}_i^Q$ is transformed as $\widetilde{\mathbf{c}}_i = \mathbf{R}_{2\text{D}} \mathbf{c}_i^Q + \mathbf{t}_{2\text{D}}$. Tree-level correspondences are then established by finding candidate matches $\mathbf{c}_j^C$ via nearest-neighbor search, subject to spatial and DBH thresholds:
\begin{equation}
\left\|\widetilde{\mathbf{c}}_i - \mathbf{c}_j^C\right\| < \tau_d, \quad |d_i^Q - d_j^C| < \tau_{\text{DBH}}.
\end{equation}
Using the matched pairs $\mathcal{M}=\{(i,j)\}$, we solve \eqref{eq:irls} once to update $(\mathbf{R}_{2\text{D}}, \mathbf{t}_{2\text{D}})$. {To support the subsequent vertical correction, TreeLoc++ next estimates residual roll and pitch from tree-axis correspondences. The same set $\mathcal{M}$ is used to estimate roll and pitch by aligning axis vectors $\mathbf{a}_i^Q$ and $\mathbf{a}_j^C$.} With yaw fixed by $\mathbf{R}_z$ extracted from $\mathbf{R}_{2\text{D}}$, we estimate the residual roll-pitch rotation $\mathbf{R}_{\phi,\psi}$ via RANSAC over tree axis correspondences:
\begin{align}
\mathbf{R}_{\phi,\psi}
&= \arg\min_{\mathbf{R}}
\sum_{(i,j)\in\mathcal{M}}
\left( 1 - (\mathbf{a}_j^C)^\top \mathbf{R}\mathbf{R}_z \mathbf{a}_i^Q \right).
\end{align}

\subsubsection{Vertical Correction and Final Scoring}

{TreeLoc++ further introduces a joint vertical correction of height, roll, and pitch using matched tree base heights.} Thanks to the roll-pitch estimation using tree axis, the remaining roll-pitch error is small. This justifies the use of a small-angle approximation when modeling vertical misalignment.
For a 3D point $(x, y, z)$, the effect of roll ($\phi$) and pitch ($\psi$) on the vertical coordinate is given by
\begin{equation}
z' = -x \cos\phi \sin\psi + y \sin\phi + z \cos\phi \cos\psi.
\end{equation}
Yaw is omitted as it does not affect the vertical component. For small $\phi$ and $\psi$, this approximates to $z' \approx z - \psi x + \phi y$.

Using tree base heights, the vertical relationship between a matched tree pair $(i,j)\in\mathcal{M}$ is approximated as
\begin{equation}
b_j^C \approx \tilde{b}_{i}^Q, \quad 
\tilde{b}_{i}^Q = \hat{b}_i^Q + \Delta z - \Delta\psi \,\hat{x}_i^Q + \Delta\phi \,\hat{y}_i^Q,
\end{equation}
where $(\Delta z,\Delta\phi,\Delta\psi)$ denote the global vertical offset and small residual roll-pitch corrections, and $(\hat{x}_i,\hat{y}_i)$ and $\hat{b}_i^Q$ denote the planar coordinates and base height of the query tree after applying the planar alignment $\mathbf{T}_{2\text{D}}$ and the roll-pitch correction $\mathbf{R}_{\phi,\psi}$.
The parameters $(\Delta z, \Delta\phi, \Delta\psi)$ are estimated by minimizing the residual
$r_{ij} = b_j^C - \tilde{b}_{i}^Q$
over all matched pairs $(i,j)\in\mathcal{M}$ using a RANSAC-based least-squares estimator for robustness.
The resulting correction is expressed as a 3D transformation:
\begin{equation}
\mathbf{T}_{6\text{D}} =
\begin{bmatrix}
\mathbf{R}_{\Delta\phi,\Delta\psi} \mathbf{R}_{\phi,\psi} & [0,0,\Delta z]^\top \\
\mathbf{0} & 1
\end{bmatrix}
\mathbf{T}_{3\text{D}},
\end{equation}
where $\mathbf{T}_{3\text{D}}$ denotes the 2D alignment $\mathbf{T}_{2\text{D}}$ lifted to $\mathrm{SE}(3)$.

\subsubsection{Final 6-DoF Transformation in Local Frames}
All transformations so far are computed in the reference-aligned frame, where roll and pitch are aligned using tree axis-based 2D projection.
To recover the full 6-DoF transformation between the original local frames, we undo this alignment using the corresponding rotations $\mathbf{R}^A_Q$ and $\mathbf{R}^A_C$.

Let $\mathbf{T}^A_Q = \mathrm{diag}(\mathbf{R}^A_Q, 1)$ and $\mathbf{T}^A_C = \mathrm{diag}(\mathbf{R}^A_C, 1)$.  
The final transformation from the query to the candidate frame is given by $\mathbf{T}^{6\text{D}}_{C \leftarrow Q} = (\mathbf{T}^A_C)^{-1} \, \mathbf{T}_{6\text{D}} \, \mathbf{T}^A_Q$. This yields a global 6-DoF relative pose purely from tree-level geometric information.

To assess alignment quality and select the best match, we compute an overlap score combining geometric consistency and spatial proximity:
\begin{equation}
\label{eq:overlap_score}
\mathcal{O}(Q,C) =
\frac{|\mathcal{M}_{Q,C}|}
{|\mathcal{T}_Q| + |\mathcal{T}_C| - |\mathcal{M}_{Q,C}|}
\cdot p(\|\mathbf{t}\|),
\end{equation}
where $|\mathcal{M}_{Q,C}|$ is the number of matched trees, and $|\mathcal{T}_Q|$, $|\mathcal{T}_C|$ are the tree counts in the query and candidate scenes.

The penalty term $p(\|\mathbf{t}\|)= \exp\left( - {\|\mathbf{t}\|^2} / {\sigma_t^2} \right)$ {introduces a spatial prior that favors candidates with smaller estimated planar displacement $\mathbf{t}$ from $\mathbf{T}^{6\text{D}}_{C \leftarrow Q}$}, where $\sigma_t$ is a scaling parameter controlling the spatial constraint.
This reflects the intuition that large translation shifts often lead to discrepancies in reconstruction quality; as the same tree is viewed from substantially different distances or angles, varying degrees of occlusion and sparsity degrade its geometric consistency. The candidate with the highest score $\mathcal{O}(Q,C)$ is selected, along with its estimated transformation $\mathbf{T}^{6\text{D}}_{C \leftarrow Q}$.

%% file: 4_experiment.tex
\section{Datasets and Evaluation Metrics}
\input{tab/dataset}
\subsection{Dataset Summary}
\begin{figure}[!t]
    \centering
    \includegraphics[width=\columnwidth]{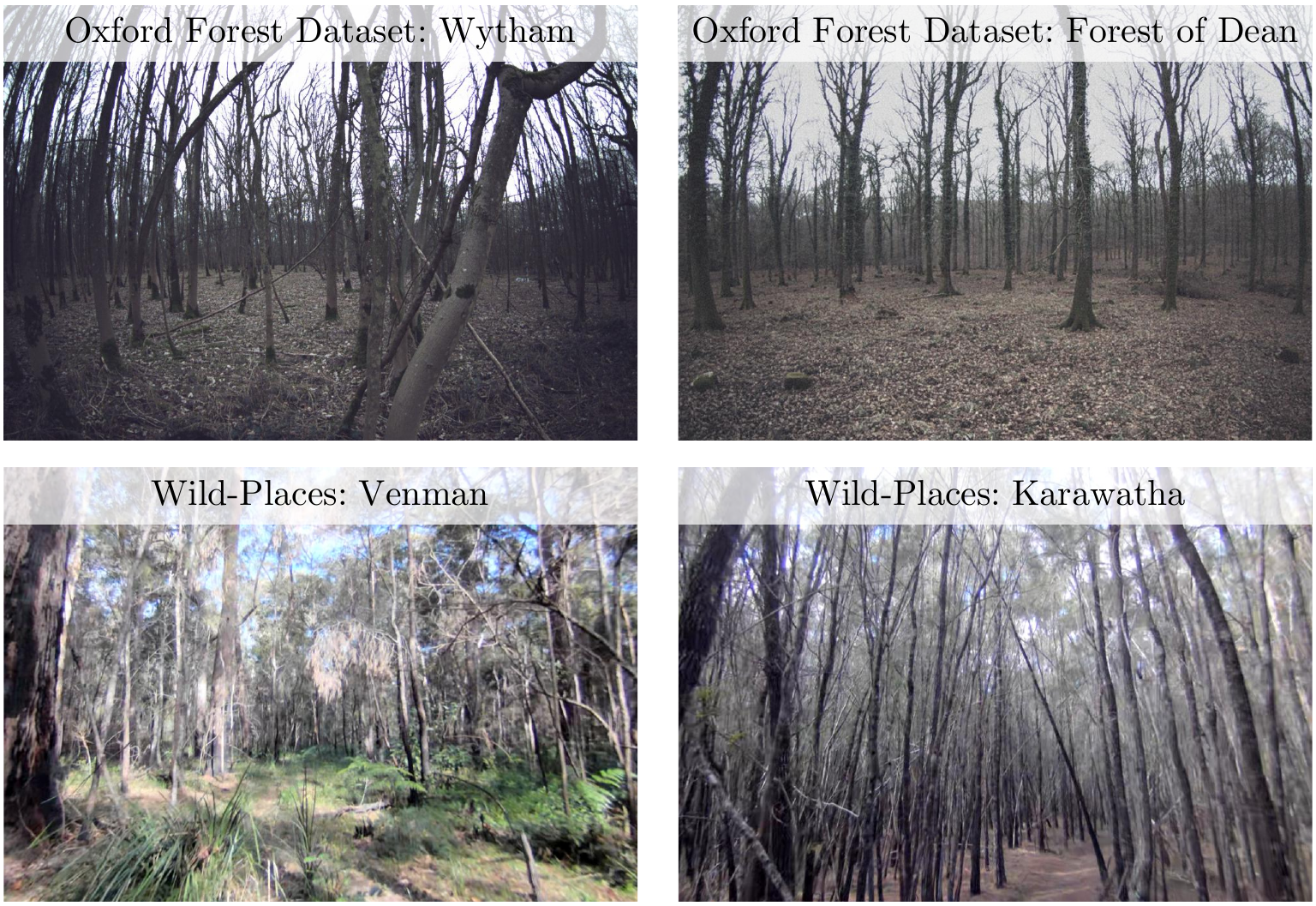}    
    \vspace{-5mm}
    \caption{Example scenes for each dataset and sequence.}
    \label{fig:dataset_overall_description}
    \vspace{-3mm}
\end{figure}

We evaluated TreeLoc++ on 27 sequences collected across diverse forest environments, including both complete plot mappings and point-to-point route traversals. Dataset characteristics are summarized in \tabref{tab:dataset_summary} and \figref{fig:dataset_overall_description}.

We first utilized the Oxford Forest Place Recognition dataset~\cite{oh2024evaluation}, which contains ten sequences collected from small forest plots across three countries.
All data were recorded using a backpack-mounted platform that incurs substantial roll and pitch motion while being carried during recording, presenting a further challenge for localization.

\begin{figure}[!t]
    \centering
    \includegraphics[width=\columnwidth]{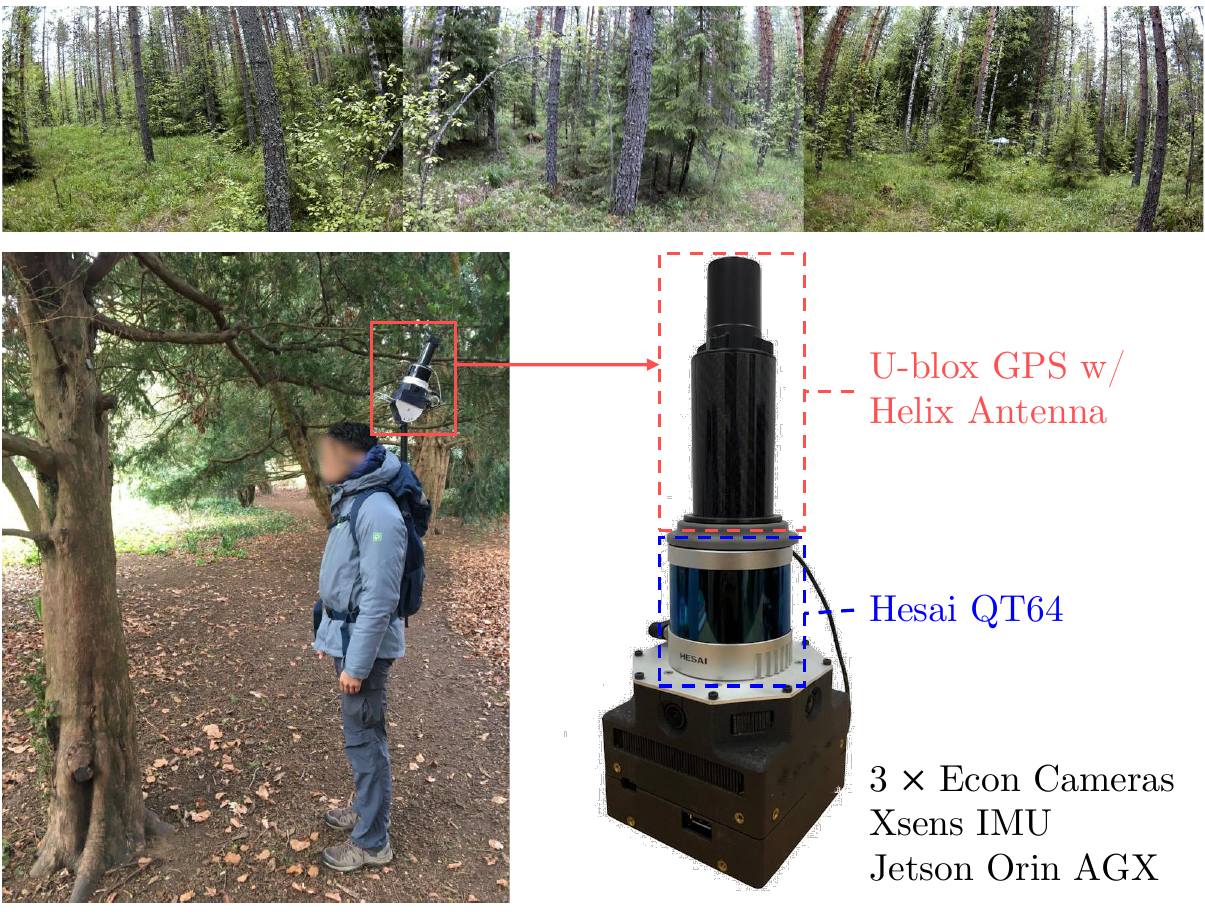}    
    \vspace{-5mm}
    \caption{Overview of the \texttt{Evo25} and acquisition platform. (Top) Example scenes captured in \texttt{Evo25}. (Bottom) Backpack-based data acquisition system equipped with a LiDAR, IMU, GPS antenna, and three cameras.}
    \label{fig:dataset_description}
    \vspace{-3mm}
\end{figure}

To assess long-term performance under pronounced appearance changes, we also introduce nine newer sequences, \texttt{Evo25:00-08}, collected two years later in the same forest as \texttt{Evo:Single}. This dataset captured environmental changes caused by vegetation growth and seasonal variation, facilitating the evaluation of long-term global localization. Ground truth trajectories for \texttt{Evo25} were obtained via multi-session alignment, with details provided in \textcolor{blue}{Appendix A}.
An overview of the data acquisition setup is shown in \figref{fig:dataset_description}.

To complement the plot inventory datasets, we used the Wild-Places dataset~\cite{knights2022wild}, which consists of eight sequences recorded in two large-scale forest parks in Australia: \texttt{Venman01-04} and \texttt{Karawatha01-04}. These sequences span substantially longer, forest access road trajectories with temporal gaps, enabling large-scale evaluation.

\subsection{Baselines and Evaluation Protocol}

To ensure fair evaluation, we standardized input preprocessing across all methods, including trajectory estimation and submap construction.
We first estimated the poses used for payload aggregation and forest inventory generation.
To reflect a realistic setup where localization operates alongside online state estimation, we used FAST-LIO2~\cite{xu2022fast} for \texttt{Evo:Single} and \texttt{Stein am Rhein}, and DLIO~\cite{chen2023direct} for the remaining Oxford sequences. For Wild-Places, ground truth poses were used.
Then, scans were aggregated into payloads and accumulated over time to form pose-centric submaps, from which forest inventories were generated.
All methods used consistent spatial cropping centered at the query pose, with regions of \unit{30}{m}$\times$\unit{30}{m} for Oxford and \unit{60}{m}$\times$\unit{60}{m} for Wild-Places to account for its larger scale.

In our intra-session evaluation, both queries and candidates used local inventory mode.
For inter-session evaluation, we considered three configurations: local-local, local-global (with uniformly sampled global poses), and global-global (with inventories from a reference trajectory).
For global baselines, the full point clouds were merged to form the global map and then the same cropping was applied during the matching. We compared TreeLoc++ to baselines for place recognition, metric localization, and pose estimation.

\input{tab/intra-session}

\noindent\textbf{Place Recognition:}
We compared TreeLoc++ with learning-based and hand-crafted methods.
Learning-based methods included point-based approaches (TransLoc3D~\cite{xu2021transloc3d}, MinkLoc3Dv2~\cite{komorowski2022improving}, LoGG3D-Net~\cite{vidanapathirana2022logg3d}, HOTFormerLoc~\cite{griffiths2025hotformerloc}) and the BEV-based methods such as BEVPlace++~\cite{luo2025bevplace++} and ForestLPR~\cite{shen2025forestlpr}. TransLoc3D and BEVPlace++ were trained on Wild-Places, while the others used pretrained checkpoints.
Hand-crafted baselines included global descriptor methods  (Scan Context+~\cite{kim2021scan}, SOLiD~\cite{kim2024narrowing}, RING++~\cite{xu2023ring++}), local descriptor methods (BTC~\cite{yuan2024btc}, MapClosure~\cite{gupta2025efficiently}), {segment-based method (NSM~\cite{tinchev2018seeing})}, and inventory-based method such as TreeLoc~\cite{jung2025treeloc}. {Because NSM localizes queries directly in the global map rather than via scan-to-scan retrieval, we used the nearest database scan to its estimated query pose as the retrieval result.}
A retrieval was considered correct if within \unit{5}{m} of the ground truth; additional analyses under varying distance thresholds are provided in \textcolor{blue}{Appendix E}. 
We report Recall@1 (R@1), Maximum Recall at 100\% Precision (MR), Maximum F1 score (MF1), and Area Under the Precision-Recall Curve (AUC).

\noindent\textbf{Metric Localization:}
We evaluated whether each method could recover an accurate pose estimation.
LoGG3D-Net and MinkLoc3Dv2 were extended with their feature-based localization, while BEVPlace++ includes its own localization module.
Hand-crafted baselines included TreeLoc, RING++, MapClosure, BTC {and NSM}.
As BEVPlace++ and RING++ only estimate $(x,y,\text{yaw})$, we report both 2D and 3D results.
We report Recall@50cm (R@50): the fraction of queries whose estimated pose is within \unit{50}{cm} and $5^\circ$ of ground truth, regardless of retrieval correctness.
The Success Rate (SR) adds the requirement that the underlying retrieval must also be a true positive.
For these successful cases, we report the Average Translation Error (ATE) and Average Rotation Error (ARE).
Further benchmark results about pose estimation with registration methods are provided in \textcolor{blue}{Appendix C}.

\section{Results}
\label{sec:experiment}
\subsection{Intra-session Place Recognition}

\subsubsection{Oxford Forest: Small and Dense Off-road Forests}
\label{sec:intra_experiment}
We begin our evaluation with intra-session cases. The first experiment focuses on intra-session place recognition using the Oxford Forest dataset, which consists of short, small-scale off-road sequences characterized by frequent revisits and substantial variations in viewpoint and occlusion.

\noindent \textbf{Evaluation protocol: }We compared TreeLoc++ against hand-crafted and learning-based baselines on four sequences. To avoid trivial matches, we excluded the 50 most recent frames from the candidate database.

\begin{figure}[!t]
    \centering
    \includegraphics[width=\columnwidth]{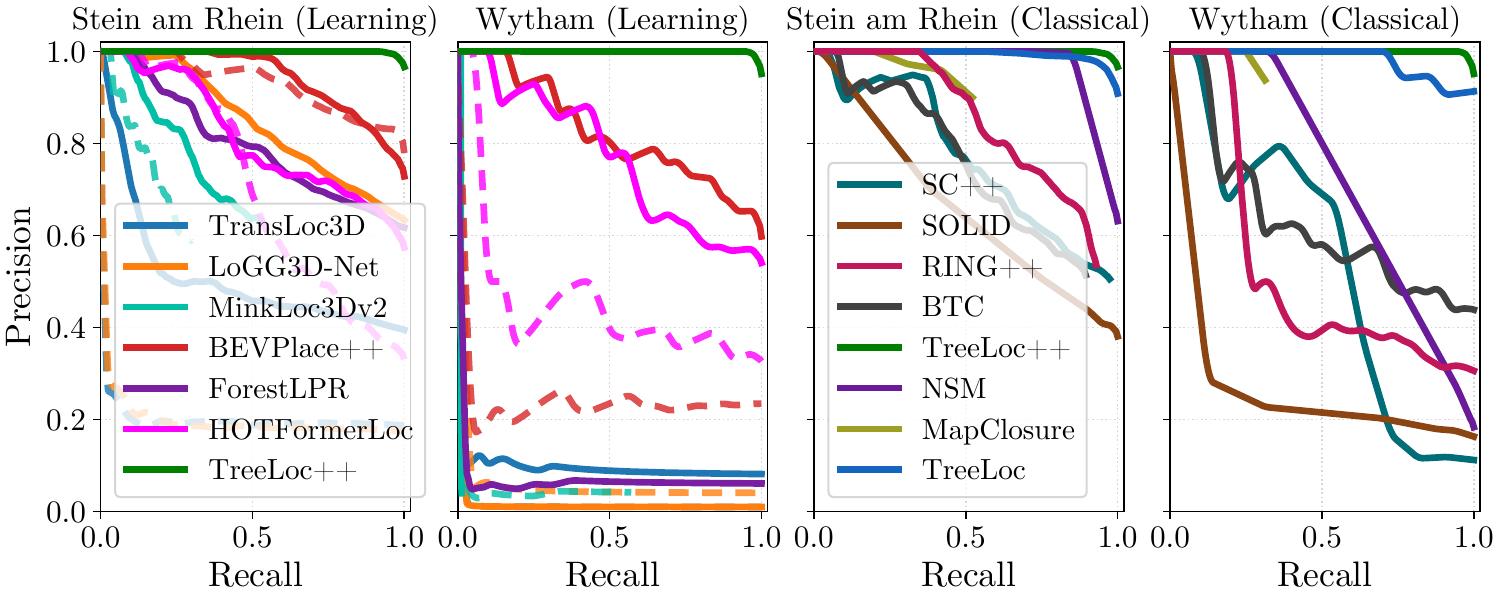}    
    \vspace{-5mm}
    \caption{(\textbf{Exp A-1}) Precision--Recall curves comparing TreeLoc++ with learning-based (left) and hand-crafted (right) methods. Some learning-based methods outperform hand-crafted baselines but remain below TreeLoc++, while hand-crafted methods show a sharp precision drop at low recall. TreeLoc++ achieves a higher AUC than TreeLoc.}
    \label{fig:auc_curve}
    \vspace{-4mm}
\end{figure}

\begin{figure}[!t]
    \centering
    \begin{subfigure}[t]{\columnwidth}
        \centering
        \includegraphics[clip=true, trim={0 170 0 0},width=\linewidth]{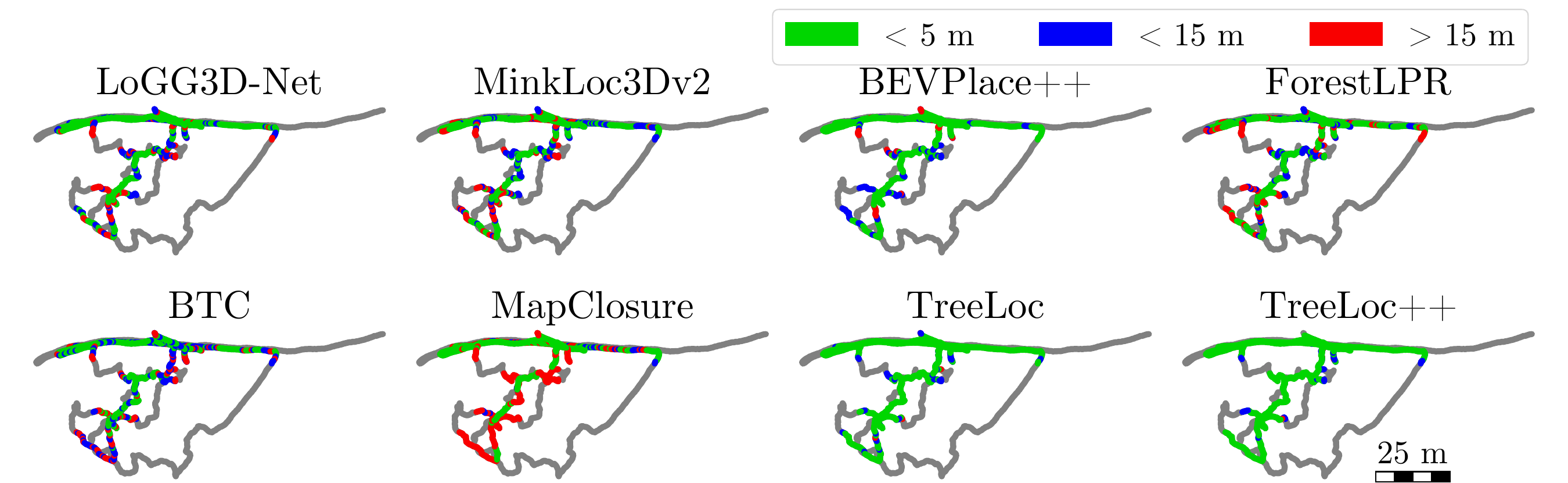}
        \caption{Intra-session Matching in \texttt{Stein am Rhein}}
        \label{fig:matching_stein}
    \end{subfigure}\hfill
    \vspace{-1mm}
    \begin{subfigure}[t]{\columnwidth}
        \centering
        \includegraphics[clip=true, trim={0 140 0 25},width=\linewidth]{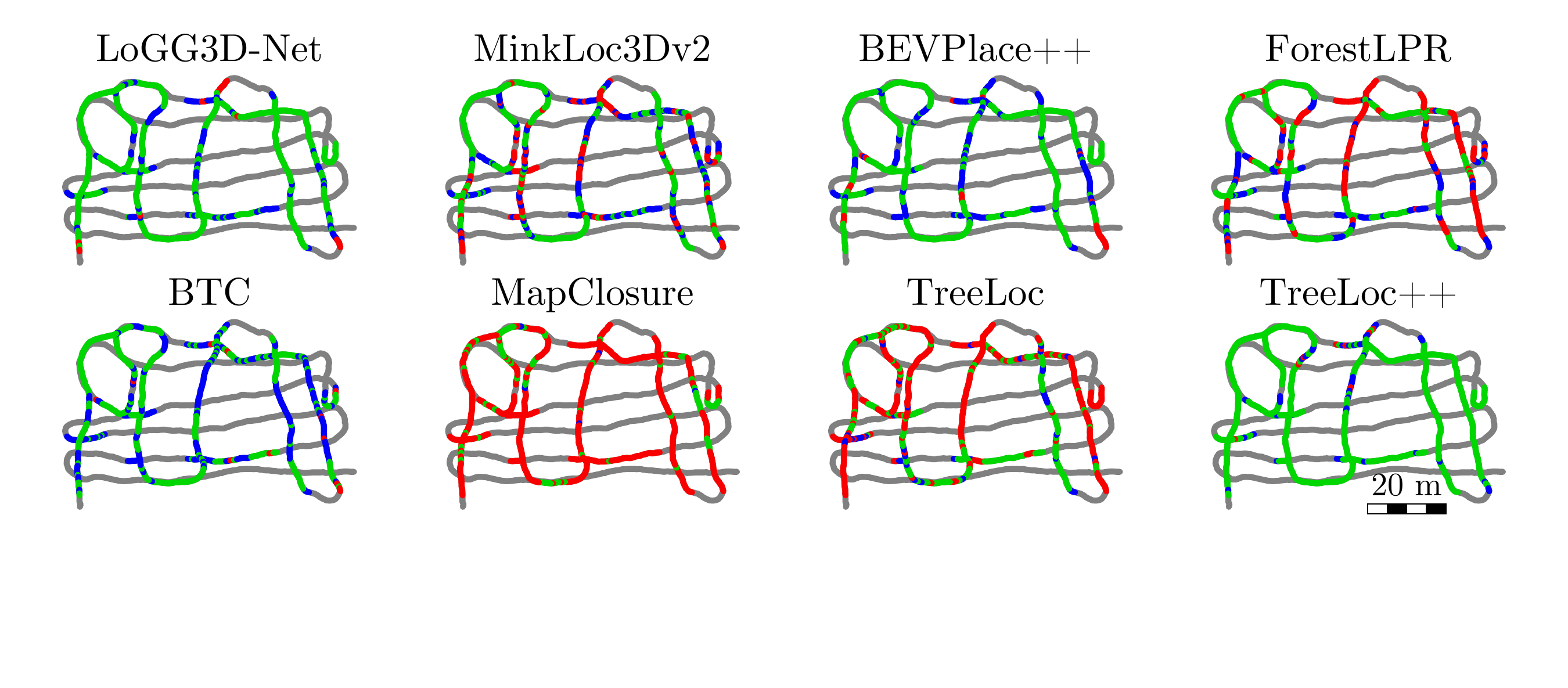}
        \caption{Intra-session Matching in \texttt{Forest of Dean}}
        \label{fig:matching_dean}
    \end{subfigure}
    \vspace{-3mm}
    \caption{(\textbf{Exp A-1}) Matching results in two sequences. TreeLoc++ identifies the most true positives (green). In failure cases, it still retrieves spatially adjacent candidates (blue) rather than distant false positives (red).}
    \label{fig:matching}
    \vspace{-4mm}
\end{figure}

\noindent \textbf{TreeLoc++: }TreeLoc++ achieved the best performance across all metrics (\tabref{tab:pr_intra_oxford_evo_dean}), maintaining high precision and the highest MR (\figref{fig:auc_curve}). This superior MR is primarily driven by the translation penalty in the overlap score defined in \eqref{eq:overlap_score}. Under frequent revisits and occlusions, where candidates exhibit similar geometry, this penalty favors spatially closer matches and improves true positive retrieval.
On \texttt{Forest of Dean}, TreeLoc++ preserved high R@1 despite repetitive geometry; its multiset-based scoring introduced in \eqref{eq:multiset} preserves each tentative triangle match, consistently outperforming TreeLoc as shown in \figref{fig:matching}.

\noindent \textbf{Hand-crafted baselines: }Hand-crafted descriptors gave poor performance in these forest sequences.
Global descriptor-based methods performed poorly, as compressing entire scans into compact global features limits discriminative power in vertically repetitive and structurally ambiguous scenes.
Likewise, local descriptor-based methods suffered from viewpoint sensitivity and unstable keypoint selection. {The segment-based NSM performed strongly on \texttt{Evo:Single} due to direct localization against the global map, but still lagged behind TreeLoc++ and degraded on the other sequences because correspondences among geometrically similar tree trunks remain ambiguous and inter-segment spatial relationships are not explicitly modeled.}
Furthermore, as seen in the PR curves in \figref{fig:auc_curve}, the baselines exhibited abrupt precision drops or early termination, indicating sensitivity to outliers and reduced matchability.

\noindent \textbf{Learning-based baselines:}
Learning-based methods show improved performance over hand-crafted baselines as shown in \tabref{tab:pr_intra_oxford_evo_dean} but still suffer from low MR and limited accuracy. While BEVPlace++ is the most competitive, its performance drops sharply in cluttered scenes like \texttt{Wytham} (\figref{fig:auc_curve}). As shown in \figref{fig:matching}, these methods yield fewer true positives and more near-matches (blue points) than TreeLoc++, failing to distinguish locations accurately. This is primarily due to a persistent domain shift of the training dataset, where inherent structural variations between forests prevent the extraction of robust, generalizable features. A detailed analysis of the impact of training data is provided in \textcolor{blue}{Appendix B}.

While the evaluation targets a small-scale setting with frequent revisits, we next consider a more challenging scenario with larger spatial extent and stronger structural aliasing.

\subsubsection{Wild-Places: Large-scale Forests with Larger Databases}
To further validate intra-session place recognition in a more challenging forest environment, we extend our experiments to the Wild-Places dataset. Compared to Oxford, Wild-Places features longer trajectories, larger databases, and fewer revisits, increasing the overall complexity of the task. Moreover, the data are collected along forest access roads, making the dataset less suitable for tree reconstruction.
The two sites present complementary characteristics: \texttt{Venman} exhibits dense, structurally repetitive scenes, whereas \texttt{Karawatha} contains open areas with sparse geometric features.

\input{tab/intra-session-wild}

\begin{figure}[!t]
    \centering
        \includegraphics[width=\columnwidth]{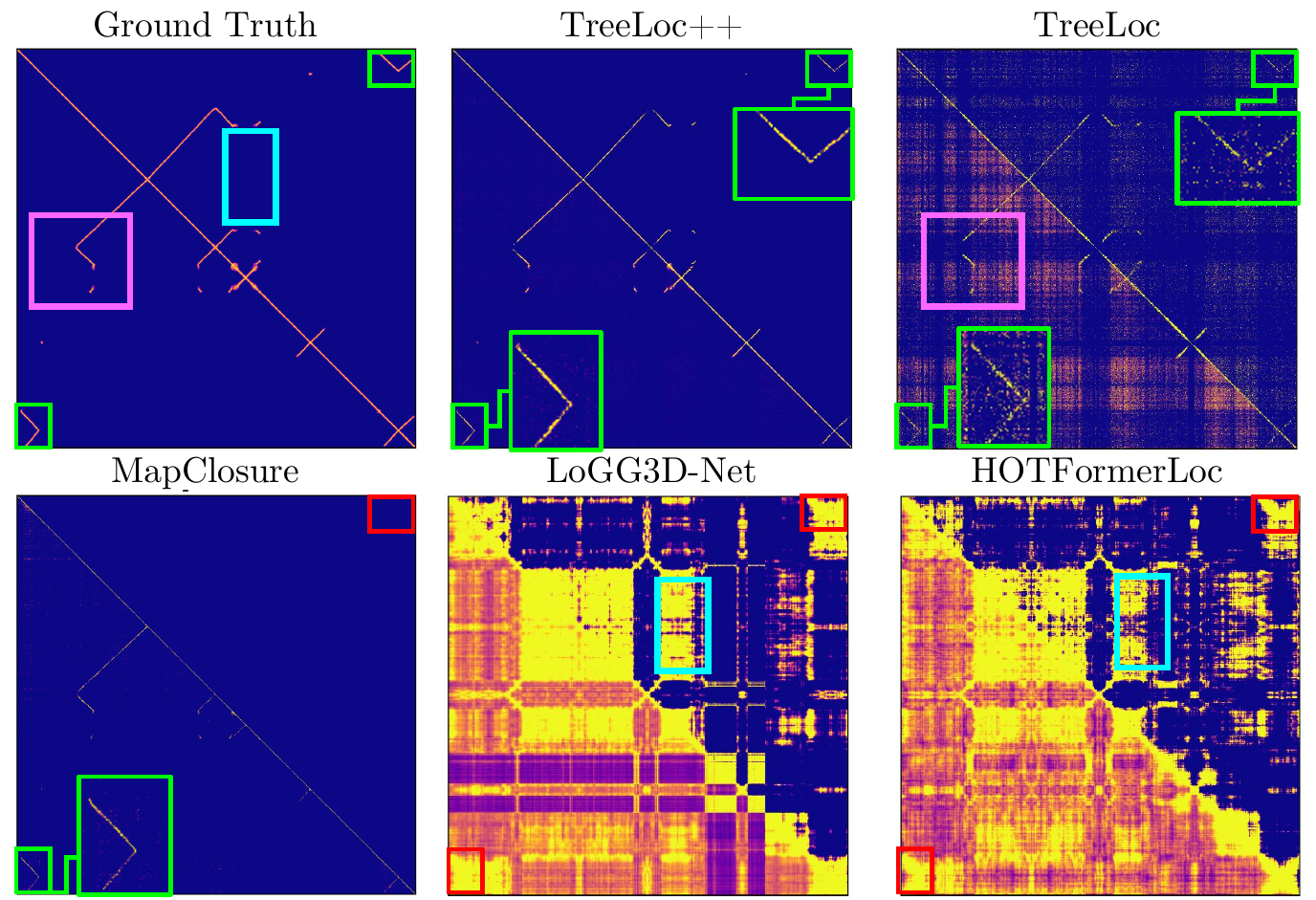} 
    \caption{(\textbf{Exp A-2}) Feature similarity matrices for \texttt{Karawatha04} (warmer colors indicate higher similarity). The {bottom-left and top-right triangular regions} show mid (20--80\%) and high (60--80\%) similarities, respectively. TreeLoc++ aligns well with ground truth{, particularly in the light-green boxed regions}; others show noisy (pink), low-contrast (red), or spuriously high similarities across unrelated regions (cyan).}

    \label{fig:feature_similarity}
    \vspace{-4mm}
\end{figure}

\begin{figure*}[!t]
    \centering
    \begin{subfigure}[t]{.611\textwidth}
        \centering
        \includegraphics[width=\linewidth]{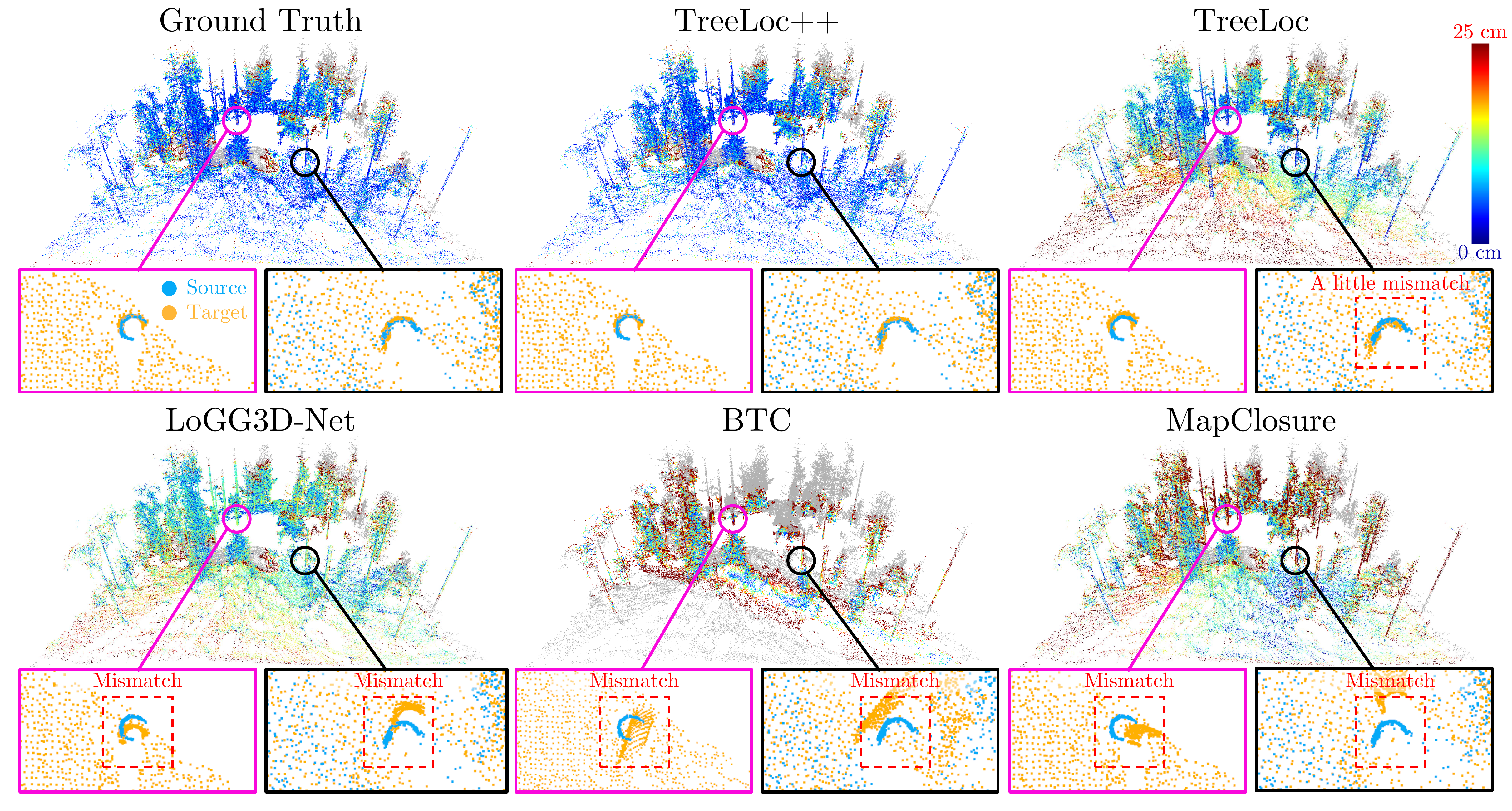}
        \caption{Registration Error in \texttt{Evo:Single}}
        \label{fig:intra_registration_a}
    \end{subfigure}
    \begin{subfigure}[t]{.38\textwidth}
        \centering
    \includegraphics[width=\linewidth]{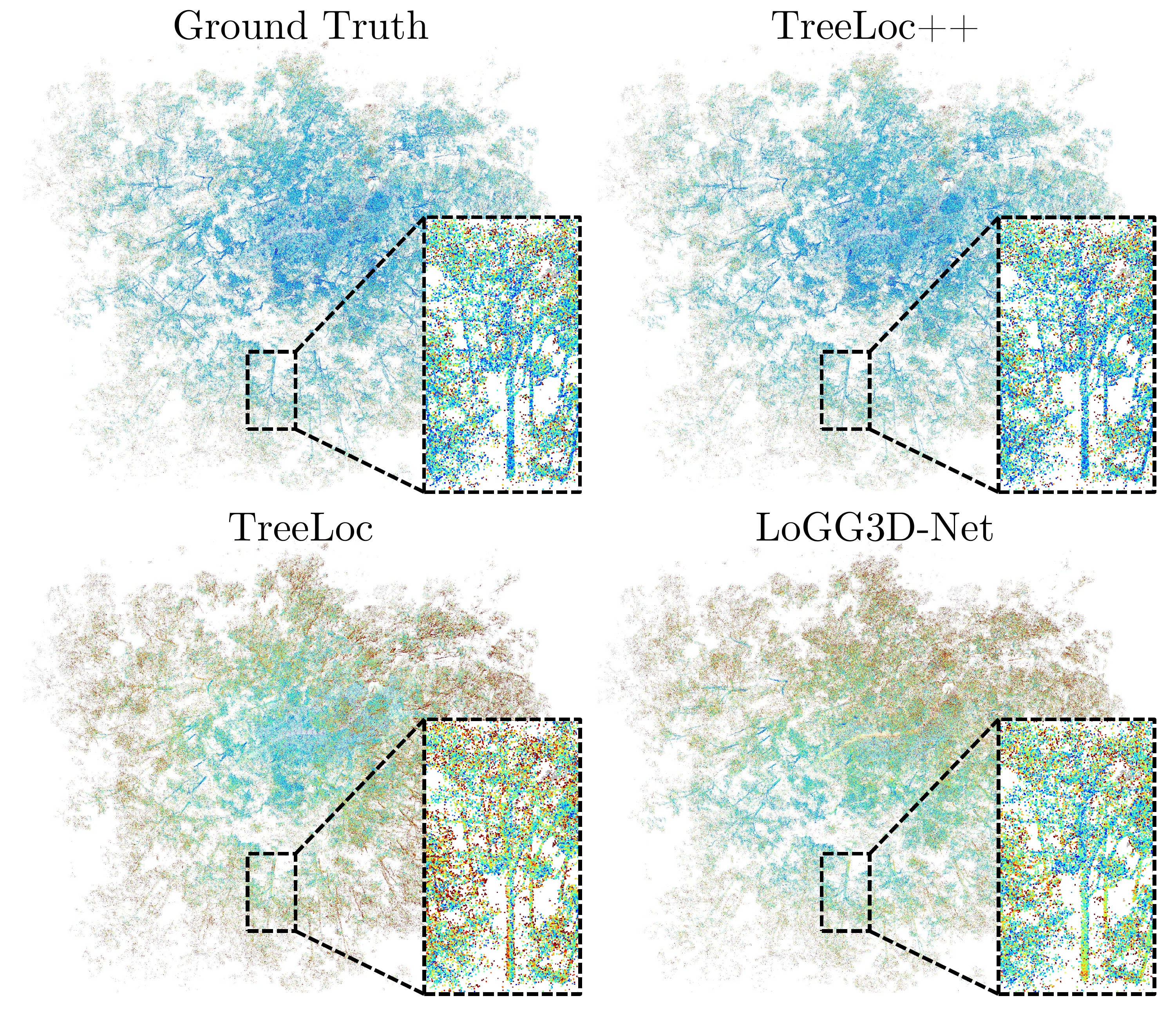}
        \caption{Registration Error in \texttt{Karawatha04}}
        \label{fig:intra_registration_b}
    \end{subfigure}
    \caption{(\textbf{Exp B}) Registration performance on \texttt{Evo} and \texttt{Karawatha04}. (a) Point-to-map nearest-neighbor distance after alignment, where TreeLoc and TreeLoc++ achieve low errors and TreeLoc++ is closest to ground truth. (zoom highlights trunk alignment). (b) In \texttt{Karawatha04}, TreeLoc++ shows blue-dominant patterns consistent with ground truth, while the other methods exhibit red-dominant regions indicating larger residual errors.}
    \label{fig:intra_registration}
    \vspace{-4mm}
\end{figure*}

\input{tab/intra_localization}

\noindent \textbf{Evaluation protocol: }We utilized four sequences from Wild-Places: \texttt{Venman03-04} and \texttt{Karawatha03-04}. These sequences were not used to train any of the learning-based baselines. We followed the same intra-session evaluation protocol as used for the Oxford dataset.

\begin{figure}[!t]
    \centering
        \includegraphics[width=\columnwidth]{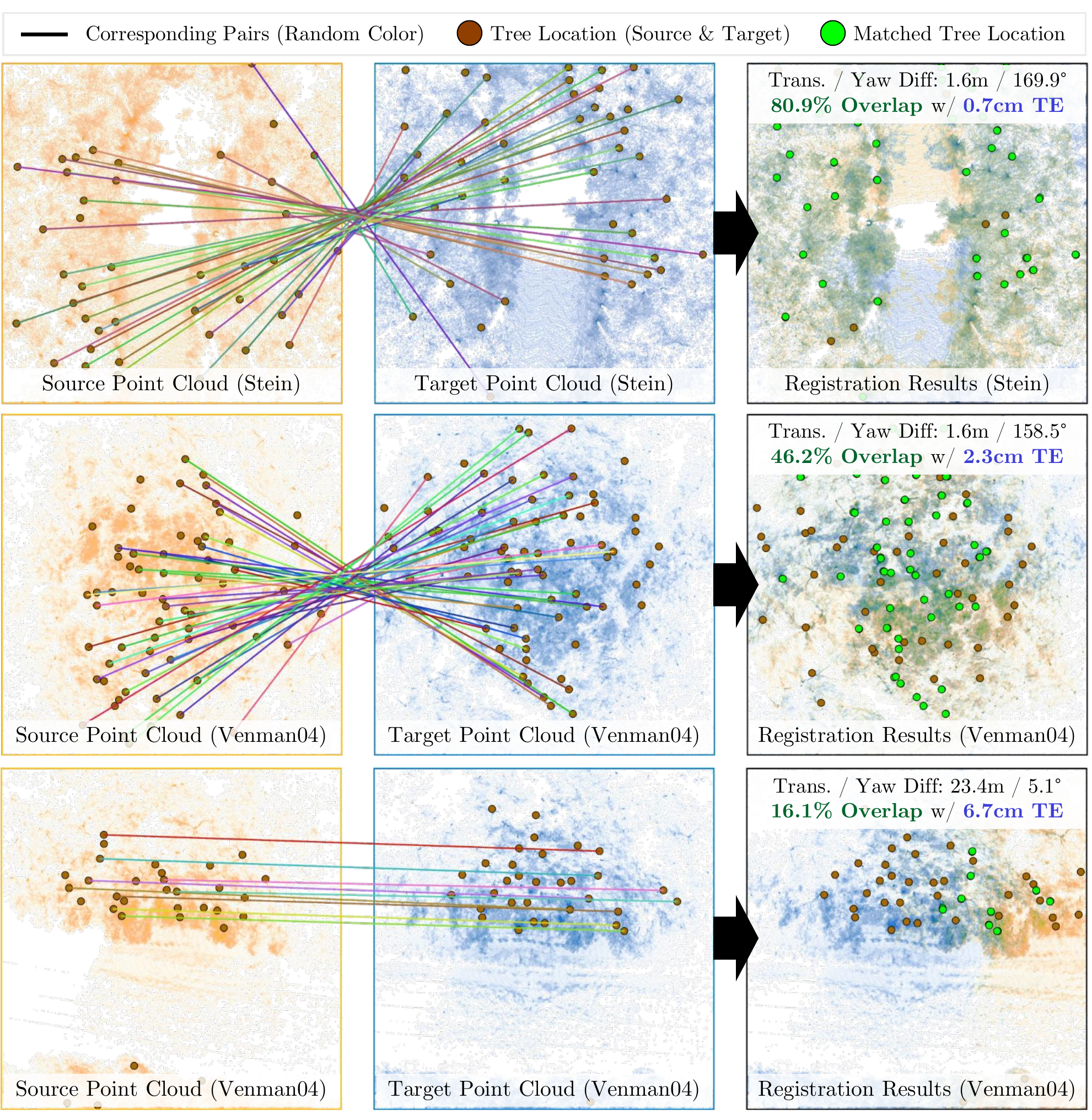} 
    \vspace{-5mm}
    \caption{(\textbf{Exp B}) Each row visualizes tree correspondences after geometric verification. TreeLoc++ establishes matches not only in high-overlap cases (top), but also in moderate (middle) and low-overlap (bottom) scenarios, even beyond the positive threshold, despite complex tree arrangements. This accurate correspondence leads to centimeter-level errors.}
    \label{fig:correspondence}
    \vspace{-3mm}
\end{figure}

\noindent \textbf{Results:}
As summarized in \tabref{tab:intra_wild}, TreeLoc++ achieved the strongest overall performance on Wild-Places.
Despite being less suitable for tree reconstruction, TreeLoc++ forms multiple triangles from common trees for reliable retrieval (\figref{fig:correspondence}), while outlier rejection modules further reduce structural ambiguities that often cause failures in TreeLoc.
By contrast, most hand-crafted baselines degrade more on Wild-Places, as they struggle to distinguish structurally similar locations in large databases, {where repetitive trunk geometry yields similar segment-level structures}.
Learning-based methods such as LoGG3D-Net and BEVPlace++ achieved relatively high recall on \texttt{Karawatha03}, but lagged behind TreeLoc++ on other metrics due to weaker discrimination.

\noindent \textbf{Similarity analysis:}
The similarity matrices in \figref{fig:feature_similarity} explain both the advantage of TreeLoc++ and the increased difficulty of Wild-Places with larger databases.
We visualize similarity matrices with $20$-$80\%$ and $60$-$80\%$ percentile ranges to highlight overall structure and high-confidence matches, respectively.
With larger databases and fewer revisits, true positives become sparse and the separation between true positives and true negatives becomes harder to maintain, making retrieval more challenging.
TreeLoc++ preserves this separation by closely following the ground truth spatial structure, enabling reliable retrieval.
In contrast, TreeLoc shows lower contrast and more noise, increasing false positives, while MapClosure has limited dynamic range, reducing true positives and increasing false negatives. Other global descriptor baselines often assign high similarity to unrelated locations, degrading precision and increasing false positives.

In summary, TreeLoc++ achieves the strongest intra-session place recognition across diverse forests without training and remains robust across scales.

\input{tab/inter-session}

\subsection{Intra-session Metric Localization}
After completing the evaluation of intra-session place recognition on two forest datasets, we turn our attention to another key strength of this work: metric-level localization performance. 
In this section, we assess whether each method can recover the relative poses following place recognition.

\noindent \textbf{Evaluation protocol: }We evaluated metric localization on four sequences with high R@1 to ensure sufficient true-positive retrievals for downstream pose estimation. Similar to SGV \cite{vidanapathirana2023spectral}, we evaluate LoGG3D-Net and MinkLoc3Dv2 using feature-based correspondences and RANSAC registration {to directly estimate the 6-DoF pose after retrieval}.

\noindent \textbf{TreeLoc++:}
As shown in \tabref{tab:wild_places_results}, TreeLoc++ achieved the best overall performance across both 2D and 3D metrics.
While TreeLoc performed well in \texttt{Stein am Rhein}, {its accuracy dropped in cluttered scenes such as \texttt{Karawatha04}.}
This degradation was more pronounced in 3D due to TreeLoc's independent estimation of roll, pitch, and height, which is prone to compounded errors.
In contrast, TreeLoc++ jointly estimates these components, yielding more stable and accurate results.
Moreover, its high R@50 score indicates successful localization even when the top-1 retrieval does not fall within the evaluation threshold.

\noindent \textbf{Baselines:}
BEV-based methods suffered from viewpoint sensitivity and resolution constraints.
BEVPlace++ and RING++ do not estimate roll and pitch, causing BEV projections to degrade under viewpoint changes.
MapClosure attempts ground-plane compensation but fails under uneven terrain or occlusions (\figref{fig:align}), and ORB-based descriptors on BEV images remain insufficiently distinctive.
{Similarly, local- and segment-based methods such as BTC and NSM are affected by partial observations and occlusions in forests. These factors destabilize BTC's local features and NSM's segment centroids, thereby limiting their localization accuracy even when NSM directly localizes against a global map.}

For learning-based methods using 3D point features, performance depended strongly on feature resolution. MinkLoc3Dv2 uses heavily downsampled representations, limiting discriminative power.
In the case of LoGG3D-Net, it retained more spatial detail but still lagged behind TreeLoc++.
Learning-based methods also incurred substantial computational overhead, as they must compute pairwise distances between high-dimensional local features to establish feature-level correspondences for metric localization.

\begin{figure*}[!t]
    \centering
    \includegraphics[width=\linewidth]{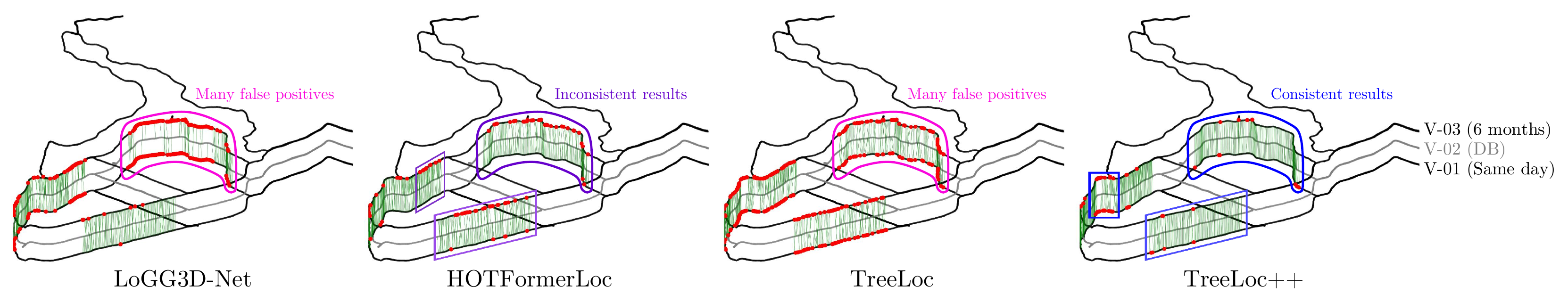}    
    \caption{(\textbf{Exp C}) Inter-session place recognition on \texttt{Venman01-03}. Following the Wild-Places protocol, evaluation is conducted on unseen regions excluding the training area. LoGG3D-Net and TreeLoc produce numerous false positives (red dots) across sessions (pink). HOTFormerLoc performs well for same-day sequences (\texttt{Venman01-02}) but degrades on \texttt{Venman03} with a 6-month gap (purple). TreeLoc++ remains consistent across temporal shifts (blue) and retrieves more true positives, even though its false positive regions also appear consistently across sessions.}

    \label{fig:venman_inter_session}
    \vspace{-4mm}
\end{figure*}

\noindent \textbf{Alignment accuracy:}
To assess alignment accuracy, we evaluated registration on the same set of successful query-match pairs across methods, measuring accuracy via nearest neighbor correspondences within \unit{50}{cm}.
As shown in \figref{fig:intra_registration_a}, TreeLoc++ produced the most accurate registration, closely similar to the ground truth.
TreeLoc and LoGG3D-Net exhibited moderate misalignments, which became more pronounced in cluttered scenes such as \texttt{Karawatha04} (\figref{fig:intra_registration_b}). BTC and MapClosure showed severe failures.
TreeLoc++'s performance stems from robust tree-level correspondences in \figref{fig:correspondence}, which remain reliable under partial overlap and enable centimeter-level accuracy comparable to point-based registration methods, as detailed in \textcolor{blue}{Appendix C}.

\subsection{Inter-session Place Recognition}
\label{sec:experiment_inter}
The previous three experiments focused exclusively on performance within intra-session settings. We now extend our evaluation to inter-session scenarios with larger temporal gaps, analyzing the robustness of the proposed method under more challenging long-term environmental changes.
We aim to determine whether methods maintain consistent performance under seasonal and structural variation.

\noindent \textbf{Evaluation protocol:}
TreeLoc++ was evaluated in an inter-session setting using the local inventory mode on \texttt{Venman01-04} and \texttt{Karawatha01-04}.
Learning-based methods were trained on Wild-Places, following the official protocol, which restricts evaluation to predefined regions, as shown in \figref{fig:venman_inter_session}.
To quantitatively assess robustness across sessions, we introduce a Stability Ratio, $S=\log\!\left(\mu/(\sigma+\epsilon)\right)$, where $\mu$ and $\sigma$ denote the mean and standard deviation of the performance metric over the 12 inter-session pairs, and $\epsilon$ is a small constant for numerical stability.
Unlike raw variance, this ratio favors methods that are both high-performing and consistent.

\begin{figure}[t]
    \centering
    \begin{subfigure}[t]{\columnwidth}
        \centering
        \includegraphics[width=\columnwidth]{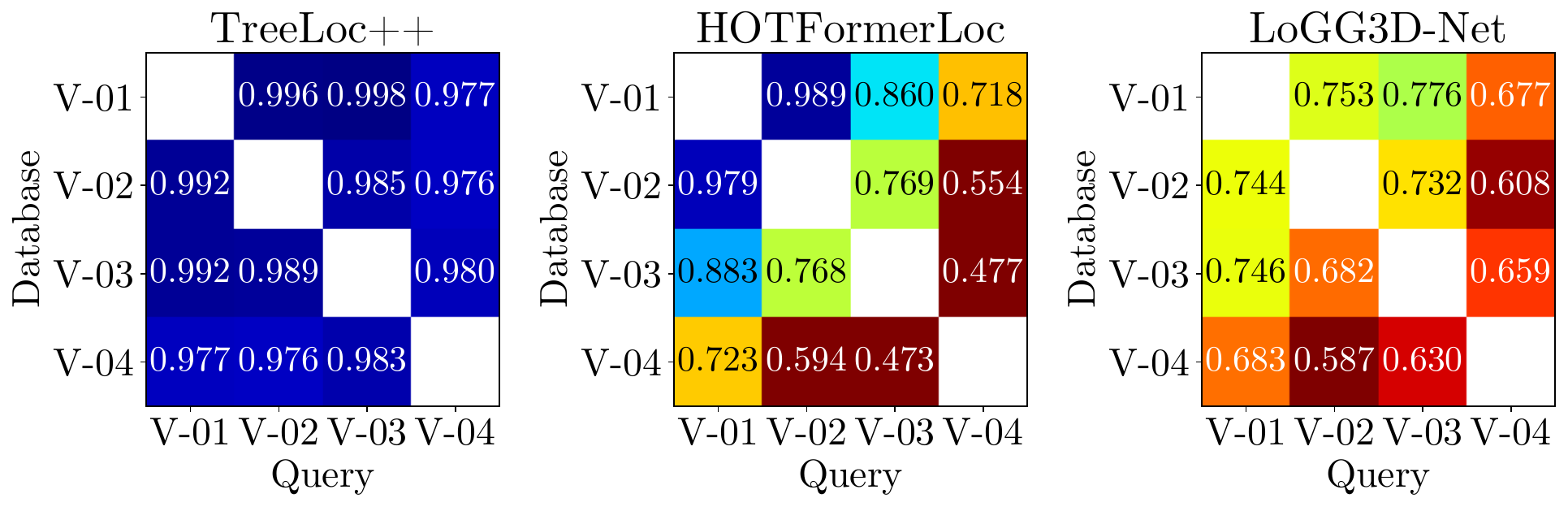}
        \vspace{-6mm}
        \caption{Accuracy in \texttt{Venman}}
        \vspace{-2mm}
        \label{fig:accuracy_venman}
    \end{subfigure}
    \begin{subfigure}[t]{\columnwidth}
        \centering
        \includegraphics[width=\columnwidth]{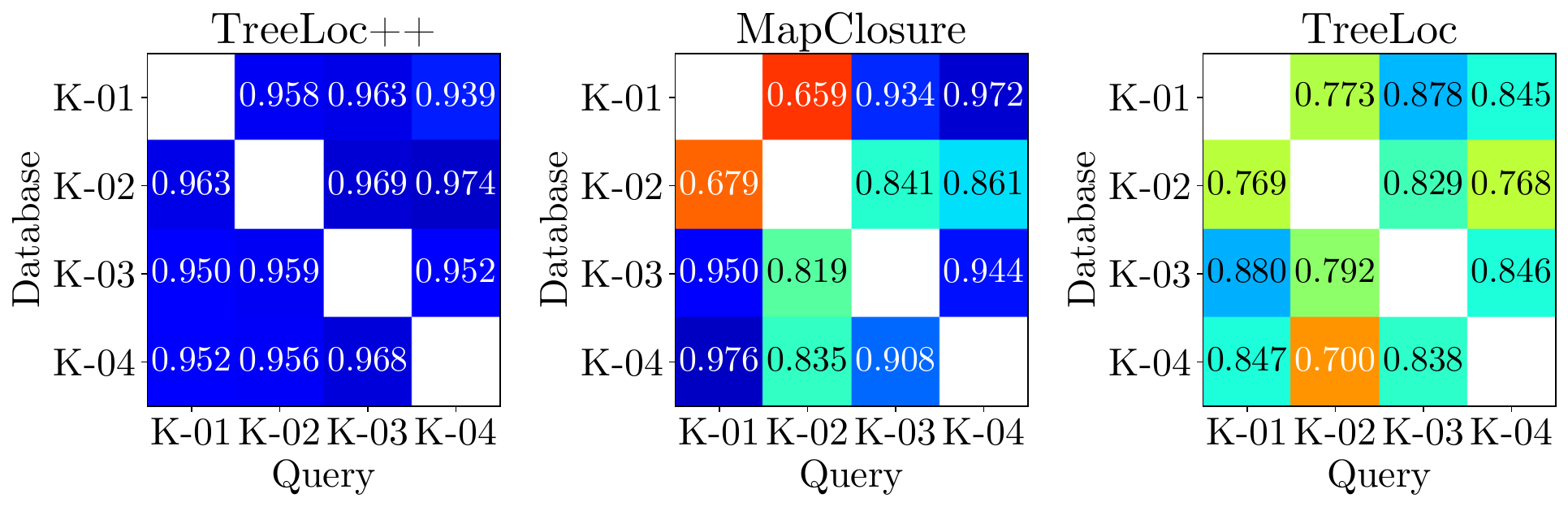}
        \vspace{-6mm}
        \caption{Accuracy in \texttt{Karawatha}}
        \label{fig:accuracy_karawatha}
    \end{subfigure}
    \vspace{-3mm}
    \caption{{(\textbf{Exp C}) Accuracy matrix for inter-session place recognition. Each cell represents the fixed-threshold accuracy for one query-database session pair. Blue tones indicate higher accuracy, while red tones indicate lower accuracy. A desirable pattern is a uniformly blue matrix, which indicates that a method maintains consistently high performance across different session pairs without requiring pair-specific threshold tuning. TreeLoc++ shows the most reliable performance across all session pairs.}}
    \label{fig:accuracy_diagram}
    \vspace{-3mm}
\end{figure}

\noindent \textbf{TreeLoc++:}
As reported in \tabref{tab:inter_session}, TreeLoc++ achieved the highest mean performance and Stability Ratio on most metrics.
Even when learning-based models were trained on the same dataset, TreeLoc++ surpassed them in both mean performance and stability, ranking second only in R@1 on \texttt{Venman}.
On \texttt{Karawatha}, recall decreased in open areas with fewer distinctive vertical structures; nevertheless, TreeLoc++ achieved Stability Ratio comparable to or higher than several learning-based methods.

\noindent \textbf{Baselines:}
Most hand-crafted methods degraded in the inter-session setting, indicating limited robustness to temporal variation.
While TreeLoc outperformed other hand-crafted baselines, it still lagged behind learning-based methods. {NSM also degraded substantially under seasonal variation, as it matches individual segment descriptors without modeling inter-segment connectivity. This makes it more sensitive to appearance changes in similar trunk segments, so that even when the mean performance remains relatively high, the Stability Ratio becomes noticeably lower.}
Among learning-based approaches, LoGG3D-Net and HOTFormerLoc achieved the highest mean performance; however, LoGG3D-Net degraded on \texttt{Venman}, and HOTFormerLoc's low Stability Ratios suggest inconsistent performance across sessions.
As shown in \figref{fig:venman_inter_session}, LoGG3D-Net produced frequent mismatches, and HOTFormerLoc deteriorated with increasing temporal gaps. Furthermore, while these methods still retrieve true positives via minimum descriptor distance, accompanying false positives necessitate an appropriate threshold to filter out incorrect matches.



\noindent \textbf{Fixed-threshold transfer:}
Consequently, we assessed inter-session robustness using fixed-threshold accuracy analysis in \figref{fig:accuracy_diagram}.
To reflect realistic deployment where thresholds cannot be retuned for each session pair, we fixed a single operating threshold per method by maximizing F1 score on \texttt{01-02} pair, and applied it unchanged to all remaining pairs.

As shown in \figref{fig:accuracy_venman} and \figref{fig:accuracy_karawatha}, TreeLoc++ was the only method that consistently achieved high accuracy across all pairs.
In contrast, baseline methods exhibited large fluctuations.
HOTFormerLoc and LoGG3D-Net degraded substantially on \texttt{Venman04} due to pronounced environmental changes.
Conversely, MapClosure and TreeLoc performed poorly on \texttt{Karawatha01-02} because low thresholds admitted many false positives, but improved in later sessions where larger appearance differences made discrimination easier.
Overall, these results suggest that the similarity scores used by the baselines are brittle under fixed thresholds.
In contrast, the proposed overlap score $\mathcal{O}(Q,C)$ maintains stable separation between true and false matches, demonstrating robust performance without session-specific tuning.


\subsection{Multi-session Metric Localization with Global DFIs}
\label{sec:relocalization}
This experiment assesses the feasibility of using multiple global forest inventories to localize incoming trajectories using local observations in a scalable and robust way.

\begin{figure}[t]
    \centering
    \includegraphics[width=\columnwidth]{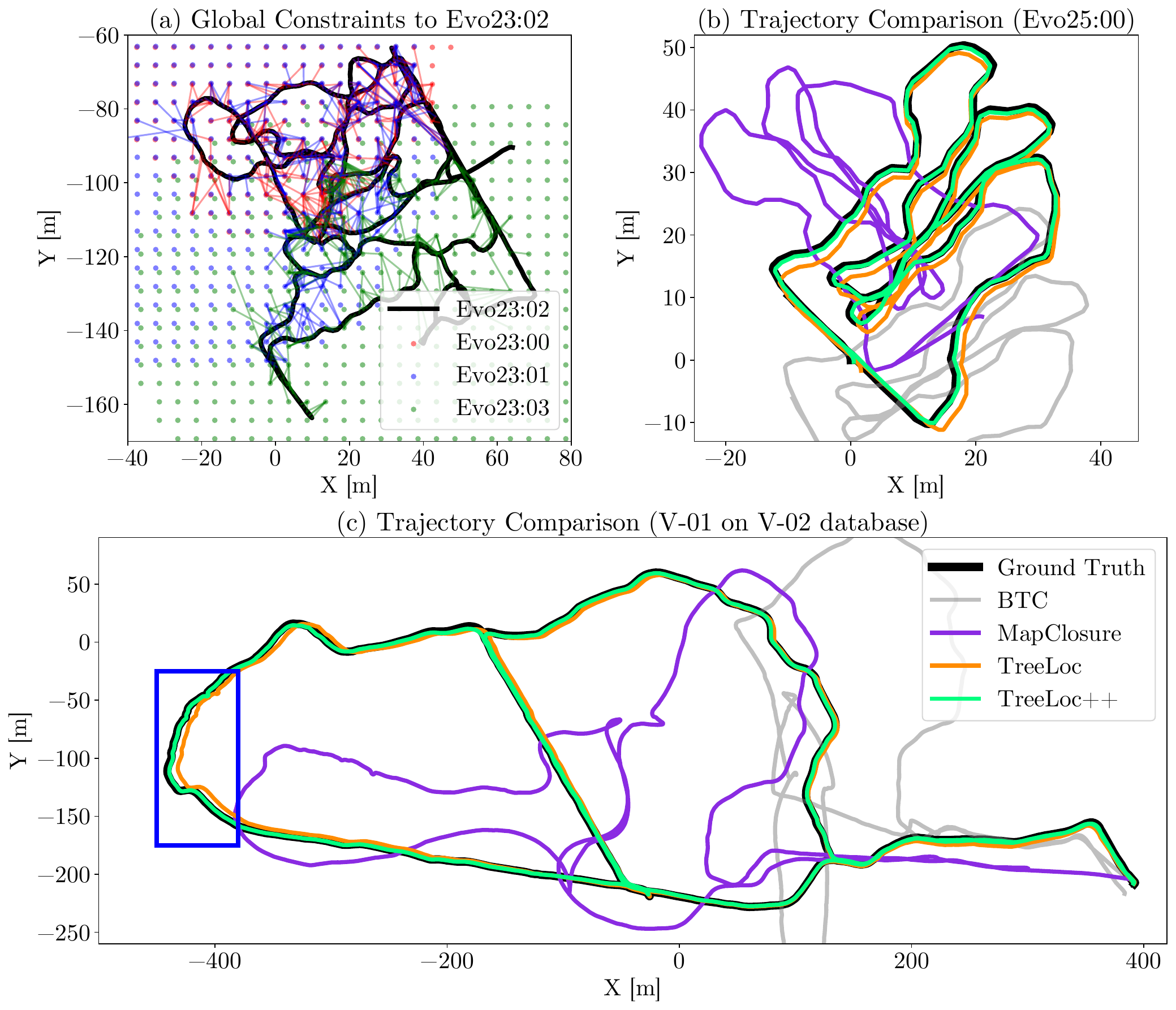}    
    \vspace{-4mm}
    \caption{(\textbf{Exp D}) Relocalization against a global forest inventory. (a) A global descriptor database is built by querying the global inventory at uniformly sampled poses on a 5 m grid, efficiently covering the full map extent. (b-c) Using relocalization constraints to align trajectories to the global frame, TreeLoc++ produces trajectories closest to the ground truth, as highlighted in the blue box. {In particular, the blue box in (c) highlights the close agreement between the aligned trajectory and the ground truth.}} 

    \label{fig:local_to_global}
    \vspace{-4mm}
\end{figure}

\noindent\textbf{Evaluation protocol:}
We evaluated relocalization against a pre-built global forest inventory that serves as a persistent reference map.
Query frames were processed in local inventory mode, while the database consisted of a single global inventory.
To cover the spatial extent of the reference map, we constructed database entries at \unit{5}{m} intervals in the horizontal plane as shown in \figref{fig:local_to_global}a.
At each sampled location, a local inventory was formed by extracting nearby trees, and the corresponding TreeLoc++ descriptor was stored in the global descriptor database.
Experiments were conducted on \texttt{Evo23:00-03}, \texttt{Evo25:00}, \texttt{02}, \texttt{03}, \texttt{05}, and \texttt{Venman01-02}.
For each sequence, initial odometry was obtained using GeoTransformer \cite{qin2023geotransformer}.
We first generated intra-session constraints using local inventories, and then incrementally added inter-session constraints by matching queries to the global descriptor database.

Constraint selection differed by method.
Baseline constraints were selected either by (i) ground truth-validated success (SR) or (ii) a confidence threshold chosen to maximize F1 from intra-session localization.
For TreeLoc and TreeLoc++, we used a fixed threshold, $\mathcal{O}(Q,C) > 0.2$.
To ensure a sufficient number of constraints under the SR-based setting, we relaxed the true-positive distance threshold to \unit{10}{m}.
All selected constraints were optimized using GTSAM \cite{dellaert2012factor}.
Trajectory accuracy was evaluated with the Evo toolkit \cite{grupp2017evo} using ATE and ARE.
Except for intra-session loop evaluations, we report results without applying SE(3) alignment during Evo evaluation. We additionally report the number of constraints produced by each method, as well as the storage required for descriptors and forest inventories.

\input{tab/local_to_global}

\noindent\textbf{Relocalization accuracy:}
As reported in \tabref{tab:local_to_global}, TreeLoc++ achieved among the lowest trajectory errors in nearly all settings by correcting raw odometry with accurate and reliable constraints.
Even when baselines were restricted to SR-validated constraints, TreeLoc++ typically attained lower or comparable error by leveraging a larger set of accurate constraints.
{When baselines used F1-based thresholds, the number of selected constraints increased due to false positives, substantially degrading trajectory accuracy.}
Overall, these results indicate that threshold-based constraint selection in existing methods is brittle in practical global relocalization, consistent with observations in earlier sections.

\figref{fig:local_to_global}a illustrates \texttt{Evo23:02}, where TreeLoc++ aligned the trajectory to the global reference frame by incorporating constraints from three inter-session sequences. As shown in \figref{fig:local_to_global}b-c, the optimized trajectories closely match the ground truth, benefiting from consistent and accurate constraints. On \texttt{Venman}, which contains 6,918 database entries, TreeLoc++ demonstrated its reliability by efficiently querying the large-scale global database and aligning all trajectories into a common world frame.

\noindent\textbf{Storage and scalability:}
TreeLoc++ also offers clear storage advantages.
Compared to other descriptor-based methods, it requires substantially less storage due to its compact representation.
Although the \texttt{Loop} configuration stores a local inventory per pose, the total storage remains on the megabyte scale, and the global inventory is even smaller, as shown in \tabref{tab:local_to_global}. 
These results show that TreeLoc++ supports accurate and scalable global trajectory estimation with minimal storage overhead.

\begin{figure*}[t]
    \centering
    \includegraphics[width=\textwidth]{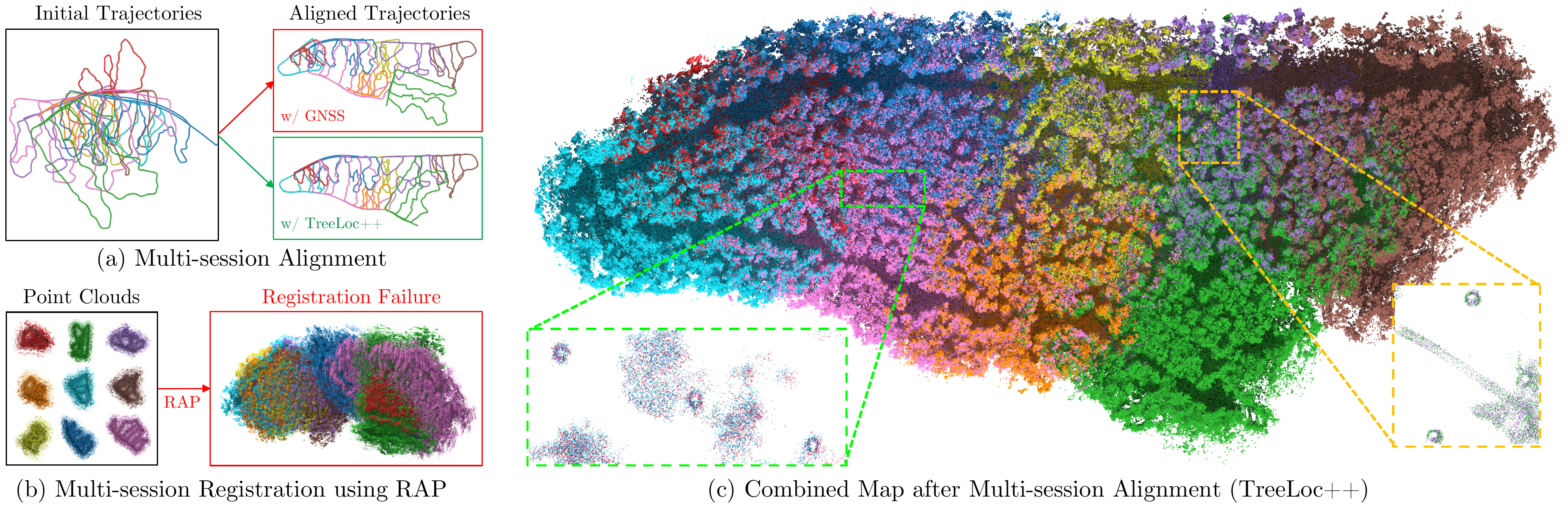}    
    \vspace{-4mm}
    \caption{(\textbf{Exp E}) Multi-session alignment results. (a) Each session trajectory is expressed in its own local frame, with all trajectories starting at the origin; GNSS-based alignment remains inconsistent across sessions, with clear misalignment in the green trajectory, whereas TreeLoc++ yields more accurate multi-session alignment. (b) Point cloud-based multi-session registration fails in repetitive forest environments due to ambiguous geometry. (c) Combined map generated using TreeLoc++ aligned poses, where the zoomed views in the {orange and green boxes} highlight the consistent alignment of tree trunks and fallen trees observed across multiple sessions, demonstrating accurate alignment.}

    \label{fig:evo2025}
    \vspace{-5mm}
\end{figure*}

\subsection{Multi-session Registration with Global DFIs}
In this section, we seek to address the following question: whether global forest inventories alone are sufficient for accurate map-to-map registration across sessions, in the absence of raw point clouds or GNSS.

\noindent\textbf{Evaluation protocol:}
This section evaluates map-to-map registration using only global forest inventories, without relying on raw point clouds.
For each session, a single global forest inventory served as the map representation.
System poses were sampled along each session, and at each sampled pose, nearby trees were extracted from the global inventory to form pose-centric inventories.
{TreeLoc++ descriptors were computed for these samples and matched across all session pairs to establish inter-session constraints, using a fixed overlap threshold of $\mathcal{O}(Q,C) > 0.2$; the rationale for this choice is discussed in \textcolor{blue}{Appendix F.2.}}

\noindent\textbf{TreeLoc++:}
\figref{fig:evo2025} illustrates the initial and aligned trajectories of \texttt{Evo25:00-08}, along with a combined point-cloud visualization obtained by transforming scans using the aligned poses.
Thanks to TreeLoc++'s high localization accuracy, trunk structures align consistently across sessions, even though raw point clouds are not used for constraint generation or registration.

\input{tab/ablation_rotation}
\begin{figure}[t]
    \centering
    \includegraphics[clip=true, trim={200 60 200 60},width=.9\columnwidth]{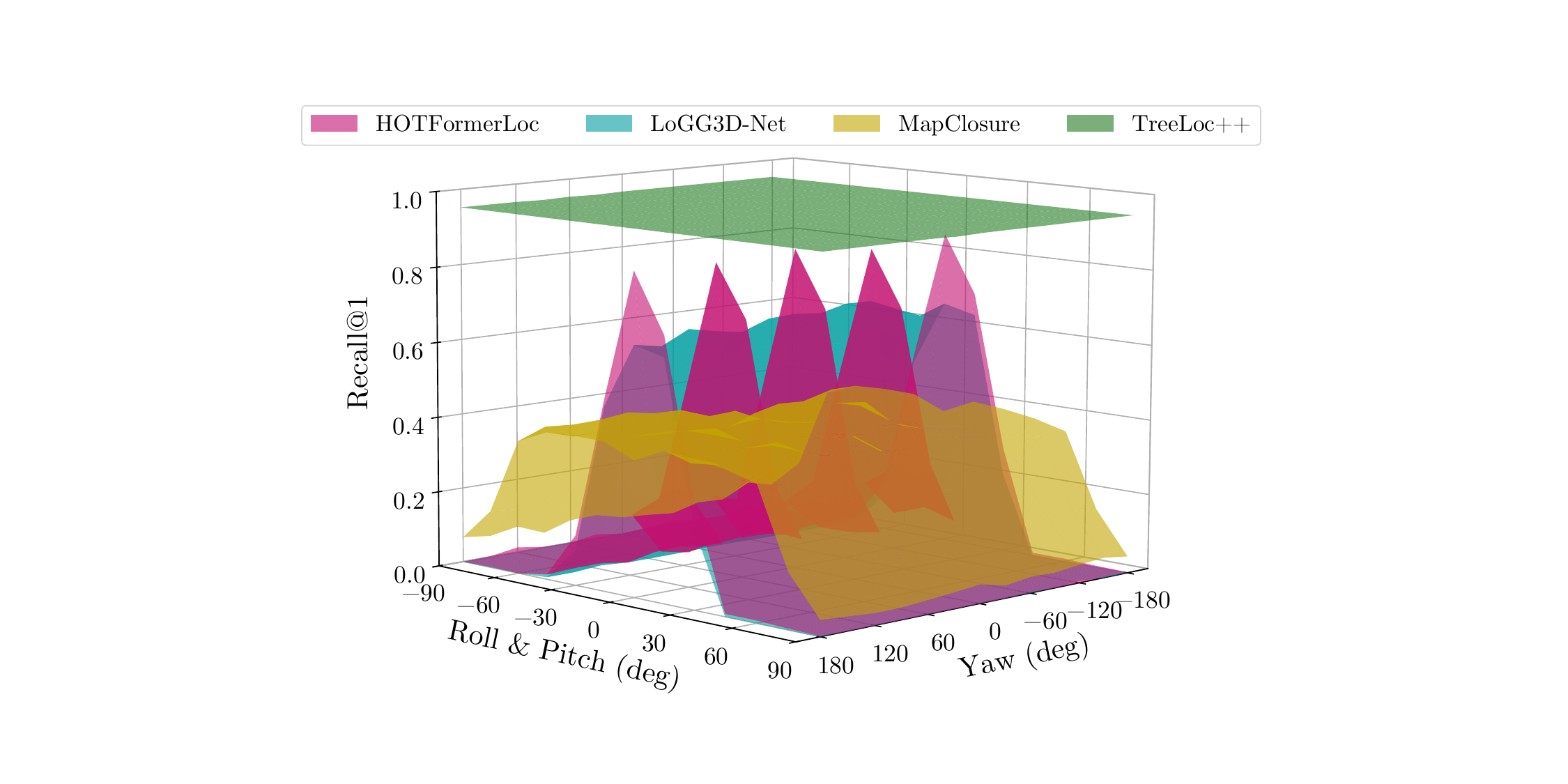}
    \caption{(\textbf{Exp F-1}) R@1 performance under synthetic orientation perturbations. {TreeLoc++ maintains high recall across the full rotation range, whereas the baselines degrade under roll-pitch perturbations.}}

    \label{fig:rotation_invariance}
    \vspace{-4mm}
\end{figure}

\noindent\textbf{GNSS and point-cloud baselines:}
For comparison, we performed multi-session alignment using GNSS by estimating a rigid transform between each session's local trajectory and its GNSS-referenced trajectory.
As shown for \texttt{Evo25:08} (green), this produced noticeable misalignments and inconsistent trajectories across sessions, indicating that GNSS provides only coarse alignment and lacks the precision needed for stem-level consistency in forests.
We further compared against RAP \cite{pan2025register}, a point cloud-based registration baseline, as shown in \figref{fig:evo2025}b.
In the unstructured and repetitive forest geometry, RAP failed to converge to a correct alignment, resulting in substantial registration errors.

\noindent\textbf{Long-term scalability:}
Finally, \figref{fig:main} shows large-scale alignment across all 15 sessions from \texttt{Evo23:00-05} and \texttt{Evo25:00-08}.
Despite a two-year gap and different LiDAR FoVs (\(104^\circ\) in 2025 vs.\ \(31^\circ\) in 2023), TreeLoc++ aligned all sessions using only the global forest inventory, with a total map size of \unit{250}{KB}.
These results highlight the robustness of TreeLoc++ and demonstrate that lightweight, geometry-based inventories support scalable multi-session registration, enabling consistent inventory maintenance and long-term ecological monitoring as detailed in \textcolor{blue}{Appendix D}.


\subsection{Robustness and Efficiency Analysis}
\subsubsection{Viewpoint Robustness}
\noindent \textbf{Evaluation Protocol:}
We evaluated the rotation invariance of TreeLoc++ using \texttt{Venman02} as the database and \texttt{Venman03} as the query.
To simulate viewpoint variation, we applied synthetic rotations to the database point clouds and DFIs.
Yaw was varied from $-180^\circ$ to $180^\circ$ in $30^\circ$ steps, and roll and pitch were jointly varied from $-90^\circ$ to $90^\circ$ in $15^\circ$ steps.
For each augmented database, we measured R@1 and quantified consistency via the Stability Ratio ($S$).

\noindent \textbf{Results:}
As shown in \tabref{tab:invariance_analysis} and \figref{fig:rotation_invariance}, all methods except TreeLoc++ failed under roll-pitch perturbations. {HOTFormerLoc exhibited cyclic local peaks under yaw rotations at 90$^{\circ}$ intervals.} LoGG3D-Net and MapClosure showed partial robustness to small tilts, but their accuracy deteriorated as perturbations increased. In contrast, TreeLoc++ maintained high R@1 across all rotations, enabled by yaw-invariant TDH/PDH and 2D triangles, and axis-based alignment that stabilizes projections under tilt. These results confirm TreeLoc++'s robustness to severe viewpoint variations.

\subsubsection{Runtime Efficiency}
\noindent \textbf{Evaluation Protocol:} 
We conducted two experiments to analyze runtime. First, we reported the total inter-session localization time, including place recognition and pose estimation (geometric verification). To compare localization efficiency across methods, we followed the large-scale setup from \secref{sec:relocalization}, using 6,918 database entries and 1,451 queries.
Second, to analyze TreeLoc++ in detail, we used \texttt{Karawatha03} to measure the runtime of descriptor generation, place recognition, and pose estimation separately. For a fair comparison, we evaluated only methods with publicly available C++ implementations on an 11th Gen Intel Core i7-11700 @ \unit{2.50}{GHz} and an NVIDIA GeForce RTX 3080 GPU.

\input{tab/time_consumption}
\begin{figure}[!t]
    \centering
    \includegraphics[width=\columnwidth, trim=0 0 0 0, clip]{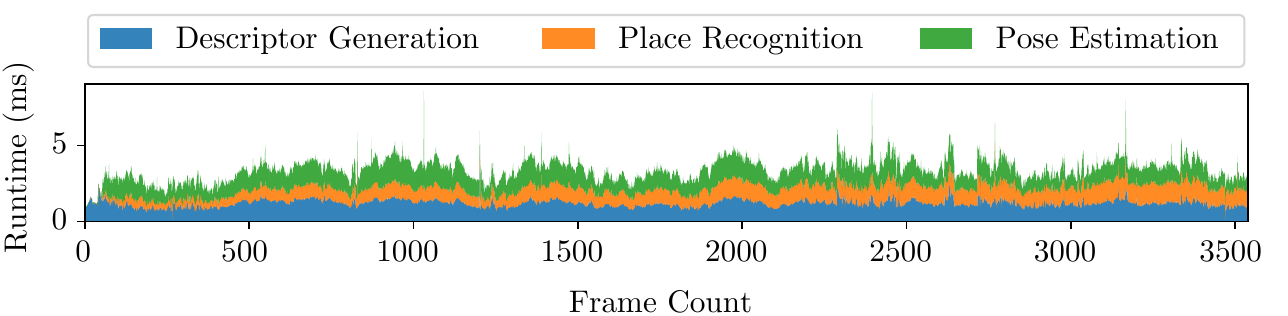}
    \caption{(\textbf{Exp F-2}) Runtime breakdown for intra-session localization in \texttt{K-03}. {Colored bands indicate descriptor generation, place recognition, and pose estimation time for each query frame. TreeLoc++ maintains consistently low per-query latency throughout the sequence, with descriptor generation remaining below {1} {ms}.}}
    \label{fig:runtime}
    \vspace{-5mm}
\end{figure}

\noindent \textbf{Results:}
As shown in \tabref{tab:efficiency_runtime}, TreeLoc++ achieved the lowest runtime among all methods. BTC incurred high runtime due to costly triangle-wise comparisons during place recognition, while MapClosure was faster by leveraging lightweight ORB-based image matching. TreeLoc++ outperformed both by avoiding per-candidate geometric checks and limiting pose estimation to a small set of top-ranked matches.
As shown in \figref{fig:runtime}, TreeLoc++ also benefits from fast descriptor generation via DFI, requiring less than \unit{1}{ms} per query. This makes on-demand descriptor computation nearly as efficient as preloading, avoiding additional I/O cost and storage.
Moreover, TreeLoc++ maintains low runtime as the database grows, due to efficient filtering using TDH and PDH. Although histogram distance computations grow with the database size, the increase remains small thanks to the low-dimensional histogram.
These design choices enable TreeLoc++ to support fast localization while conserving compute for downstream tasks like mapping and planning.

\input{tab/ablation_coarse}
\begin{figure}[t]
    \centering
    \includegraphics[width=\columnwidth]{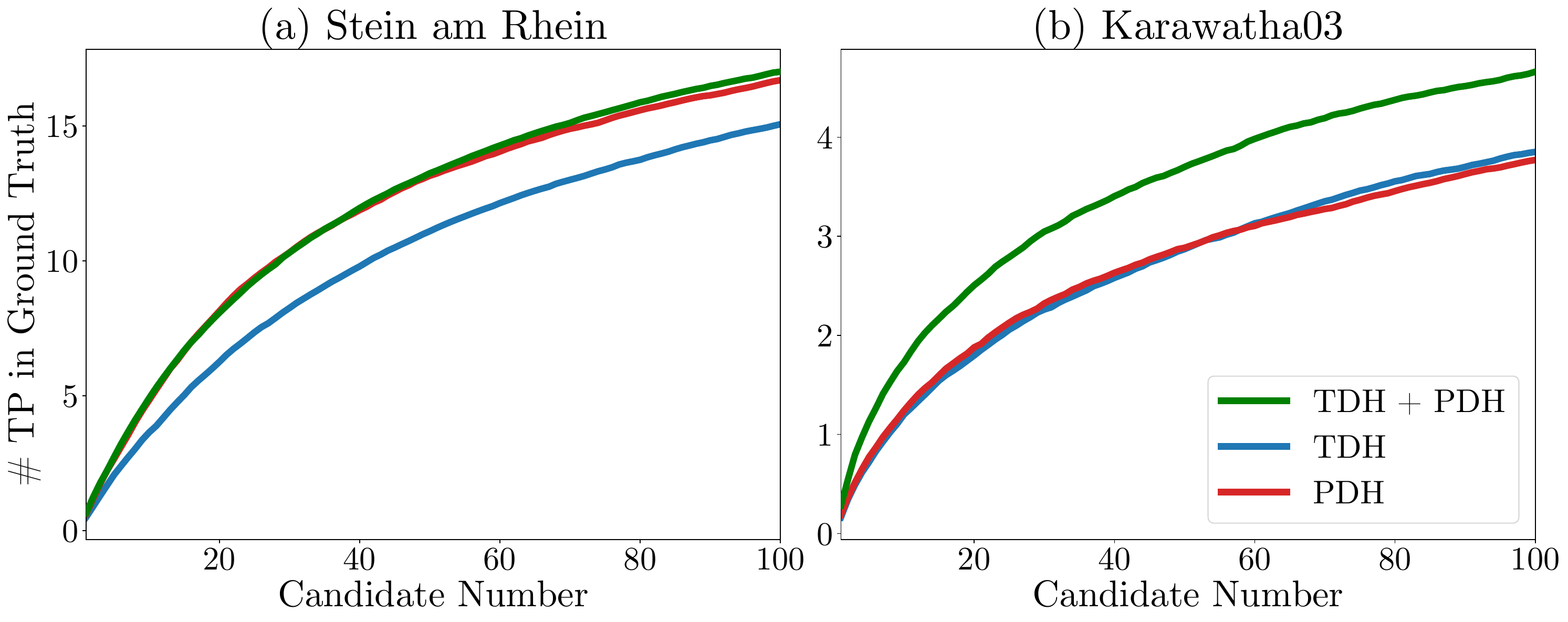}    
    \caption{(\textbf{Exp G-1}) Number of true positives versus candidate count {in coarse retrieval. In \texttt{Stein am Rhein}, TDH+PDH stays only slightly above PDH because PDH already retrieves most true positives in this compact scene. In \texttt{Karawatha03}, the gap becomes larger, showing that TDH and PDH provide complementary cues in a larger and more structurally ambiguous environment.}}

    \label{fig:candidate_pdh}
    \vspace{-5mm}
\end{figure}

\subsection{Ablation Studies}
The contribution of each TreeLoc++ component was analyzed through an ablation study. Experiments were conducted on two contrasting environments: the small-scale \texttt{Stein am Rhein} and the large-scale \texttt{Karawatha03}. {Complementary analyses of parameter sensitivity, robustness to DFI quality, and the statistical comparison between TreeLoc and TreeLoc++ are provided in \textcolor{blue}{Appendix F} and \textcolor{blue}{G}.}

\subsubsection{Ablation on Coarse Retrieval with TDH and PDH}
\noindent \textbf{Evaluation Protocol: }
We assessed the impact of TDH and PDH by incrementally enabling each component and measuring candidate retrieval performance.
We report R@1 and R@50, the average number of true positives (\# TP) in the retrieved candidate set, and the false negative rate (FNR), defined as the fraction of queries for which the ground truth match is absent from the retrieved candidates.

\noindent \textbf{Results:}
As shown in \tabref{tab:descriptor_ablation}, \texttt{Stein am Rhein} exhibited only minor changes in recall, since its limited spatial extent already yields many true positives in coarse retrieval. {Accordingly, PDH alone already recovers most true positives, and combining TDH and PDH provides only a marginal additional benefit, slightly increasing the number of true positives while reducing the FNR. In this compact setting, PDH also retrieves more true positives than TDH because pairwise-distance statistics are already stable thanks to the sufficient number of observed trees.} On the larger \texttt{Karawatha03}, this combination substantially improved both recall and FNR. {In this larger and more structurally ambiguous scene, PDH alone can be less distinctive because different locations may share similar pairwise-distance distributions, while TDH alone remains sensitive to reference-centered layout variations caused by missing trees or slight translation shifts. Their combination therefore provides complementary cues from pairwise geometric stability in PDH and location-centered distribution information in TDH, yielding more true positives, lower FNR, and better place recognition performance.} {\figref{fig:candidate_pdh} also reflects this difference: in \texttt{Stein am Rhein}, the TDH+PDH curve stays only slightly above PDH because PDH already retrieves most true positives, whereas in \texttt{Karawatha03}, the gap becomes larger because the combined descriptor better suppresses structurally ambiguous candidates.}
When using only triangle descriptors without histograms, geometrically similar but distant triangles frequently collide in the hash space, causing false scene matches.
In contrast, combining histograms stabilizes retrieval by jointly encoding inter-tree relationships and location-centered distributions, providing complementary cues for robust candidate selection.

\input{tab/ablation}

\begin{figure}[t]
    \centering
    \begin{subfigure}[t]{\columnwidth}
        \centering
        \includegraphics[width=\columnwidth]{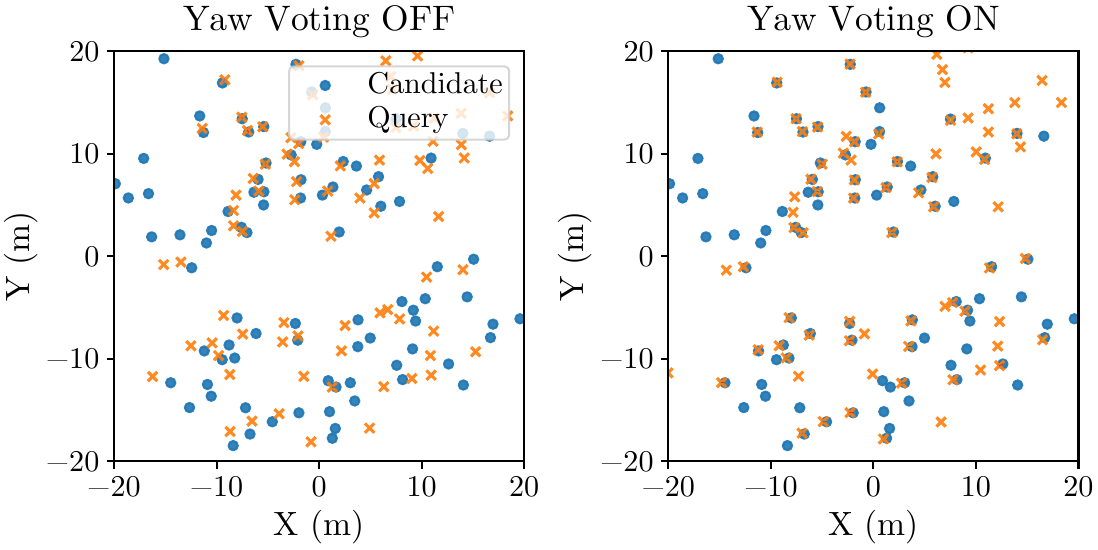}
        \vspace{-3mm}
    \end{subfigure}
    \begin{subfigure}[t]{\columnwidth}
        \centering
        \includegraphics[width=\columnwidth]{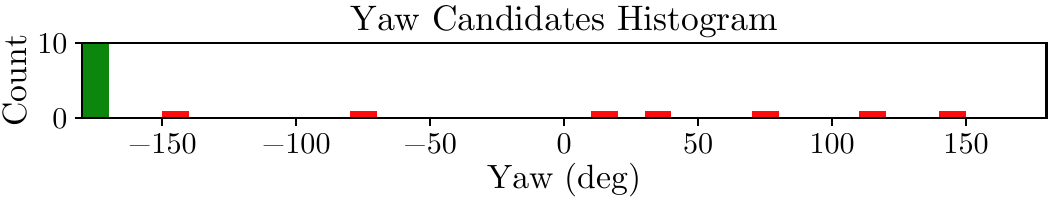}
        \caption{Effect of Yaw Voting in \texttt{Karawatha03}}
        \label{fig:ablation_yaw}
    \end{subfigure}
    \begin{subfigure}[t]{\columnwidth}
        \centering
        \includegraphics[width=\columnwidth]{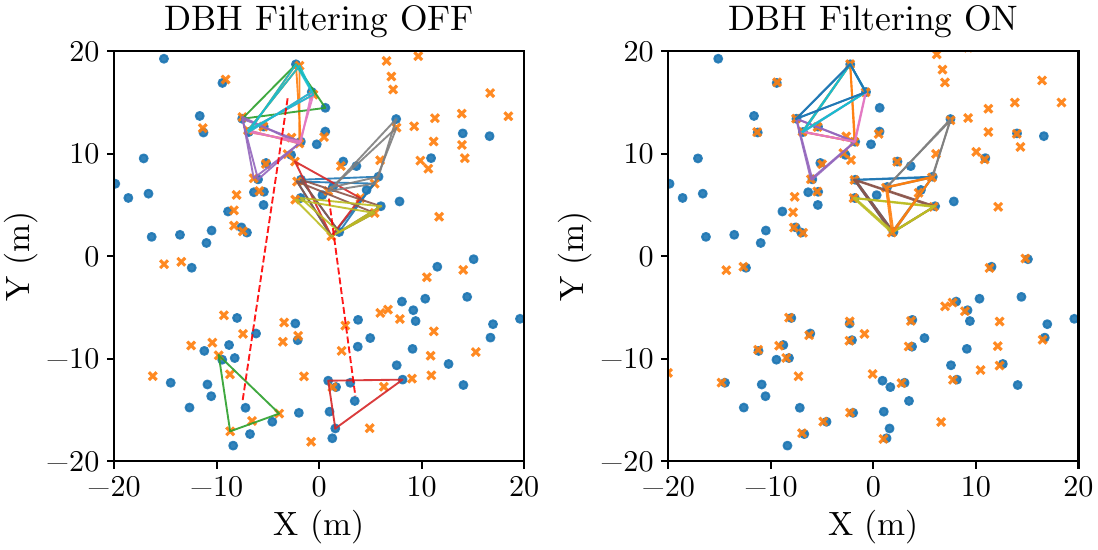}
        \caption{Effect of DBH Filtering in \texttt{Karawatha03}}
        \label{fig:ablation_dbh}
    \end{subfigure}
    \vspace{-2mm}
\caption{(\textbf{Exp G-2}) Ablation of yaw voting and DBH filtering. (a) Registration with yaw voting disabled and enabled, where a histogram built from triangle matches suppresses orientation outliers by selecting the dominant yaw, improving query-candidate alignment. (b) 10 triangle matches with DBH filtering disabled and enabled; consistent colors indicate corresponding pairs. Without DBH filtering, outlier pairs (red dashed) remain and degrade pose estimation, while DBH filtering removes them.} 
    \label{fig:ablation_outlier}
    \vspace{-3mm}
\end{figure}

\subsubsection{Ablation on Candidate Re-ranking and Outlier Rejection}
\noindent \textbf{Evaluation Protocol:}
We evaluated three refinement strategies: the translation penalty, yaw voting, and DBH filtering.
Since these components influence both place recognition and metric localization, we report R@1, AUC, R@50, and SR.

\noindent \textbf{Results:}
As shown in \tabref{tab:refinement_ablation_updated}, enabling the translation penalty improved R@1, AUC, and SR, while R@50 changed only marginally.
By promoting spatially adjacent candidates, it increased R@1, and this led to more accurate overlap scoring, resulting in higher AUC.
The number of correctly retrieved true positives also increased, which raised the SR.
Yaw voting and DBH filtering further improved performance by suppressing outlier triangle correspondences, leading to more accurate scoring and correspondence selection.
These gains were more pronounced on \texttt{Karawatha03}, where the large inventory generated a substantial number of triangles, making the matching stage more prone to outliers.
\figref{fig:ablation_yaw} illustrates that yaw voting enables more reliable pose estimation on \texttt{Karawatha03}; despite the presence of outlier correspondences, it selects a consistent inlier set around the dominant yaw, improving query-candidate alignment.
DBH filtering further improves performance when applied after yaw voting by removing residual outliers that yaw voting alone cannot suppress. As shown in \figref{fig:ablation_dbh}, these outliers otherwise lead to inaccurate pose estimation, whereas DBH filtering yields more consistent correspondences.

\subsubsection{Ablation on TreeLoc++ Extensions over TreeLoc}
\noindent \textbf{Evaluation Protocol:}
We evaluated three extensions in TreeLoc++ over TreeLoc: reusing preceding inventories to increase the number of trees, planar alignment via IRLS, and vertical correction for full 6-DoF pose refinement.
Their impact was quantified using R@1, R@50, ATE, and ARE.

\input{tab/ablation_others}

\noindent \textbf{Results:}
As shown in \tabref{tab:system_ablation}, reusing preceding windows had little effect on \texttt{Stein am Rhein}, where most frames already contain sufficient tree observations.
In contrast, in open or sparse areas of \texttt{Karawatha03}, this reuse increased the number of matched triangles, improving recall.
IRLS refinement was particularly effective in occluded or sparsely observed scenes.
When tree centers were inconsistently estimated, IRLS stabilized the 2D alignment around the initial estimate, producing more consistent correspondences and improved recall.
While these two strategies mainly improved retrieval, their impact on ATE and ARE was limited because the pose refinement stage was unchanged.
In contrast, vertical correction substantially reduced ATE and ARE by jointly optimizing roll, pitch, and height, enabling accurate 6-DoF localization without degrading recall.

%% file: tab/dataset.tex
\begin{table}[t]
\vspace{-1mm}
\centering
\caption{Summary of datasets used for evaluation.}
\label{tab:dataset_summary}
\resizebox{\columnwidth}{!}{%
\renewcommand{\arraystretch}{1.2}
\begin{tabular}{ll|c|>{\raggedright\arraybackslash}p{0.5\columnwidth}}
\toprule
\multicolumn{2}{c|}{\textbf{Dataset / Sequence}} & \textbf{Length} & \textbf{Environment Characteristics} \\
\midrule

\multirow{10}{*}{\rotatebox{90}{Oxford Forest Place Recognition}}
& \multirow{2}{*}{\texttt{Evo:Single}} & \multirow{2}{*}{\unit{1.53}{km}} & Tall mixed-species forest, \\
& & & moderate density \\ \cmidrule(lr){2-4}

& \texttt{Stein am Rhein} & \unit{0.70}{km} & Sparse coniferous trees, flat terrain \\ \cmidrule(lr){2-4}

& \multirow{2}{*}{\texttt{Wytham}} & \multirow{2}{*}{\unit{0.70}{km}} & Ground vegetation with \\
& & & cluttered trees, severe occlusion \\ \cmidrule(lr){2-4}

& \multirow{2}{*}{\texttt{Forest of Dean}} & \multirow{2}{*}{\unit{0.75}{km}} & Widely spaced oak trees, \\
& & & minimal ground vegetation \\ \cmidrule(lr){2-4}

& \multirow{2}{*}{\texttt{Evo23:00-05}} & \multirow{2}{*}{\unit{4.69}{km}} & Same forest as \texttt{Evo}, \\
& & & different traversals \\
\midrule

\multirow{2}{*}{\rotatebox{90}{Ours}}
& \multirow{2}{*}{\texttt{Evo25:00-08}} & \multirow{2}{*}{\unit{3.29}{km}} & Same forest as \texttt{Evo}, vegetation \\
& & & growth, seasonal change \\
\midrule

\multirow{4}{*}{\rotatebox{90}{Wild-Places}}
& \multirow{2}{*}{\texttt{Venman01-04}} & \multirow{2}{*}{\unit{12.68}{km}} & Moderate vegetation density, mixed \\
& & & terrain, relatively stable appearance \\ \cmidrule(lr){2-4}

& \multirow{2}{*}{\texttt{Karawatha01-04}} & \multirow{2}{*}{\unit{20.24}{km}} & Dense vegetation, uneven terrain, \\
& & & pronounced appearance changes \\
\bottomrule
\end{tabular}%
}
\vspace{-3mm}
\end{table}

%% file: tab/intra-session.tex
\definecolor{Best}{HTML}{def3e6}
\definecolor{Second}{HTML}{ecf8f1}

\newcommand{\best}[1]{\cellcolor{Best}{\bfseries #1}}
\newcommand{\secondbest}[1]{\cellcolor{Second}{\textit{#1}}}

\begin{table}[t]
\centering
\caption{(\textbf{Exp A-1}) Intra-session place recognition performance in Oxford Forest Place Recognition Dataset. \textbf{Bold}: Best, \textit{Italic}: Second-best.}
\label{tab:pr_intra_oxford_evo_dean}
\resizebox{\columnwidth}{!}{%
\begin{tabular}{ll|cccc|cccc}
\toprule
\multicolumn{2}{c|}{\multirow{2}{*}{\textbf{Dimension / Method}}} & 
\multicolumn{4}{c|}{\texttt{Evo:Single}} & 
\multicolumn{4}{c}{\texttt{Forest of Dean}} \\
\multicolumn{2}{c|}{} & 
R@1 & MR & MF1 & AUC & 
R@1 & MR & MF1 & AUC \\ 
\midrule

\multirow{6}{*}{\rotatebox{90}{\shortstack{Learning-based\\(Wild-Places)}}}
& TransLoc3D     & 0.425 & 0.001 & 0.596 & 0.570 & 0.570 & 0.002 & 0.726 & 0.659\\
& LoGG3D-Net     & 0.521 & 0.082 & 0.685 & 0.722 & \secondbest{0.776} & 0.182 & 0.874 & 0.776 \\
& MinkLoc3Dv2    & 0.470 & 0.010 & 0.608 & 0.462 & 0.492 & 0.058 & 0.652 & 0.582\\
& BEVPlace++     & 0.872 & 0.351 & 0.932 & 0.973 & 0.712 & 0.212 & 0.837 & 0.858\\
& ForestLPR      & 0.465 & 0.001 & 0.635 & 0.611 & 0.439 & 0.058 & 0.610 & 0.567\\
& HOTFormerLoc   & 0.530 & 0.019 & 0.743 & 0.820 & 0.583 & 0.126 & 0.748 & 0.780 \\
\midrule

\multirow{8}{*}{\rotatebox{90}{\shortstack{Hand-crafted\\(Untrained)}}}
& SC++           & 0.347 & 0.326 & 0.762 & 0.696 & 0.426 & 0.333 & 0.759 & 0.862 \\
& SOLID          & 0.222 & 0.000 & 0.372 & 0.352 & 0.155 & 0.071 & 0.367 & 0.388\\
& RING++         & 0.567 & 0.126 & 0.223 & 0.125 & 0.674 & 0.194 & 0.781 & 0.718 \\
& BTC            & 0.553 & {0.531} & 0.761 & 0.864 & 0.559 & \secondbest{0.772} & 0.720 & 0.692\\
& MapClosure     & 0.359 & 0.064 & 0.528 & 0.359 & 0.166 & 0.014 & 0.285 & 0.154\\
& {NSM}       & {0.836} & {\secondbest{0.762}} & {{0.961}} & {\secondbest{0.992}} & {0.486} & {0.676} & {0.885} & {\secondbest{0.912}} \\
& TreeLoc        & \secondbest{0.890} & 0.228 & \secondbest{0.955} & {0.984} & 0.348 & 0.045 & 0.897  & {0.891} \\
& \cellcolor[HTML]{f3f7fc}\textbf{TreeLoc++} & \best{0.956} & \best{0.933} & \best{0.991} & \best{0.999} & \best{0.891} & \best{0.898} & \best{0.985}  & \best{0.997} \\

\bottomrule
\end{tabular}%
}
\small
\vspace{-0.1mm}
{$\,$}

\scriptsize 
\raggedright 
$\cdot$ Recall@1 (R@1), Maximum Recall at 100\% Precision (MR), Maximum F1 score (MF1), and Area Under the Precision--Recall Curve (AUC).
\vspace{-3mm}
\end{table}

%% file: tab/intra-session-wild.tex
\definecolor{Best}{HTML}{def3e6}
\definecolor{Second}{HTML}{ecf8f1}

\begin{table}[t]
\centering
\caption{(\textbf{Exp A-2}) Intra-session place recognition performance in Wild-Places.}
\label{tab:intra_wild}
\resizebox{\columnwidth}{!}{%
\begin{tabular}{ll|ccc|ccc|ccc}
\toprule
\multicolumn{2}{c|}{\multirow{2}{*}{\textbf{Category / Method}}} & 
\multicolumn{3}{c|}{\texttt{Venman03}} & 
\multicolumn{3}{c|}{\texttt{Venman04}} & 
\multicolumn{3}{c}{\texttt{Karawatha03}} \\
\multicolumn{2}{c|}{} & 
R@1 & MR & MF1 & 
R@1 & MR & MF1 & 
R@1 & MR & MF1 \\ 
\midrule


\multirow{6}{*}{\rotatebox{90}{\shortstack{Learning-based\\(Wild-Places)}}}
& TransLoc3D     & 0.632 & 0.003 & 0.775 & 0.801 & 0.003 & 0.890 & 0.693 & 0.001 & 0.819 \\
& LoGG3D-Net     & 0.635 & 0.000 & 0.777 & 0.814 & 0.113 & 0.898 & \best{0.867} & 0.009 & \secondbest{0.929} \\
& MinkLoc3Dv2    & \secondbest{0.722} & 0.049 & \secondbest{0.793} & 0.853 & 0.037 & 0.921 & 0.814 & 0.114 & 0.894 \\
& BEVPlace++     & 0.354 & 0.000 & 0.530 & 0.885 & {0.234} & 0.940 & 0.842 & {0.152} & 0.915 \\
& ForestLPR      & 0.501 & 0.025 & 0.668 & 0.743 & 0.045 & 0.853 & 0.599 & 0.010 & 0.749 \\
& HOTFormerLoc   & 0.621 & 0.018 & 0.766 & \secondbest{0.911} & 0.037 & \secondbest{0.953} & 0.849 & 0.069 & 0.921 \\
\midrule

\multirow{8}{*}{\rotatebox{90}{\shortstack{Hand-crafted\\(Untrained)}}}
& SC++           & 0.014 & 0.000 & 0.222 & 0.262 & 0.050 & 0.494 & 0.368 & 0.059 & 0.583 \\
& SOLID          & 0.030 & 0.000 & 0.060 & 0.204 & 0.013 & 0.453 & 0.139 & 0.008 & 0.267 \\
& RING++         & 0.123 & 0.000 & 0.429 & 0.301 & 0.026 & 0.482 & 0.411 & 0.042 & 0.583 \\
& BTC            & 0.158 & 0.030 & 0.364 & 0.168 & 0.061 & 0.579 & 0.236 & 0.022 & 0.563 \\
& MapClosure     & 0.180 & 0.000 & 0.305 & 0.741 & 0.164 & 0.851 & 0.585 & 0.051 & 0.738 \\
& {NSM}       & {0.016} & {\secondbest{0.333}} & {0.500} & {0.524} & {\secondbest{0.645}} & {0.899} & {0.503} & {\secondbest{0.617}} & {0.880} \\
& TreeLoc        & 0.624 & {0.057} & 0.792 & 0.662 & 0.016 & 0.809 & 0.479 & 0.017 & 0.738 \\
& \cellcolor[HTML]{f3f7fc}\textbf{TreeLoc++} & \best{0.940} & \best{0.978} & \best{0.990} & \best{0.924} & \best{0.980} & \best{0.993} & \secondbest{0.864} & \best{0.897} & \best{0.982} \\
\bottomrule
\end{tabular}%
}
\small
\vspace{-0.1mm}
{$\,$}

\scriptsize 
\raggedright 
$\cdot$ Recall@1 (R@1), Maximum Recall at 100\% Precision (MR), and Maximum F1 score (MF1)
\vspace{-5mm}
\end{table}

%% file: tab/intra_localization.tex
\definecolor{Best}{HTML}{def3e6}
\definecolor{Second}{HTML}{ecf8f1}

\begin{table}[t]
\vspace{-2mm}
\centering
\caption{(\textbf{Exp B}) Metric-level localization performance on two datasets.}
\label{tab:wild_places_results}
\resizebox{\columnwidth}{!}{%
\begin{tabular}{ll|cccc|cccc}
\toprule
\multicolumn{2}{c|}{\multirow{2}{*}{\textbf{Dimension / Method}}} & 
\multicolumn{4}{c|}{\texttt{Stein am Rhein}} & 
\multicolumn{4}{c}{\texttt{Karawatha04}} \\
\multicolumn{2}{c|}{} & 
R@50 & SR & ATE & ARE & 
R@50 & SR & ATE & ARE \\ 
\midrule

\multirow{10}{*}{\rotatebox{90}{\shortstack{2D}}}
& LoGG3D-Net               & 0.676 & 0.613 & 0.150 & 0.478 & \secondbest{0.900} & \secondbest{0.860} & 0.131 & 0.229 \\
& MinkLoc3Dv2              & 0.037 & 0.037 & 0.310 & 1.878 & 0.088 & 0.088 & 0.283 & 1.686 \\
& BEVPlace++                    & 0.853 & 0.746 & 0.152 & 0.644 & 0.502 & 0.495 & 0.217 & 0.299 \\
& RING++                        & 0.432 & 0.410 & 0.206 & 1.247 & 0.453 & 0.401 & 0.266 & 0.976 \\
& BTC                            & 0.593 & 0.386 & 0.231 & 0.843 & 0.527 & 0.292 & 0.158 & 0.422 \\
& MapClosure                    & 0.526 & 0.479 & 0.142 & 0.635 & 0.638 & 0.601 & 0.147 & 0.489 \\
& {NSM}                      & {0.605} & {0.605} & {0.082} & {0.363} & {0.700} & {0.700} & {0.059} & {\secondbest{0.170}} \\
& TreeLoc                       & \secondbest{0.976} & \secondbest{0.906} & \secondbest{0.047} & \secondbest{0.197} & 0.654 & 0.465 & \secondbest{0.055} & {0.187} \\
& TreeLoc++                     & \best{0.982} & \best{0.941} & \best{0.046} & \best{0.142} & \best{0.924} & \best{0.902} & \best{0.038} & \best{0.140} \\
\midrule \midrule

\multirow{8}{*}{\rotatebox{90}{\shortstack{3D}}}
& LoGG3D-Net               & 0.656 & 0.599 & 0.173 & 1.051 & \secondbest{0.891} & \secondbest{0.860} & 0.157 & \secondbest{0.510} \\ 
& MinkLoc3Dv2              & 0.018 & 0.018 & 0.269 & 2.682 & 0.046 & 0.046 & 0.327 & 2.540 \\
& BTC                            & 0.310 & 0.243 & 0.239 & 2.219 & 0.403 & 0.272 & 0.227 & {1.678} \\
& MapClosure                    & 0.526 & 0.479 & 0.147 & \secondbest{0.756} & 0.635 & 0.597 & 0.148 & 0.563 \\
& {NSM}                      & {0.523} & {0.523} & {\secondbest{0.120}} & {1.348} & {0.650} & {0.650} & {\secondbest{0.105}} & {0.759} \\
& TreeLoc                       & \secondbest{0.937} & \secondbest{0.878} & {0.121} & 1.378 & 0.593 & 0.423 & {0.135} & 1.151 \\
& TreeLoc++                     & \best{0.939} & \best{0.906} & \best{0.087} & \best{0.470} & \best{0.924} & \best{0.902} & \best{0.068} & \best{0.307} \\
\bottomrule
\end{tabular}%
}
\small
\vspace{-0.1mm}
{$\,$}

\scriptsize 
\raggedright 
$\cdot$ Recall@\unit{50}{cm} (R@50), Success Rate (SR), Average Translation Error (ATE) [m], and Average Rotation Error (ARE) [$^{\circ}$].
\vspace{-5mm}
\end{table}

%% file: tab/inter-session.tex
\begin{table*}[t]
\centering
\caption{(\textbf{Exp C}) Inter-session place recognition performance in Wild-Places.}
\label{tab:inter_session}
\resizebox{\textwidth}{!}{%
\begin{tabular}{ll|cccccccc|cccccccc}
\toprule
\multicolumn{2}{c|}{\multirow{3}{*}{\textbf{Category / Method}}} & 
\multicolumn{8}{c|}{\texttt{Venman01-04} (12 query-database pairs)} & 
\multicolumn{8}{c}{\texttt{Karawatha01-04} (12 query-database pairs)} \\
\cline{3-18}
\multicolumn{2}{c|}{} & 
\multicolumn{2}{c}{R@1} & 
\multicolumn{2}{c}{MR} & 
\multicolumn{2}{c}{MF1} & 
\multicolumn{2}{c|}{AUC} &
\multicolumn{2}{c}{R@1} & 
\multicolumn{2}{c}{MR} & 
\multicolumn{2}{c}{MF1} & 
\multicolumn{2}{c}{AUC} \\
\multicolumn{2}{c|}{} &
Mean & $S$ &
Mean & $S$ &
Mean & $S$ &
Mean & $S$ &
Mean & $S$ &
Mean & $S$ &
Mean & $S$ &
Mean & $S$ \\
\midrule


\multirow{6}{*}{\rotatebox{90}{\shortstack{Learning-based \\ (Wild-Places)}}}
& TransLoc3D     & 0.553 & 0.464 & 0.008 & -0.048 & 0.707 & 0.678 & 0.604 & 0.511 & 0.711 & 1.316 & 0.013 & -0.103 & 0.831 & 1.561 & 0.756 & 1.268 \\
& LoGG3D-Net     & 0.660 & \best{1.242} & 0.003 & -0.282 & 0.813 & \secondbest{1.382} & 0.732 & 0.872 & \best{0.897} & \best{1.848} & 0.002 & -0.278 & \secondbest{0.948} & \best{2.299} & 0.921 & \secondbest{1.638} \\
& MinkLoc3Dv2    & 0.755 & 0.782 & 0.033 & 0.161 & 0.857 & 1.030 & 0.818 & 0.881 & 0.849 & \secondbest{1.607} & 0.062 & 0.150 & 0.919 & 1.889 & 0.911 & 1.592 \\
& BEVPlace++     & 0.735 & 0.701 & 0.091 & -0.025 & 0.840 & 0.920 & 0.841 & 0.842 & 0.767 & 0.934 & 0.016 & -0.193 & 0.868 & 1.179 & 0.767 & 0.822 \\
& ForestLPR      & 0.718 & 0.695 & 0.023 & 0.006 & 0.842 & 1.006 & 0.839 & 1.010 & 0.790 & 1.136 & 0.072 & 0.024 & 0.885 & 1.366 & 0.885 & 1.205 \\
& HOTFormerLoc   & \best{0.899} & 1.173 & {0.155} & -0.057 & \secondbest{0.946} & 1.448 & \secondbest{0.938} & \secondbest{1.350} & \secondbest{0.885} & 1.341 & 0.088 & 0.026 & 0.941 & 1.628 & \secondbest{0.937} & 1.471 \\
\midrule

\multirow{8}{*}{\rotatebox{90}{\shortstack{Hand-crafted \\ (Untrained)}}}
& SC++           & 0.245 & 0.429 & 0.056 & 0.129 & 0.425 & 0.631 & 0.396 & 0.430 & 0.370 & 0.632 & {0.117} & 0.146 & 0.560 & 0.817 & 0.564 & 0.704 \\
& SOLID          & 0.150 & 0.199 & 0.014 & -0.263 & 0.288 & 0.321 & 0.241 & 0.093 & 0.158 & 0.357 & 0.000 & 0.000 & 0.300 & 0.501 & 0.211 & 0.318 \\
& RING++         & 0.322 & 0.448 & 0.057 & -0.004 & 0.503 & 0.624 & 0.500 & 0.458 & 0.341 & 0.713 & 0.030 & -0.143 & 0.509 & 0.848 & 0.402 & 0.439 \\
& BTC            & 0.185 & 0.543 & 0.006 & -0.337 & 0.331 & 0.594 & 0.216 & 0.338 & 0.068 & 0.169 & 0.005 & -0.193 & 0.221 & 0.150 & 0.128 & 0.024 \\
& MapClosure     & 0.478 & 0.338 & 0.054 & -0.008 & 0.618 & 0.459 & 0.465 & 0.279 & 0.484 & 0.420 & 0.036 & -0.121 & 0.641 & 0.571 & 0.479 & 0.347 \\
& {NSM}        & {0.287} & {0.014} & \secondbest{{0.444}} & \secondbest{{0.221}} & {0.701} & {0.634} & {0.824} & {0.859} & {0.248} & {0.086} & \secondbest{{0.544}} & \secondbest{{0.545}} &{ 0.828} & {0.948} & {0.907} & {1.270} \\
& TreeLoc        & 0.766 & 0.940 & 0.046 & -0.019 & 0.890 & 1.202 & 0.853 & 0.954 & 0.486 & 0.938 & 0.003 & {0.785} & {0.858} & 1.747 & 0.838 & 1.577 \\
& \cellcolor[HTML]{f3f7fc}\textbf{TreeLoc++} & \secondbest{0.896} & \secondbest{1.207} & \best{0.970} & \best{1.469} & \best{0.993} & \best{2.303} & \best{0.998} & \best{3.281} & 0.728 & 1.266 & \best{0.849} & \best{1.008} & \best{0.974} & \secondbest{2.149} & \best{0.995} & \best{2.801} \\

\bottomrule
\end{tabular}%
}
\small
\vspace{-0.1mm}
{$\,$}

\scriptsize 
\raggedright 
$\cdot$ Recall@1 (R@1), Maximum Recall at 100\% Precision (MR), Maximum F1 score (MF1), Area Under the Precision--Recall Curve (AUC), and Stability Ratio ($S$).
\vspace{-3mm}
\end{table*}

%% file: tab/local_to_global.tex
\begin{table}[t]
\centering
\caption{(\textbf{Exp D}) Global relocalization results, where loop closure and inter-session constraints are added incrementally from additional sequences.}
\label{tab:local_to_global}
\resizebox{\columnwidth}{!}{%
\begin{tabular}{lll|cc|cc|cc|cc}
\toprule
\multicolumn{2}{l}{} & & \multicolumn{2}{c|}{LoGG3D-Net} & \multicolumn{2}{c|}{BTC} & \multicolumn{2}{c|}{MapClosure} & TreeLoc & \cellcolor[HTML]{f3f7fc}\textbf{TreeLoc++} \\ 
 &  & Metrics & SR & F1 & SR & F1 & SR & F1 &  \multicolumn{2}{c}{$\mathcal{O} > 0.2$} \\ \midrule

\multirow{16}{*}{\rotatebox{90}{\texttt{25:00}}} & \multirow{4}{*}{\rotatebox{90}{\texttt{+ Loop}}} 
 & Size & \multicolumn{2}{c|}{\secondbest{1.2 MB}} & \multicolumn{2}{c|}{44MB} & \multicolumn{2}{c|}{2.6 MB} & \multicolumn{2}{c}{\best{227.9 KB}} \\
 & & \# Const. & 16 & \best{55} & 5 & 14 & 1 & 1 & 33 & \secondbest{38} \\
 & & ATE  & 0.35 & 10.40 & 0.37 & \secondbest{0.31} & 0.46 & 0.46 & \best{0.31} & 0.32 \\
 & & ARE  & 1.53 & 45.61 & 1.63 & \best{1.33} & 1.96 & 1.96 & \secondbest{1.49} & 1.53 \\ \cmidrule(lr){2-11}

 & \multirow{4}{*}{\rotatebox{90}{\texttt{+ 25:02}}} 
 & Size & \multicolumn{2}{c|}{\secondbest{3.7 MB}} & \multicolumn{2}{c|}{90.8 MB} & \multicolumn{2}{c|}{7.7 MB} & \multicolumn{2}{c}{\best{11.2 KB}} \\
 & & \# Const. & 0 & 29 & 0 & 0 & 0 & \secondbest{25} & 2 & \best{58} \\
 & & ATE  & 33.73 & 20.60 & 33.80 & 33.78 & 34.05 & 15.20 & \secondbest{10.21} & \best{0.57} \\
 & & ARE  & 41.58 & 63.96 & 42.20 & 41.52 & 42.48 & 24.36 & \secondbest{7.93} & \best{1.75} \\ \cmidrule(lr){2-11}

 & \multirow{4}{*}{\rotatebox{90}{\texttt{+ 25:03}}} 
 & Size & \multicolumn{2}{c|}{\secondbest{2.0 MB}} & \multicolumn{2}{c|}{20.2 MB} & \multicolumn{2}{c|}{2.9 MB} & \multicolumn{2}{c}{\best{5.9 KB}} \\
 & & \# Const. & 0 & \secondbest{17} & 0 & 1 & 0 & 15 & 7 & \best{24} \\
 & & ATE  & 33.73 & 19.35 & 33.80 & 23.97 & 34.05 & 11.50 & \secondbest{2.74} & \best{0.60} \\
 & & ARE  & 41.58 & 53.14 & 42.20 & 26.64 & 42.48 & 37.06 & \secondbest{7.05} & \best{1.92} \\ \cmidrule(lr){2-11}

 & \multirow{4}{*}{\rotatebox{90}{\texttt{+ 25:05}}} 
 & Size & \multicolumn{2}{c|}{\secondbest{3.9 MB}} & \multicolumn{2}{c|}{74MB} & \multicolumn{2}{c|}{7.3 MB} & \multicolumn{2}{c}{\best{14.9 KB}} \\
 & & \# Const. & 0 & 1 & 0 & 0 & 0 & \secondbest{6} & 3 & \best{25} \\
 & & ATE  & 33.73 & 18.66 & 33.80 & 23.97 & 34.05 & 25.93 & \secondbest{1.96} & \best{0.53} \\
 & & ARE  & 41.58 & 51.55 & 42.20 & 26.64 & 42.48 & 80.74 & \secondbest{3.49} & \best{1.91} \\ \midrule \midrule

\multirow{8}{*}{\rotatebox{90}{\texttt{V-01}}} & \multirow{4}{*}{\rotatebox{90}{\texttt{+ Loop}}} 
 & Size & \multicolumn{2}{c|}{\secondbest{11.3 MB}} & \multicolumn{2}{c|}{1.60 GB} & \multicolumn{2}{c|}{21.7 MB} & \multicolumn{2}{c}{\best{3.2 MB}} \\
 & & \# Const. & 314 & \best{1400} & 61 & 0 & 84 & 336 & 273 & \secondbest{338} \\
 & & ATE  & \best{6.74} & 174.78 & 7.18 & 175.88 & 6.88 & 7.95 & 7.53 & \secondbest{6.76} \\
 & & ARE  & \secondbest{2.94} & 71.39 & 3.06 & 66.90 & 2.96 & 3.52 & 3.36 & \best{2.92} \\ \cmidrule(lr){2-11}

 & \multirow{4}{*}{\rotatebox{90}{\texttt{+ V-02}}} 
 & Size & \multicolumn{2}{c|}{\secondbest{54.1 MB}} & \multicolumn{2}{c|}{2.00 GB} & \multicolumn{2}{c|}{130.5 MB} & \multicolumn{2}{c}{\best{68.7 KB}} \\
 & & \# Const. & 217 & \best{1451} & 6 & 46 & 0 & 19 & 69 & \secondbest{602} \\
 & & ATE  & \secondbest{1.72} & 146.71 & 2.90 & 295.14 & 23.60 & 104.68 & 5.47 & \best{1.23} \\
 & & ARE  & \best{0.99} & 52.05 & 1.56 & 81.21 & 5.00 & 75.77 & 2.25 & \secondbest{1.13} \\ \bottomrule
\end{tabular}%
}
\small
\vspace{-0.1mm}
{$\,$}

\scriptsize 
\raggedright 
$\cdot$ Validated success constraints (SR), F1 score-based constraints (F1), descriptor and inventory storage (Size), number of selected constraints (\# Const.), Average Translation Error (ATE) [m], and Average Rotation Error (ARE) [$^{\circ}$].
\vspace{-3mm}
\end{table}

%% file: tab/ablation_rotation.tex
\begin{table}[!t]
\centering
\caption{(\textbf{Exp F-1}) Invariance analysis across different orientation}
\label{tab:invariance_analysis}

\resizebox{\columnwidth}{!}{%
\begin{tabular}{l | ccc | ccc | ccc}
\toprule
\multirow{2}{*}{Methods} &
  \multicolumn{3}{c|}{Yaw (13 pairs)} &
  \multicolumn{3}{c|}{Roll \& Pitch (13 pairs)} &
  \multicolumn{3}{c}{Full Rot. (169 pairs)} \\
 &
  Mean & $\sigma$ & $S$ & Mean & $\sigma$ & $S$ & Mean & $\sigma$ & $S$ \\ \midrule
MapClosure   & 0.363  & 0.002 & 2.281          & 0.273  & 0.132 & 0.316          & 0.281  & 0.130 & 0.336          \\
LoGG3D-Net   & 0.664  & 0.002 & 2.592          & 0.168  & 0.258 & -0.186         & 0.166  & 0.252 & -0.181         \\
HOTFormerLoc & 0.385  & 0.361 & 0.028          & 0.191  & 0.288 & -0.179         & 0.109  & 0.196 & -0.255         \\
\cellcolor[HTML]{f3f7fc}\textbf{TreeLoc++}    & \best{0.951} & \best{0.001} & \best{3.024} & \best{0.952} & \best{0.001} & \best{3.075} & \best{0.952} & \best{0.001} & \best{3.200} \\ \bottomrule
\end{tabular}%
}
\small
\vspace{-0.1mm}
{$\,$}

\scriptsize 
\raggedright 
$\cdot$ Standard deviation ($\sigma$) and Stability Ratio ($S$).
\vspace{-4mm}
\end{table}

%% file: tab/time_consumption.tex


\begin{table}[!t]
\centering
\caption{(\textbf{Exp F-2}) Total runtime for all queries on large databases [s]}
\label{tab:efficiency_runtime}
\resizebox{0.75\columnwidth}{!}{%
\begin{tabular}{l|cc|c}
\toprule
Method & Place Recognition & Pose Estimation & Total Time \\
\midrule
BTC         & 382.29      & 10.49    & 392.78 \\
MapClosure  & 5.60        & 4.46     & 10.06 \\
\cellcolor[HTML]{f3f7fc}\textbf{TreeLoc++} & \best{2.98} & \best{3.43} & \best{6.41} \\
\bottomrule
\end{tabular}%
}
\vspace{-4mm}
\end{table}

%% file: tab/ablation_coarse.tex
\begin{table}[t]
\centering
\caption{(\textbf{Exp G-1}) Ablation studies on TDH and PDH}
\label{tab:descriptor_ablation}
\resizebox{\columnwidth}{!}{%
\begin{tabular}{cc|cccc|cccc}
\toprule
\multicolumn{2}{c|}{Config.} &
\multicolumn{4}{c|}{\texttt{Stein am Rhein}} &
\multicolumn{4}{c}{\texttt{Karawatha03}} \\ 
TDH & PDH & \# TP & FNR  & R@1 & R@50 & \# TP & FNR  & R@1 & R@50 \\ \midrule
\xmark & \xmark & - & - & 0.939 & 0.935 & - & - & 0.820 & 0.845 \\
\checkmark & \xmark & 15.069 & 0.024 & 0.933 & \best{0.943} & 3.852 & 0.063 & 0.834 & 0.863 \\
\xmark & \checkmark & 16.701 & 0.022 & 0.939 & 0.933 & 3.771 & 0.076 & 0.817 & 0.845 \\
\cellcolor[HTML]{f3f7fc} 
\checkmark & \cellcolor[HTML]{f3f7fc}\checkmark & \best{17.016} & \best{0.020} & \best{0.941} & 0.939 & \best{4.662} & \best{0.038} & \best{0.864} & \best{0.874} \\ 
\bottomrule
\end{tabular}%
}
\small
\vspace{-0.1mm}
{$\,$}

\scriptsize 
\raggedright 
$\cdot$ Number of true positives (\# TP), False Negative Rate (FNR), Recall@1 (R@1) and Recall@\unit{50}{cm} (R@50).
\vspace{-3mm}
\end{table}

%% file: tab/ablation.tex
\begin{table}[t]
\centering
\caption{(\textbf{Exp G-2}) Ablation studies on refinements}
\label{tab:refinement_ablation_updated}
\resizebox{\columnwidth}{!}{%
\begin{tabular}{c c c|cccc|cccc}
\toprule
\multicolumn{3}{c|}{{Refinement}} &
\multicolumn{4}{c|}{\texttt{Stein am Rhein}} &
\multicolumn{4}{c}{\texttt{Karawatha03}} \\ 
Pen. & Vot. & Filt. &
R@1 & AUC & R@50 & SR & 
R@1 & AUC & R@50 & SR \\ \midrule
\xmark & \xmark & \xmark & 0.902 & 0.983 & 0.933 & 0.872 & 0.644 & 0.825 & 0.809 & 0.577 \\
\checkmark & \xmark & \xmark & 0.927 & \best{0.998} & 0.935 & 0.894 & 0.803 & 0.995 & 0.808 & 0.730 \\
\checkmark & \checkmark & \xmark & 0.933 & \best{0.998} & \best{0.939} & 0.902 & 0.852 & \best{0.998} & 0.850 & 0.796 \\
\rowcolor[HTML]{f3f7fc} 
\checkmark & \checkmark & \checkmark & \best{0.941} & \best{0.998} & \best{0.939} & \best{0.906} & \best{0.864} & \best{0.998} & \best{0.874} & \best{0.816} \\ 
\bottomrule
\end{tabular}%
}
\small
\vspace{-0.1mm}
{$\,$}

\scriptsize 
\raggedright 
$\cdot$ Spatial Penalty (Pen.), Yaw Voting (Vot.), DBH filtering (Filt.), Recall@1 (R@1), Area Under the Precision--Recall Curve (AUC), Recall@\unit{50}{cm} (R@50) and Success Rate (SR).
\vspace{-4mm}
\end{table}

%% file: tab/ablation_others.tex
\begin{table}[t]
\centering
\caption{(\textbf{Exp G-3}) Ablation studies on system components}
\label{tab:system_ablation}
\resizebox{\columnwidth}{!}{%
\begin{tabular}{c c c|cccc|cccc}
\toprule
\multicolumn{3}{c|}{{System Components}} &
\multicolumn{4}{c|}{\texttt{Stein am Rhein}} &
\multicolumn{4}{c}{\texttt{Karawatha03}} \\ 
Reuse & IRLS & Corr. &
R@1  & R@50  & ATE  & ARE  & 
R@1  & R@50  & ATE  & ARE  \\ \midrule
\xmark & \xmark & \xmark & 0.939 & 0.915 & 0.152 & 1.331 & 0.825 & 0.844 & 0.099 & 0.730 \\
\checkmark & \xmark & \xmark & 0.939 & 0.921 & 0.154 & 1.356 & 0.852 & 0.852 & 0.101 & 0.737 \\
\checkmark & \checkmark & \xmark & \best{0.941} & 0.927 & 0.149 & 1.359 & \best{0.864} & 0.871 & 0.103 & 0.751 \\
\rowcolor[HTML]{f3f7fc} 
\checkmark & \checkmark & \checkmark & \best{0.941} & \best{0.939} & \best{0.087} & \best{0.470} & \best{0.864} & \best{0.874} & \best{0.080} & \best{0.260} \\ 
\bottomrule
\end{tabular}%
}
\small
\vspace{-0.1mm}
{$\,$}

\scriptsize 
\raggedright 
$\cdot$ Preceding-inventory reuse (Reuse), Iterative Reweighted Least Squares (IRLS), Joint Vertical Correction (Corr.), Recall@1 (R@1), Recall@\unit{50}{cm} (R@50), Average Translation Error (ATE) [m], and Average Rotation Error (ARE) [$^{\circ}$].
\vspace{-4mm}
\end{table}

%% file: 5_limitation.tex
\section{Limitations of TreeLoc++}
\begin{figure}[!t]
    \centering
    \begin{subfigure}[t]{\columnwidth}
        \centering
        \includegraphics[width=\linewidth]{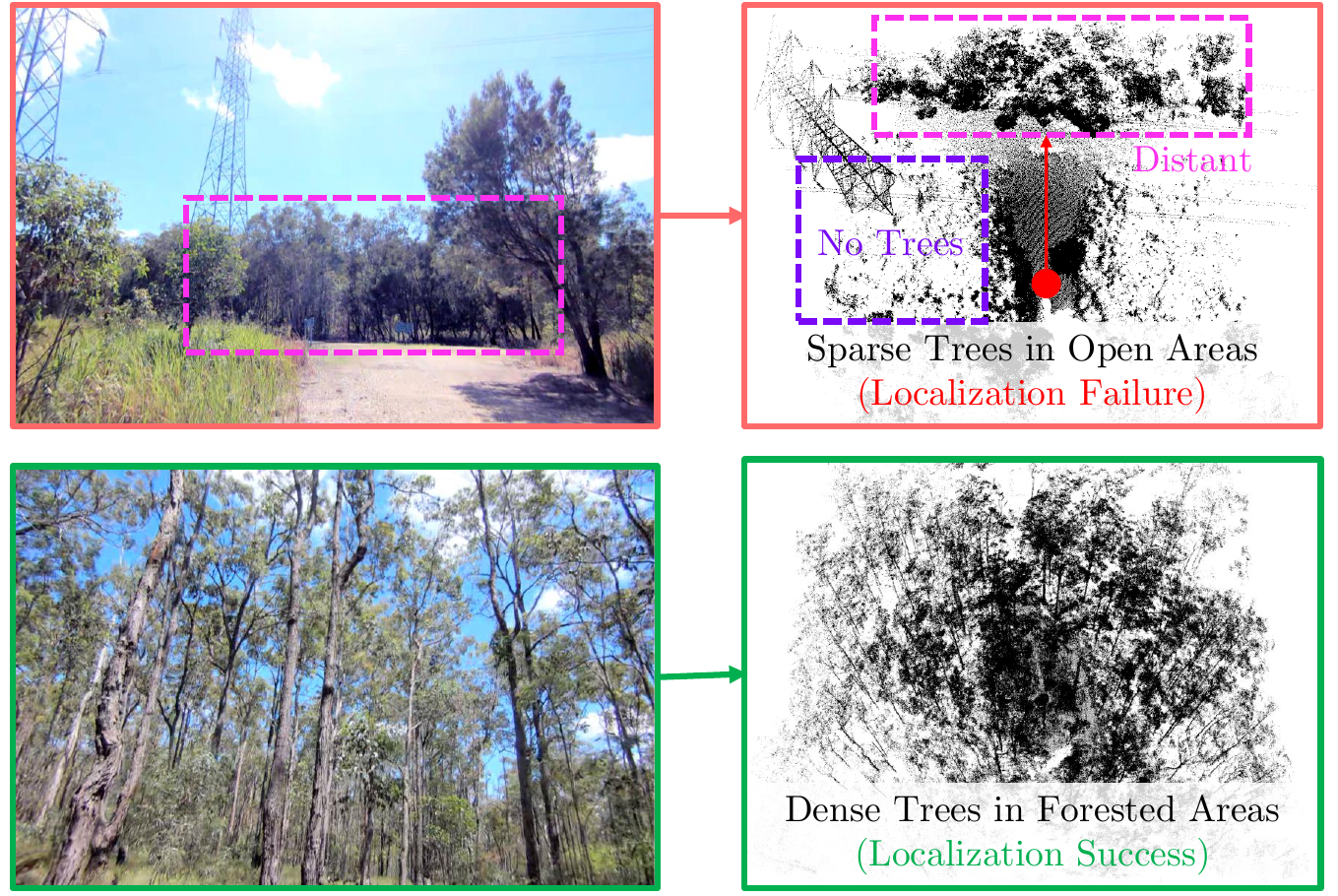}
        \caption{Examples of Localization Failure and Success across Different Areas}
        \label{fig:localization_failure}
    \end{subfigure}\hfill
    \vspace{-1mm}
    \begin{subfigure}[t]{\columnwidth}
        \centering
        \includegraphics[width=\linewidth]{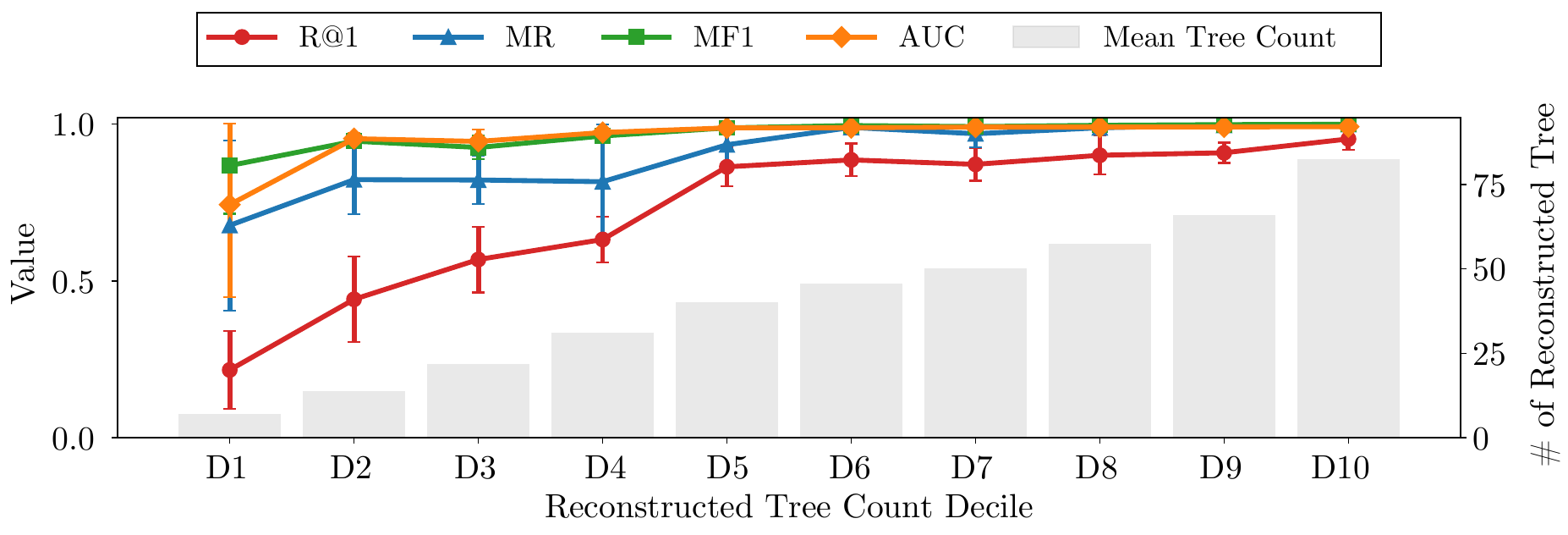}
        \caption{Sensitivity to Reconstructed Tree Count in \texttt{Karawatha}}
        \label{fig:tree_count}
    \end{subfigure}
    \begin{subfigure}[t]{\columnwidth}
        \centering
        \includegraphics[width=\linewidth]{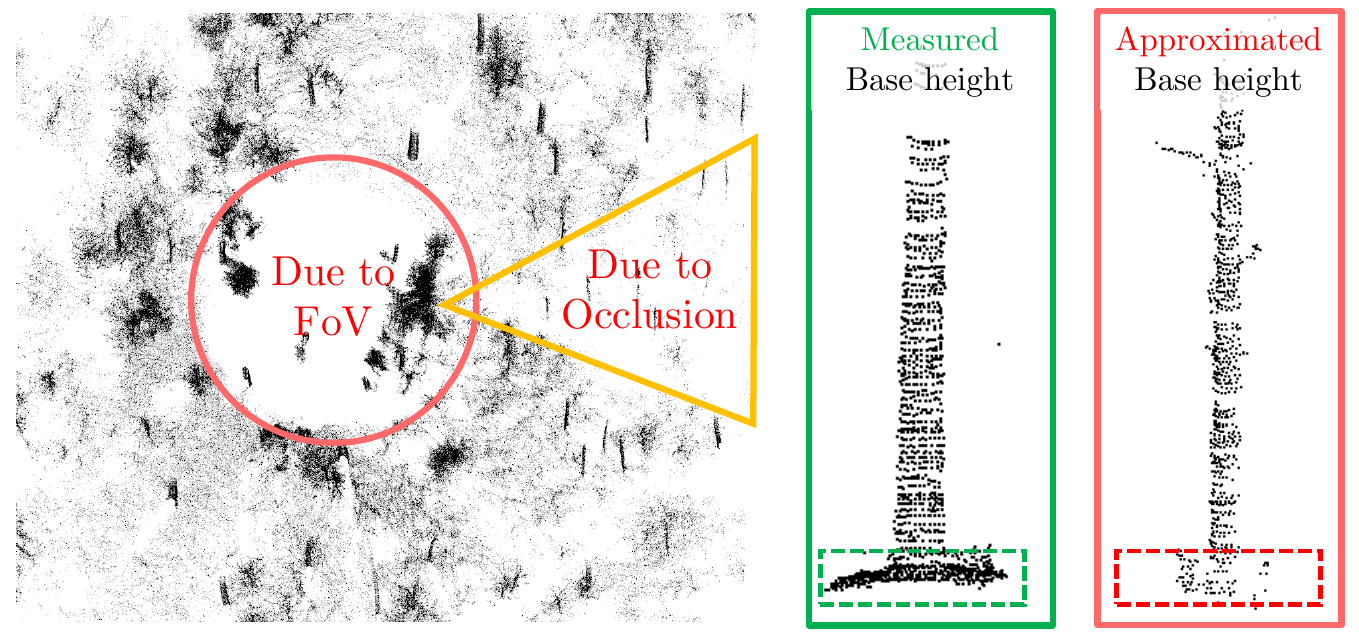}
        \caption{Inaccurate Base Height from FoV and Occlusion}
        \label{fig:inaccurate_height}
    \end{subfigure}
    \vspace{-3mm}
    \caption{Limitations of TreeLoc++. (a) Sparse and distant tree observations in open areas provide insufficient geometric constraints, leading to failure. In contrast, dense forests enable successful localization through abundant correspondences. {(b) Inter-session place recognition performance on \texttt{Karawatha} versus reconstructed tree count, where D1-D10 denote deciles from the sparsest to the densest scans. Performance generally improves with increasing tree count, showing that sparse scans provide weaker geometric constraints and lower overlap.} (c) Limited sensor FoV and occlusions prevent direct ground measurements, requiring an approximation of the base height that may introduce errors relative to the true base height.}
    \label{fig:limitation}
    \vspace{-3mm}
\end{figure}

\noindent \textbf{Failure Cases in Open Areas:}
TreeLoc++ performance varies substantially between the open areas and deep-forest regions.
In the open areas of \texttt{Karawatha03}, tree observations tend to be sparse and distant, providing insufficient geometric constraints for robust localization as shown in \figref{fig:localization_failure}. This often leads to degraded downstream pose estimation. However, in deep forests, the density of nearby trunks allows the system to achieve highly successful localization through stable correspondences. {As also shown in \figref{fig:tree_count}, the average inter-session results across all pairs in \texttt{Karawatha} indicate that recall decreases when only a small number of trees are available, and gradually recovers as tree density increases. In contrast, F1 score and AUC degrade much less, suggesting that TreeLoc++ can still evaluate candidate matches precisely even in sparse scenes.} While TreeLoc++ is environment-dependent, it naturally complements GNSS: open areas typically offer strong GNSS reception thanks to the lack of canopy obstruction, whereas dense forests provide richer geometric features for reliable localization when GNSS becomes unreliable. This suggests a practical deployment strategy that uses GNSS in open areas and switches to TreeLoc++ as tree density increases.

\noindent \textbf{Inaccurate Base Height under Terrain Occlusion:}
TreeLoc++ relies on estimated base heights for full 6-DoF alignment, so base height errors can destabilize roll, pitch, and height refinement.
As shown in \figref{fig:inaccurate_height}, this occurs when limited sensor FoV or severe occlusions prevent sufficient terrain observations around a tree. In such cases, the base height must be approximated from nearby data, which can deviate from the true value.
This issue is more pronounced in local inventory mode due to limited payload coverage. In contrast, global inventory mode mitigates it by leveraging all explored payloads during tree reconstruction, providing broader multi-view coverage for base height estimation.
Nevertheless, the practical impact is marginal: the error increase from 3-DoF to 6-DoF alignment remains at the centimeter level with stable rotation error. Furthermore, RANSAC-based filtering effectively rejects base height outliers, as supported by results in \tabref{tab:wild_places_results} and \textcolor{blue}{Appendix~C}.

%% file: 6_conclusion.tex
\section{Conclusion}
This paper presented TreeLoc++, a global localization framework for forest environments based on \ac{DFI}s that eliminates the need to store raw point clouds. TreeLoc++ builds local and global inventories by reconstructing tree instances from aggregated payloads. For localization, it performs stem-axis-based alignment and 2D projection, and retrieves candidates via a coarse-to-fine matching strategy, starting with TDH and PDH, and 2D triangle descriptor for efficient retrieval. It then rejects outliers through DBH filtering and yaw-consistent inlier voting, and refines the 6-DoF pose through geometric verification with constrained optimization of roll, pitch, and height.
We evaluated TreeLoc++ on 27 sequences from three datasets, including multi-session data collected two years apart. The results demonstrate robust place recognition and centimeter-level localization accuracy, while maintaining a compact map representation of just 250~KB across 15 sessions covering \unit{7.98}{km} trajectories. TreeLoc++ outperformed both hand-crafted and learning-based baselines in accuracy and achieved the lowest runtime. Ablation studies further confirmed that each proposed component contributes to the overall performance and robustness.
Future work will focus on inventory construction. Although TreeLoc++ is accurate and efficient, the current pipeline relies on individual tree segmentation and forest inventory generation using RealtimeTrees, which requires aggregating point cloud payloads. With improvements to \ac{DFI} construction in the future, we would envisage TreeLoc++ achieving further gains in localization performance.

%% file: appendix.tex
\normalsize
\section{Appendix}
\subsection{Ground Truth Generation for Evo25}
To generate consistent ground truth trajectories for the nine sequences in \texttt{Evo25:00-08}, we first estimated local trajectories using VILENS~\cite{wisth2022vilens} with LiDAR and IMU data. The sessions were then coarsely aligned based on common locations. To further refine this alignment, inter-session constraints were added at the point cloud level via robust pairwise frame matching using TEASER++~\cite{yang2020teaser}. All TEASER++ matches were manually verified to ensure high alignment quality. Candidate frame pairs were selected via nearest neighbor matching across all possible session combinations. Finally, all local trajectories and inter-session constraints were jointly optimized using GTSAM to produce globally consistent multi-session trajectories.

\begin{figure}[!t]
    \centering
    \begin{subfigure}[t]{\columnwidth}
        \centering
        \includegraphics[width=\columnwidth]{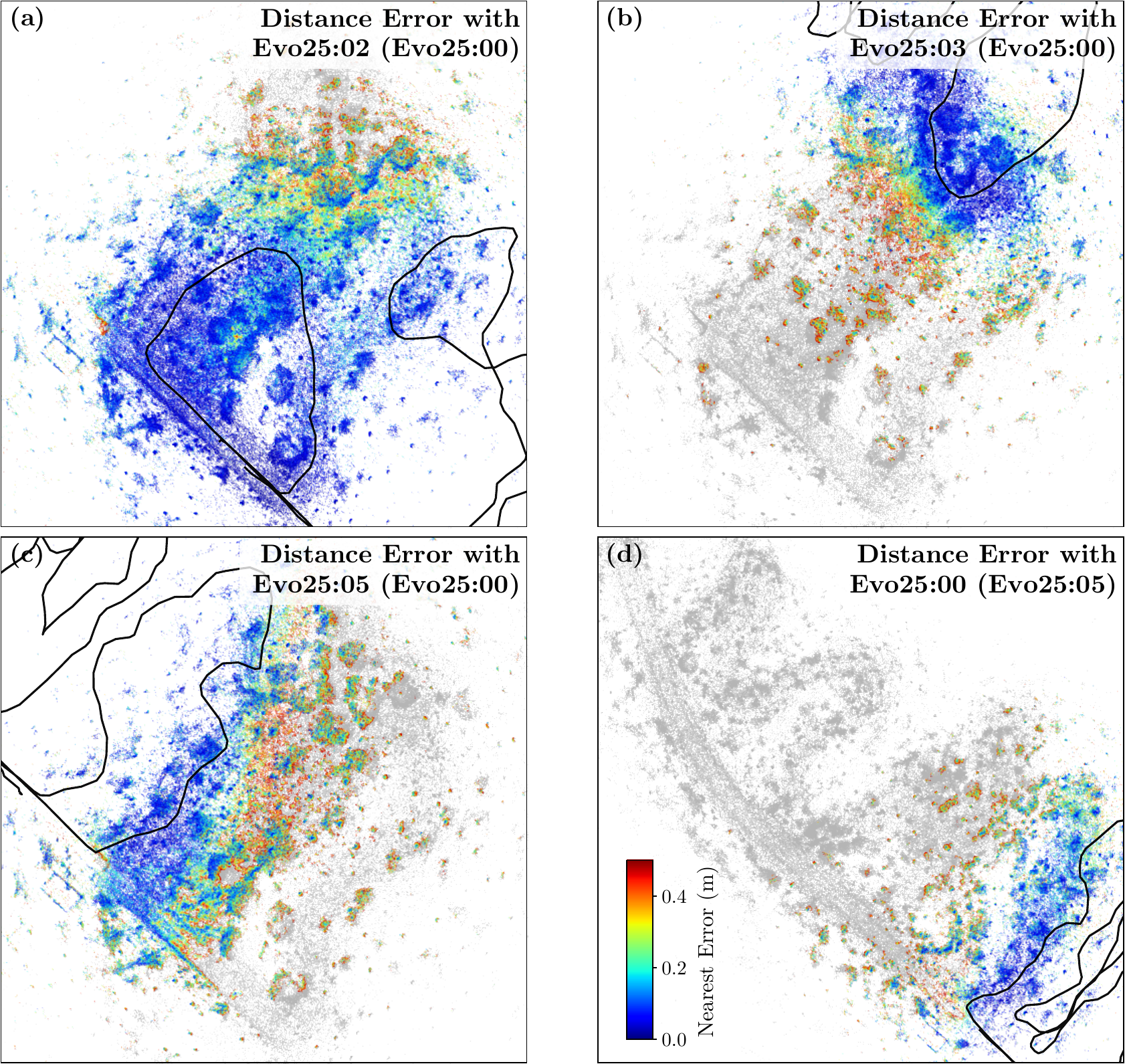}
        \vspace{-2mm}
    \end{subfigure}
    
    \begin{subfigure}[t]{\columnwidth}
        \centering
        \includegraphics[width=\columnwidth]{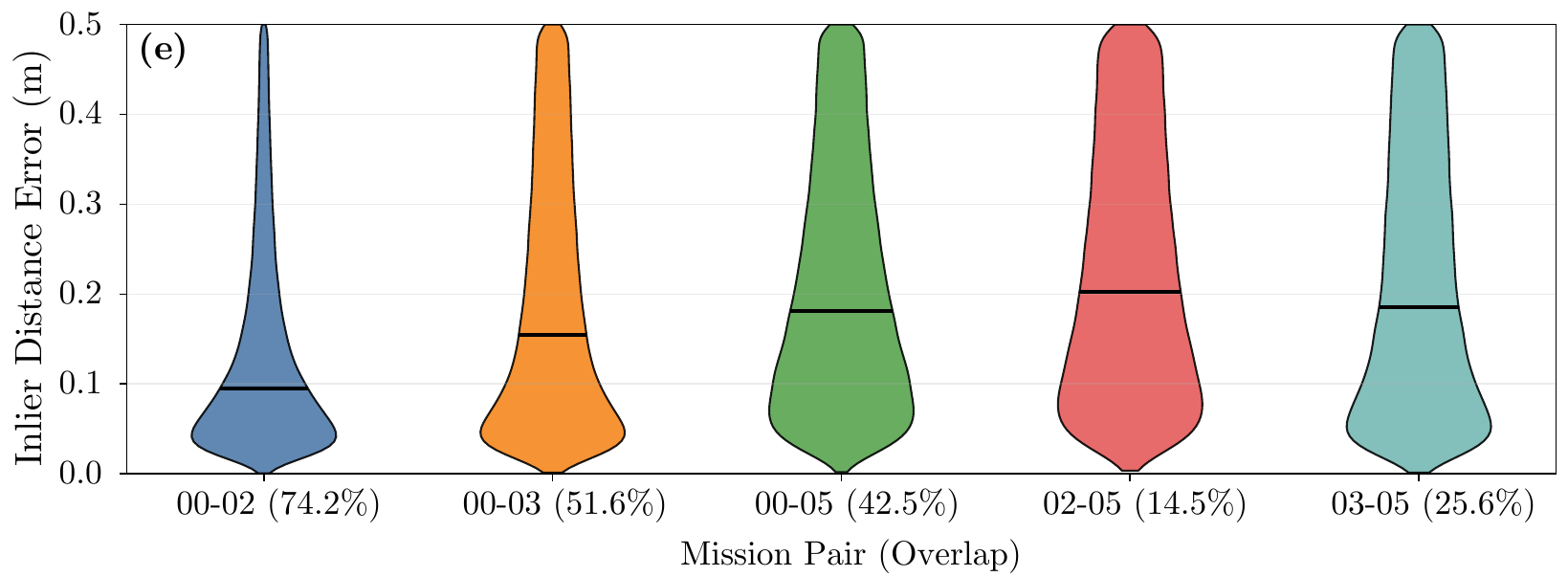}
    \end{subfigure}
    \vspace{-5mm}
\caption{{Registration consistency across \texttt{Evo25} missions. (a-c) Top-view error maps of \texttt{Evo25:00} with respect to \texttt{Evo25:02}, \texttt{Evo25:03}, and \texttt{Evo25:05}. (d) Top-view error map of \texttt{Evo25:05} with respect to \texttt{Evo25:00}. Points whose nearest neighbor lies within \unit{0.5}{m} are treated as inliers and color-coded by error (blue: low, red: high), while invalid correspondences are shown in gray. Black lines indicate the optimized trajectory of the reference mission. Overlapping regions mostly show low error, whereas boundary regions with limited overlap show larger distances. (e) Violin plots of inlier error distributions for representative mission pairs, with overlap ratios in parentheses and the median shown by a black bar. The densest portion of the distribution remains below {0.1} {m}.}}
    \label{fig:gt_registration}
    \vspace{-3mm}
\end{figure}

{To quantify the geometric reliability of this pseudo-ground truth, we evaluated cross-session registration consistency using point-to-point nearest-neighbor distances, following the same nearest-neighbor error analysis used in \figref{fig:intra_registration} and a point-cloud distance-based validation similar to~\cite{knights2022wild}. Using representative mission point clouds from \texttt{Evo25:00}, \texttt{Evo25:02}, \texttt{Evo25:03}, and \texttt{Evo25:05}, which were also used in \secref{sec:relocalization}, we transformed them into the optimized global frame and measured nearest-neighbor distances between mission pairs. As shown in \figref{fig:gt_registration}a-d, regions with substantial overlap mostly exhibit low error, whereas boundary regions with limited overlap show larger distances because sparse LiDAR observations provide fewer reliable cross-session correspondences.}

{\figref{fig:gt_registration}e summarizes these trends using violin plots of the distance error distributions for representative mission pairs. Although the median error becomes slightly larger as overlap decreases, the medians remain around or below \unit{0.2}{m}, and the highest-density portions of all distributions lie below \unit{0.1}{m}. These results indicate that the optimized trajectories are sufficiently accurate to serve as pseudo-ground truth for evaluating TreeLoc++. The remaining uncertainty is mainly confined to low overlap boundary regions, while the evaluated trajectory segments are mostly drawn from revisited areas with substantial overlap. Therefore, its impact on the main evaluation is expected to be small.}

\subsection{Learning-based Methods with Training Datasets}
We analyzed the impact of training data on learning-based baselines. Each learning-based method was evaluated under three training regimes: (1) urban, (2) in-domain forest, and (3) out-of-domain forest.

\begin{figure}[!t]
    \centering
    \begin{subfigure}[t]{\columnwidth}
        \centering
        \includegraphics[width=\columnwidth]{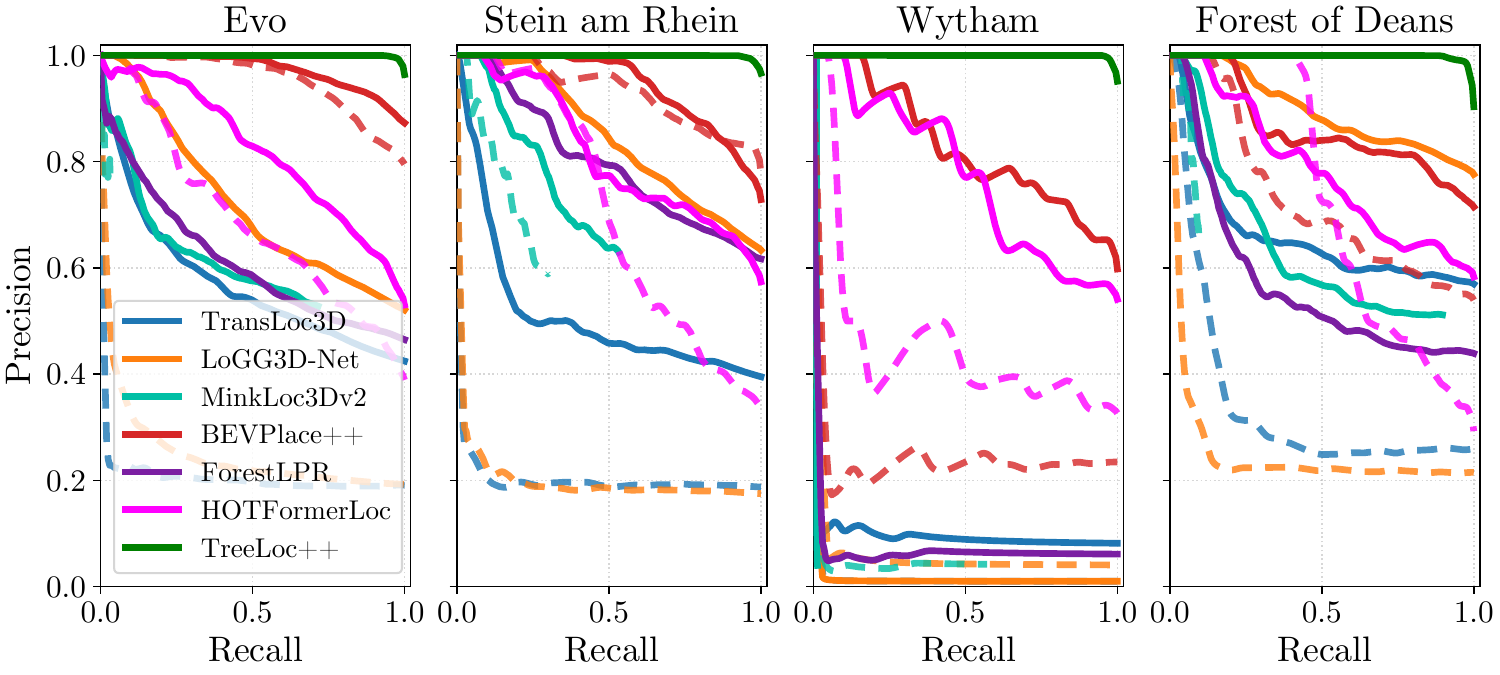}
        \vspace{-6mm}
        \caption{Precision-Recall Curve in Oxford Forest Place Recognition Dataset}
        \vspace{-2mm}
        \label{fig:pr_curve_appendix}
    \end{subfigure}
    \begin{subfigure}[t]{\columnwidth}
        \centering
        \includegraphics[width=\columnwidth]{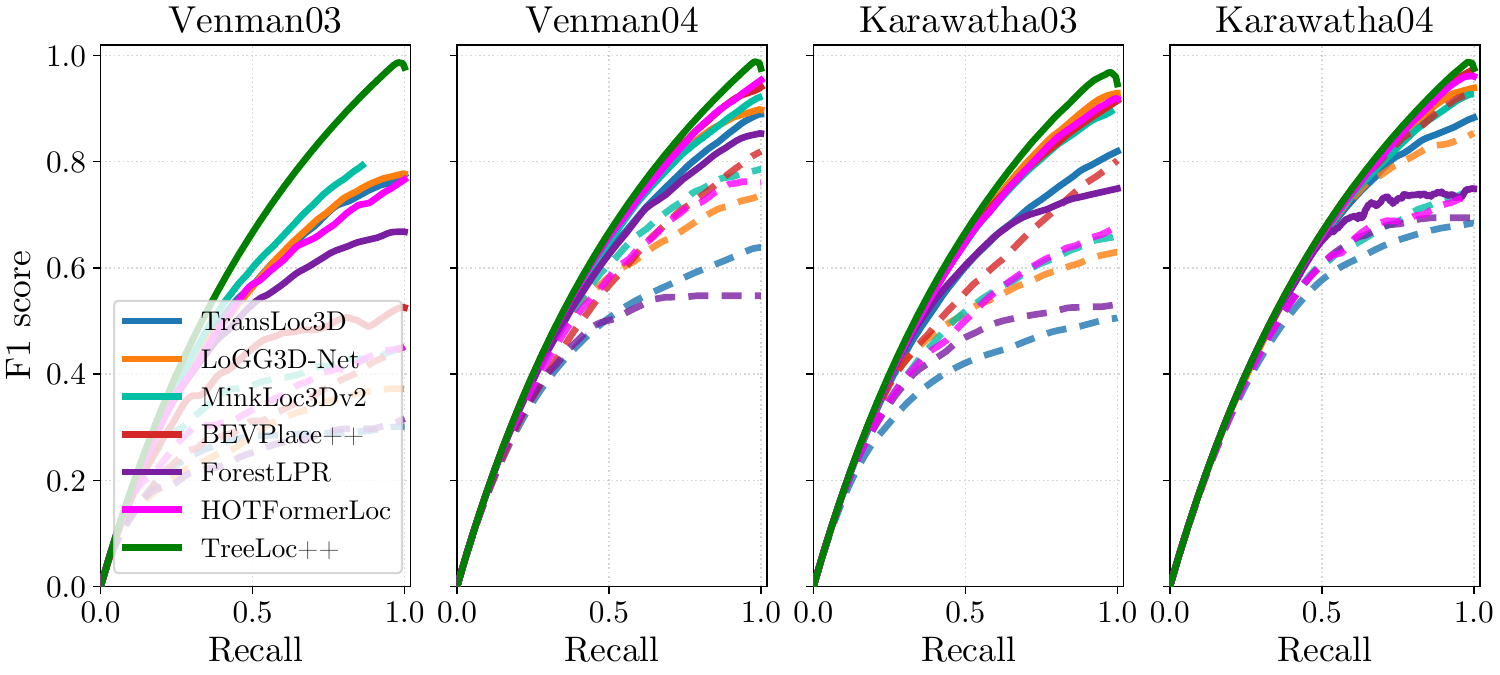}
        \vspace{-6mm}
        \caption{F1-Recall Curve in Wild-Places}
        \vspace{-2mm}
        \label{fig:f1_curve_appendix}
    \end{subfigure}
\caption{Precision--Recall and F1--Recall curves showing the impact of training domains.
(Top) Urban-trained models (dashed) perform worse on Oxford Forest, compared to models trained on forest datasets (solid), indicating limited generalization to natural environments.
(Bottom) Even across forest datasets, models trained on Oxford Forest (dashed) perform worse on Wild-Places, compared to those trained directly on Wild-Places (solid), highlighting challenges in cross-forest generalization.}
    \vspace{-3mm}
\end{figure}

\noindent \textbf{Urban vs. Forest:}
We analyzed the impact of training data on learning-based baselines by using urban datasets (KITTI~\cite{geiger2013vision} and Oxford RobotCar~\cite{maddern20171}) for training, and evaluating the models on the Oxford Forest Place Recognition Dataset. BEVPlace++ and LoGG3D-Net were trained on KITTI, while the other models used Oxford RobotCar. All models employed pretrained checkpoints, except for TransLoc3D.
As shown in \tabref{tab:intra_appendix_oxford}, models trained on urban datasets exhibited limited generalization to forest environments. This trend is also reflected in \figref{fig:pr_curve_appendix}, where urban-trained models underperformed in terms of PR curves, with performance degrading across all sequences.
In contrast, models trained on Wild-Places achieved moderate performance, particularly in structured scenes such as \texttt{Stein am Rhein}. While training on Wild-Places improved overall performance, their performance still remained limited across the evaluation sequences. This suggests two key insights: first, that models trained on urban datasets struggle to generalize to forest environments, even though such datasets are large-scale and readily available, which limits their practical applicability; and second, that a domain gap persists even within forest environments, degrading performance across evaluation sequences. We further investigate this forest domain gap in the Wild-Places experiments.

\input{tab/appendix_learning}

\input{tab/appendix_learning_wild}

\noindent \textbf{In-domain vs. out-of-domain forest:}
We examined the effect of training data from different forest domains by comparing models trained on Wild-Places and the Oxford Forest Place Recognition Dataset, evaluated on Wild-Places sequences. As shown in \tabref{tab:intra_appendix_wild}, methods trained on Oxford Forest showed a clear drop in performance, comparable to that of urban-trained models, despite both datasets being forest environments. Similarly, in \figref{fig:f1_curve_appendix}, methods trained on Oxford Forest consistently yielded lower F1 scores, indicating decreased retrieval quality. These results suggest that, even within the same broader domain, forest environments can differ substantially, making it difficult for learning-based methods to generalize. This highlights the potential of well-designed hand-crafted methods to provide more robust performance across diverse forest conditions. In particular, TreeLoc++, which achieves learning-comparable performance without training, emerges as a strong alternative to learning-based approaches in forest environments.

\input{tab/intra_pose_estimation}

\begin{figure}[t]
    \centering
    \begin{subfigure}[t]{\columnwidth}
        \centering
        \includegraphics[width=\columnwidth]{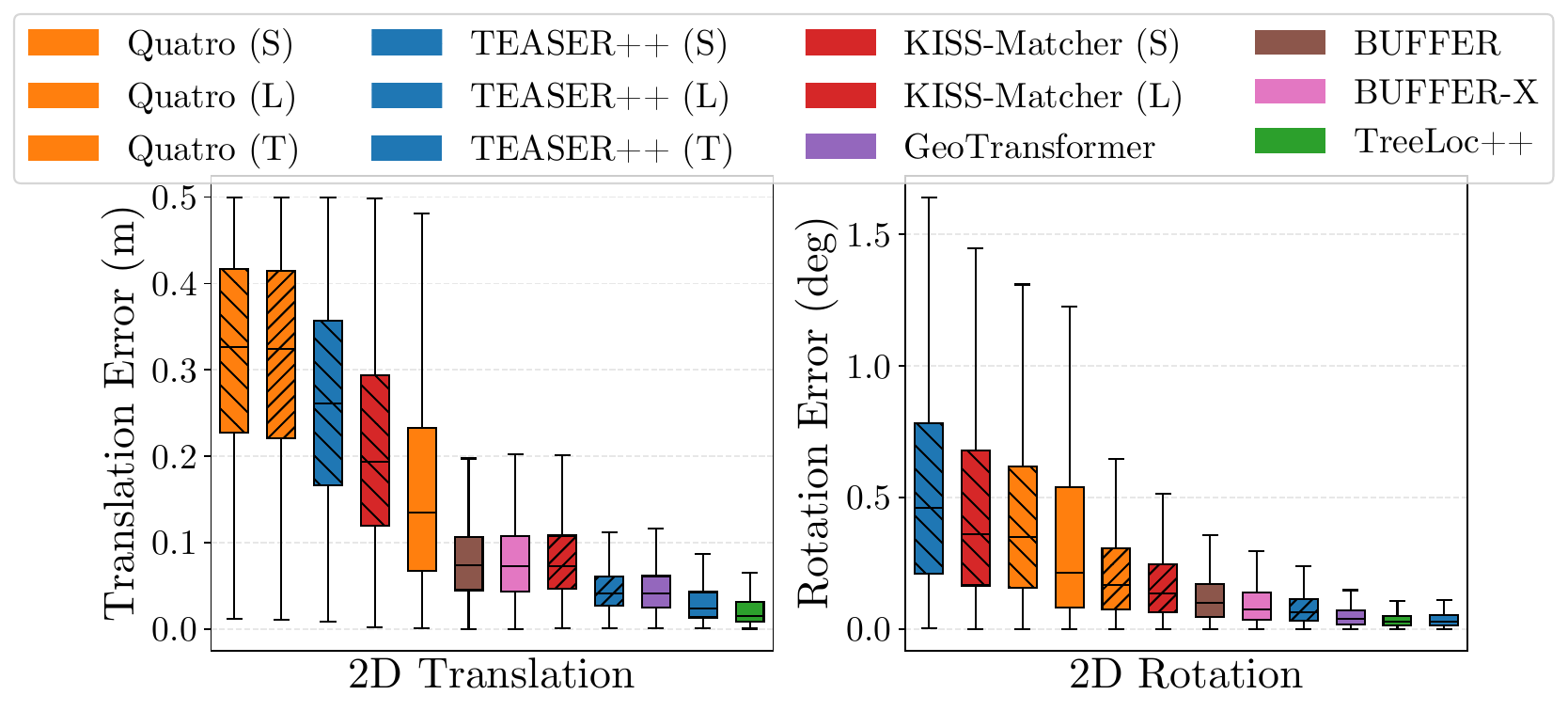}
        \vspace{-6mm}
        \caption{2D Pose Estimation Error in \texttt{Venman03}}
        \vspace{-2mm}
        \label{fig:2d_venman}
    \end{subfigure}
    \begin{subfigure}[t]{\columnwidth}
        \centering
        \includegraphics[width=\columnwidth]{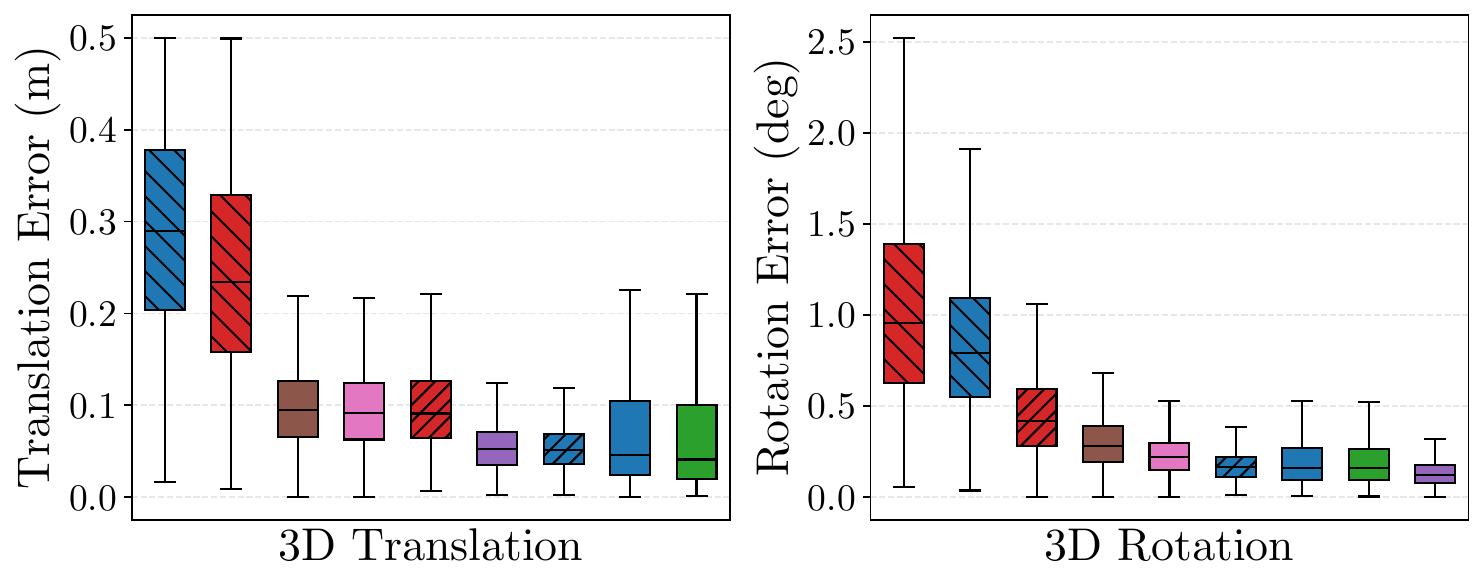}
        \vspace{-6mm}
        \caption{3D Pose Estimation Error in \texttt{Venman03}}
        \vspace{-2mm}
        \label{fig:3d_venman}
    \end{subfigure}
    \caption{Box plots of pose errors on the \texttt{Venman03} for the {1--4} {m} interval. Boxes indicate the IQR, with the median marked inside. TreeLoc++ consistently exhibits low median error and tight error distribution.
}
    \vspace{-3mm}
\end{figure}

\begin{figure}[t]
    \centering
    \includegraphics[width=\columnwidth]{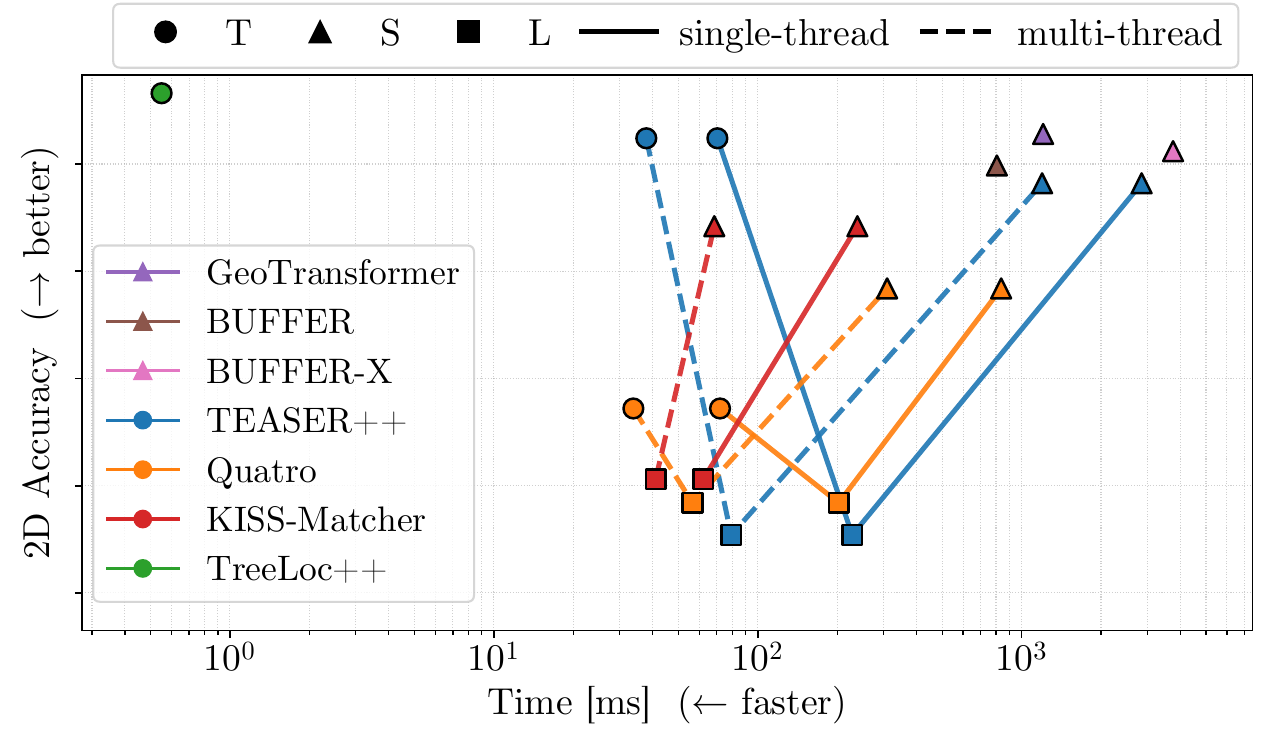}       
    \caption{2D pose estimation error with time consumption on the \texttt{Evo:Single} for the {1--4} {m} translation interval}
    \label{fig:error_time}
    \vspace{-3mm}
\end{figure}

\begin{figure*}[!t]
    \centering
    \begin{subfigure}[t]{.28\textwidth}
        \centering
        \includegraphics[width=\linewidth, trim=0 -70 0 0, clip]{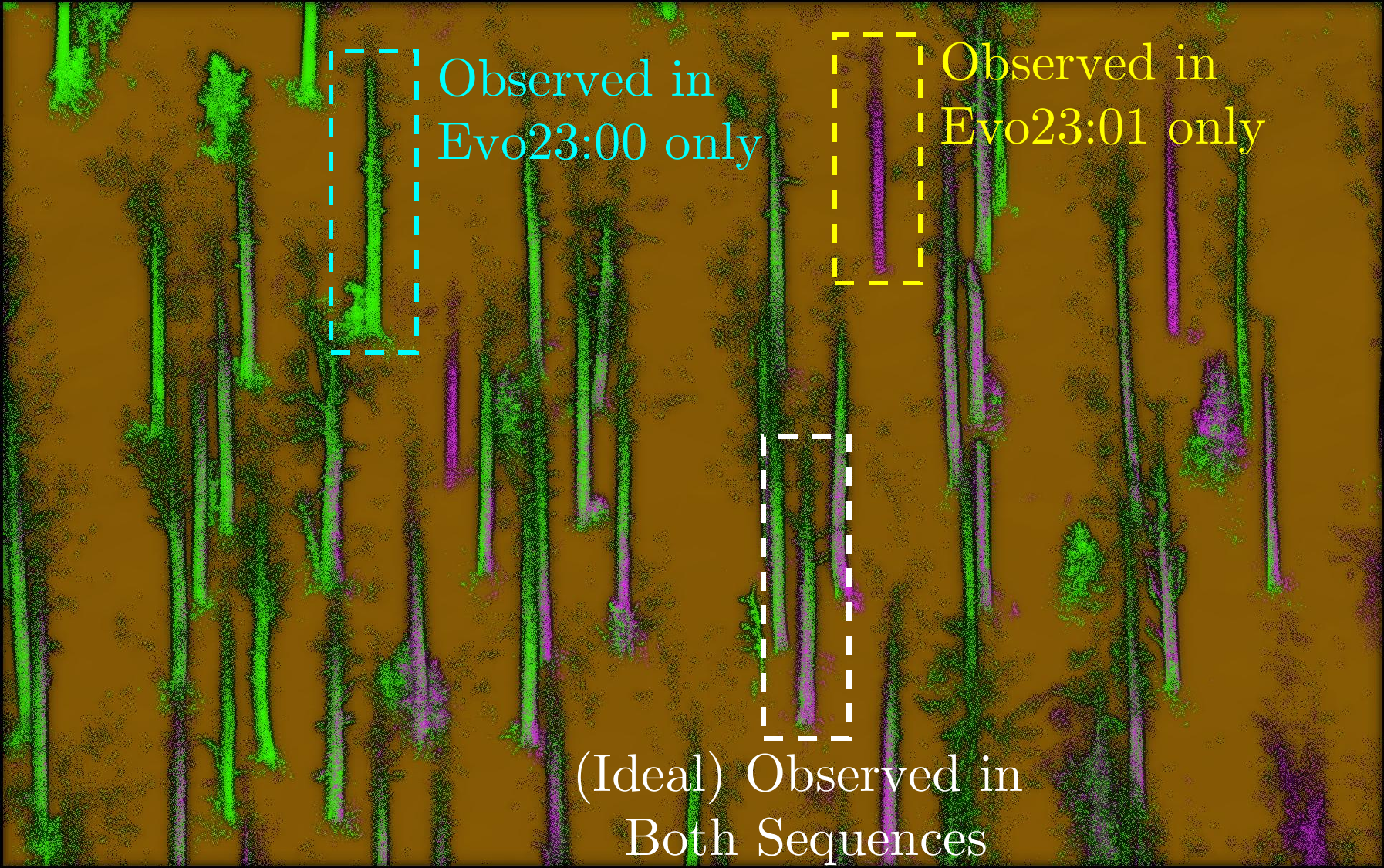}
        \caption{Session-wise Tree Reconstructions}
        \label{fig:trunk_visualization}
    \end{subfigure}
    \begin{subfigure}[t]{.32\textwidth}
        \centering
        \includegraphics[width=\linewidth,trim=0 0 -30 0, clip]{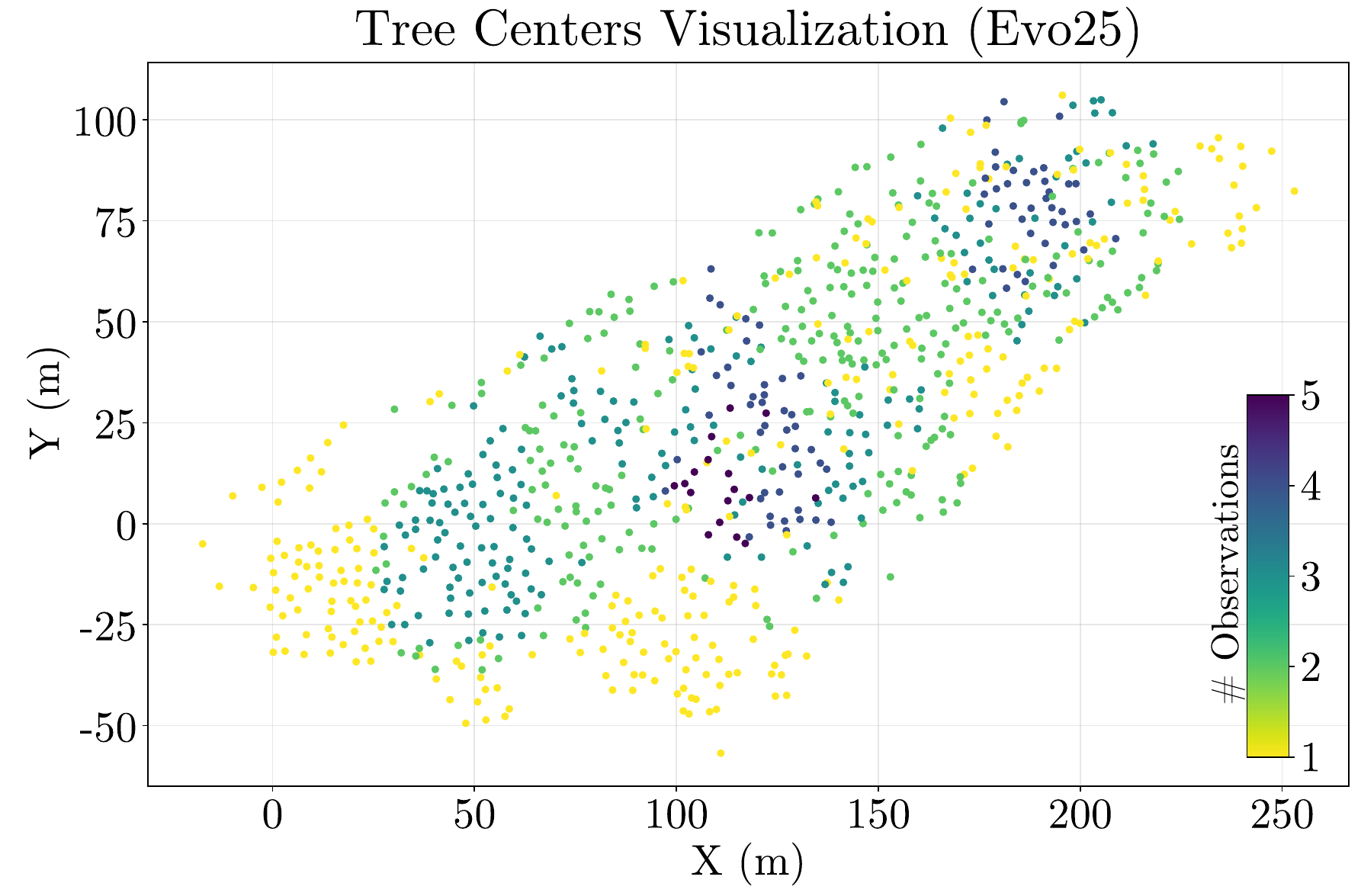}
        \caption{Tree Distribution and Observations in \texttt{Evo25}}
        \label{fig:center_distribution}
    \end{subfigure}
    \begin{subfigure}[t]{.38\textwidth}
        \centering
        \includegraphics[width=\linewidth]{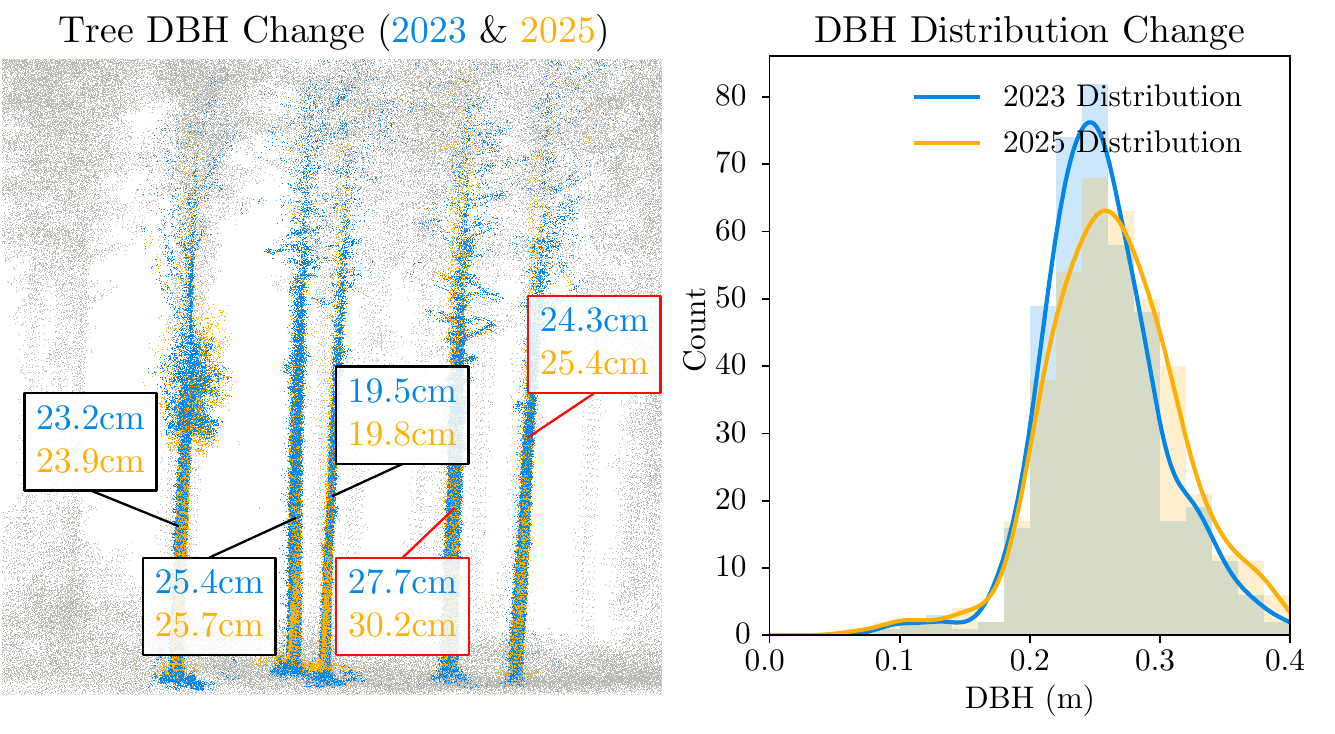}
        \caption{DBH Changes between \texttt{Evo23} and \texttt{Evo25}}
        \label{fig:dbh_summary}
    \end{subfigure}
    \vspace{-2mm}
\caption{(a) Trees reconstructed by RealtimeTrees in \texttt{Evo23:00-01}. Different trajectories lead to session-specific observations, motivating inventory management for partially covered areas. (b) Tree-center distributions in \texttt{Evo25} after TreeLoc++ alignment; brighter colors indicate lower observation rates, highlighting regions requiring additional exploration. (c) Tree point clouds and corresponding DBH measurements from 2023 and 2025 (left). Continuous ecological monitoring is enabled by tracking tree correspondences over time.
The DBH distribution for trees in the 20–80\% growth percentile range (right) shows a slight shift to the right.}

    \vspace{-3mm}
\end{figure*}

\subsection{Pose Estimation Benchmark: Registration Methods}
\noindent \textbf{Evaluation protocol:}
We evaluated TreeLoc++ as a registration method against learning-based baselines (GeoTransformer~\cite{qin2023geotransformer}, BUFFER~\cite{ao2023buffer}, BUFFER-X~\cite{seo2025buffer}) and hand-crafted baselines (Quatro~\cite{lim2022single}, TEASER++~\cite{yang2020teaser}, KISS-Matcher~\cite{lim2025kiss}).
To evaluate performance across spatial separations, we grouped query-candidate pairs into translation ranges of \unit{1-4}{m}, \unit{4-7}{m}, and \unit{7-10}{m}.
For each range, we selected two candidates per query from TreeLoc++ retrieval outputs (one above and one below the midpoint) based on overlap ratio, without guaranteeing true positives.
To avoid bias toward nearby candidates, we disabled the translation penalty \(p(\|\mathbf{t}\|)\) during candidate selection.
Classical baselines were evaluated with two voxel resolutions, small (S) and large (L), as well as a tree-only setting (T) using stem centers and base heights.
Voxel sizes were \unit{0.5/1.0}{m} for \texttt{Evo:Single} and \unit{1.0/2.0}{m} for \texttt{Venman03}. 
We also tested whether TreeLoc++ tree-level features (stem centers and base heights) improve existing registration pipelines, and compared runtime to assess efficiency.
Learning-based baselines were evaluated on the \unit{1-4}{m} range using official configurations and pretrained checkpoints.

\noindent \textbf{TreeLoc++:}
{As summarized in \tabref{tab:intra_registration}, TreeLoc++ achieved highly competitive 2D pose accuracy across all distance ranges.}
In the \unit{1-4}{m} range, it outperformed both classical and learning-based baselines in translation and rotation error, as visualized in \figref{fig:2d_venman}.
In 3D, accuracy degraded with distance for all methods; nevertheless, TreeLoc++ maintained centimeter-level errors comparable to dense point cloud-based approaches. \figref{fig:3d_venman} shows increased interquartile range (IQR), but the medians remained low, supporting the effectiveness of TreeLoc++ for metric localization.

\noindent \textbf{Tree-based vs. point-based registration:}
To evaluate the value of tree-level features, we compared TreeLoc++ with TEASER++ (T).
They achieved comparable accuracy on sparse inputs, indicating that our matching strategy offers robustness similar to that of maximum-clique optimization.
In contrast, classical point-based baselines were more sensitive: Quatro performed poorly due to restrictive orientation assumptions, TEASER++ required fine voxel resolutions to achieve high success, and KISS-Matcher offered an efficiency-accuracy trade-off.

These results further highlight a trade-off between sparse and dense representations.
Specifically, dense point cloud-based methods (S) gain an advantage in 3D stability, as evidenced by their stable performance in \texttt{Venman03}, by utilizing ground points to provide comprehensive 6-DoF constraints.
By comparison, tree-based methods can become unstable when tree centers and base points are degraded by limited visibility or long-range occlusion.
In practice, TreeLoc++ mitigates this by using the translation penalty \(p(\|\mathbf{t}\|)\) during retrieval to favor closer candidates, where tree-level features are more distinctive, thereby improving robustness for metric localization.

\noindent \textbf{Runtime:}
TreeLoc++ also provides a clear computational advantage by combining compact tree-level representations with retrieval-informed correspondences.
As shown in \figref{fig:error_time}, it completes full pose estimation in under \unit{1}{ms} on a single CPU thread by reusing correspondences and avoiding expensive clique solving.
In contrast, classical and learning-based baselines typically require over \unit{100}{ms}, and can exceed \unit{1}{s} at small voxel sizes. This makes them impractical for real-time deployment despite their accuracy, whereas TreeLoc++ incurs minimal latency while maintaining high precision.

In summary, TreeLoc++ achieves accurate and efficient 2D localization, while delivering 3D localization performance competitive with point cloud-based registration methods at a substantially lower computational cost.
Its comparable performance to TEASER++ (T) indicates that tree inventory features, when paired with precise matching, are sufficient for robust forest localization.

\input{tab/appendix_threshold}

\subsection{Multi-Session Digital Forest Inventory Maintenance}

\noindent\textbf{Problem setting:}
We demonstrate a practical \ac{DFI} maintenance pipeline by updating tree traits across multiple sessions (Fig.~\ref{fig:pipeline}) and illustrating how the updated inventory can be used in long-term forest monitoring.
In practical forest deployments, robots cannot exhaustively explore all regions and instead revisit forests along a limited set of traversed paths, yielding uneven observation coverage over time.
Consequently, reconstructions can be session-dependent: some trees are successfully reconstructed in one session but missed in another due to viewpoint, occlusion, or partial coverage, as illustrated in \figref{fig:trunk_visualization}.
This motivates multi-session updates, where observations are fused across sessions to recover missing trees and refine tree-level attributes, maintaining a consistent inventory for the regions of interest.

\noindent\textbf{Multi-session fusion:}
Building on the multi-session registration results, we establish tree correspondences across sessions by transforming inventories into a common global frame and associating them via spatial proximity and stem traits. Instead of re-running the reconstruction on aggregated clusters, we consolidate the per-session tree traits by adopting the attributes from the session with the minimum DBH for each unique tree. This approach is designed to mitigate the overestimation of DBH typically caused by alignment errors, such as odometry drift or sensor noise. This unified inventory ensures more robust tree-level attributes for consistent downstream analyses.

\noindent\textbf{Observation-aware map consistency:}
The updated multi-session \ac{DFI} integrates observations across sessions into a consistent stem map, enabled by TreeLoc++'s robust global alignment and tree association. \figref{fig:center_distribution} shows the spatial distribution of stem centers and the number of sessions in which each tree was reconstructed. Trees in frequently revisited areas appear across multiple sessions, whereas those near the map boundaries are reconstructed in fewer sessions. This spatial pattern qualitatively indicates successful consolidation of observations in well-covered regions, while also revealing under-observed areas that may require further exploration.

\noindent\textbf{Temporal analysis using tree-level attributes:} \figref{fig:dbh_summary} illustrates the potential for long-term ecological monitoring by comparing tree-level attributes across sessions. Using TreeLoc++, we identify consistent tree correspondences between \texttt{Evo23} and \texttt{Evo25}, enabling reliable tracking of individual trees over time. However, detecting DBH changes over a two-year interval is challenging, as actual biological growth is limited. Furthermore, LiDAR sensor noise, such as the \unit{3}{cm} accuracy of the Hesai QT64 used in \texttt{Evo25}, introduces measurement variability. Consequently, some trees in \figref{fig:dbh_summary} (left, red boxes) show DBH variations exceeding the typical growth rate of \unit{2-4}{mm} per year. Nevertheless, the DBH distribution for the 20-80\% percentile range (\figref{fig:dbh_summary}, right) exhibits a slight rightward shift consistent with long-term growth. This morphological stability allows TreeLoc++ to leverage trees as reliable long-term landmarks, supporting robust global localization and consistent ecological monitoring across sessions.

\subsection{Ablation Study on Positive Threshold Selection}
\noindent \textbf{Evaluation Protocol:}
We assessed how localization performance varied with different true positive thresholds by evaluating MF1, R@50, and ATE. We considered three thresholds: \unit{3}{m} (as used in Wild-Places), \unit{10}{m} (from the Oxford Forest Place Recognition dataset), and \unit{5}{m} as a middle ground adopted in this manuscript. As in previous experiments, LoGG3D-Net and MinkLoc3Dv2 employed feature-based matching and RANSAC-based registration, without re-ranking.

\noindent \textbf{Performance Trends:}
As shown in \tabref{tab:appendix_localization}, most methods exhibited increased MF1 as the threshold grew, while R@50 and ATE tended to degrade. This behavior was expected, as larger thresholds admitted more distant candidates as true positives, increasing recall and thus MF1. However, this also reduced geometric overlap between matched pairs, leading to less reliable correspondences and increased translation error. This trend was consistent across both 2D and 3D localization.

\noindent \textbf{Stability across Thresholds:}
TreeLoc++ maintained strong performance across all thresholds and was the only method that consistently achieved centimeter-level ATE. In contrast, most baseline methods showed large variations in MF1 and R@50 as the threshold changed, indicating a strong dependence on the threshold and reduced robustness to the definition of true positives. This variation indicated that the similarity scoring and correspondence selection mechanisms did not reliably capture real-world spatial relationships, leading to exaggerated sensitivity to threshold values and reduced stability in localization performance.

\noindent \textbf{Conclusion:}
These findings highlighted that TreeLoc++ delivered consistent and robust localization performance across a wide range of true positive thresholds, in contrast to baseline methods that were highly sensitive to threshold selection. This stability underscored its suitability for real-world deployment, where fixed thresholds may not always be well defined.

\begin{figure}[t]
    \centering
    \includegraphics[width=\columnwidth]{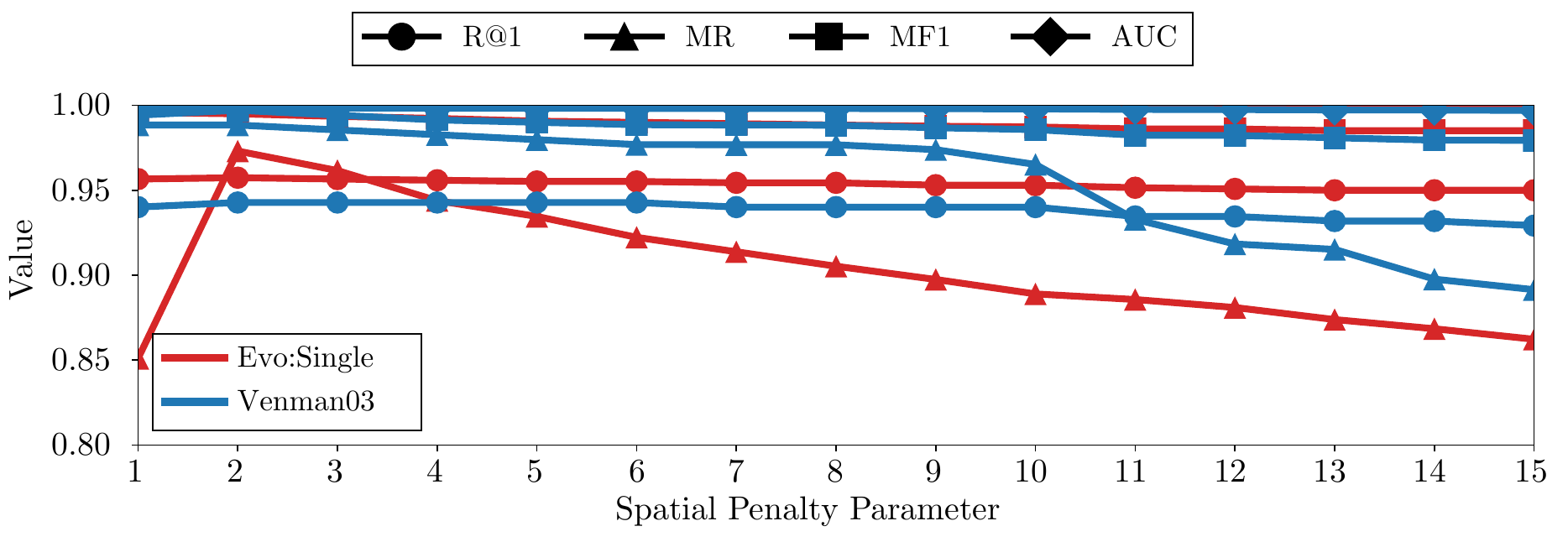}       
    \caption{{Sensitivity of TreeLoc++ to the spatial penalty parameter $\sigma_t$ on \texttt{Evo:Single} and \texttt{Venman03}. The plots report Recall@1 (R@1), Maximum Recall at 100\% precision (MR), Maximum F1 score (MF1), and Area Under the Precision--Recall Curve (AUC). Performance is generally stable across a wide range of $\sigma_t$, showing that the method is not highly sensitive to this parameter except for MR.}}
    \label{fig:tau_sweep}
    \vspace{-3mm}
\end{figure}

\subsection{Ablation Study on Parameter Sensitivity}

\subsubsection{Effect of the Spatial Penalty Parameter}

\noindent \textbf{Evaluation Protocol:}
{We evaluated the sensitivity of TreeLoc++ to the spatial penalty parameter $\sigma_t$, which controls the translation penalty in the overlap score $\mathcal{O}(Q,C)$. Following the same intra-session setup used in \secref{sec:intra_experiment}, we varied $\sigma_t$ from 1 to 15 and reported R@1, MR, MF1, and AUC on \texttt{Evo:Single} and \texttt{Venman03}.}

\noindent \textbf{Performance Trends:}
{As shown in \figref{fig:tau_sweep}, TreeLoc++ remained stable over a broad range of $\sigma_t$. Among the reported metrics, MR was the most sensitive, because a larger $\sigma_t$ relaxed the translation penalty and retained more geometrically distant candidates. Nevertheless, on both datasets the reduction in MR stayed within about 0.1 up to $\sigma_t=15$, while the other metrics varied only slightly, indicating limited practical sensitivity.}

\noindent \textbf{Parameter Choice:}
{In practice, $\sigma_t$ can be selected according to the spatial range over which the estimated pose is intended to remain reliable. In our experiments, we set $\sigma_t=5$, consistent with the positive distance criterion used for place recognition. Although $\sigma_t=5$ was not always the best-performing value, it provided a stable and reliable operating point across both datasets.}

\begin{figure}[!t]
    \centering
    \begin{subfigure}[t]{\columnwidth}
        \centering
        \includegraphics[width=\columnwidth]{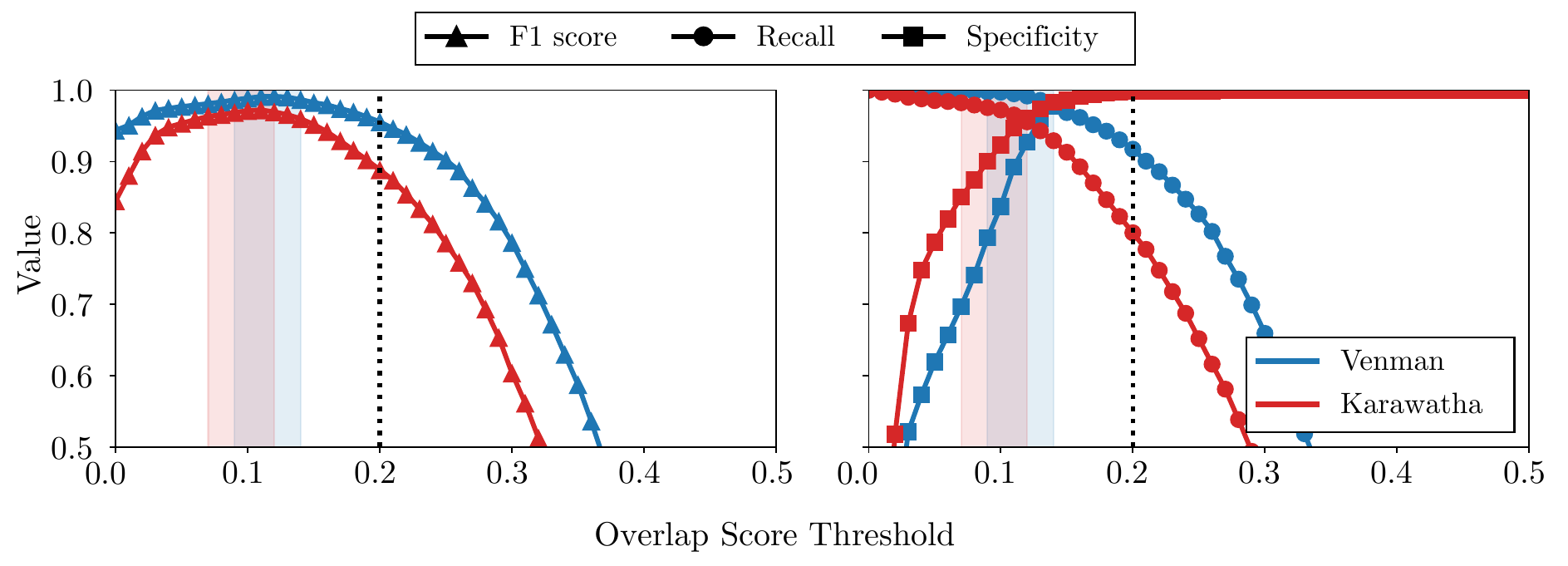}
        \vspace{-6mm}
        \caption{{Inter-session place recognition under varying overlap score thresholds.}}
        \label{fig:overlap_sensitivity1}
    \end{subfigure}
    \begin{subfigure}[t]{\columnwidth}
        \centering
        \includegraphics[width=\columnwidth]{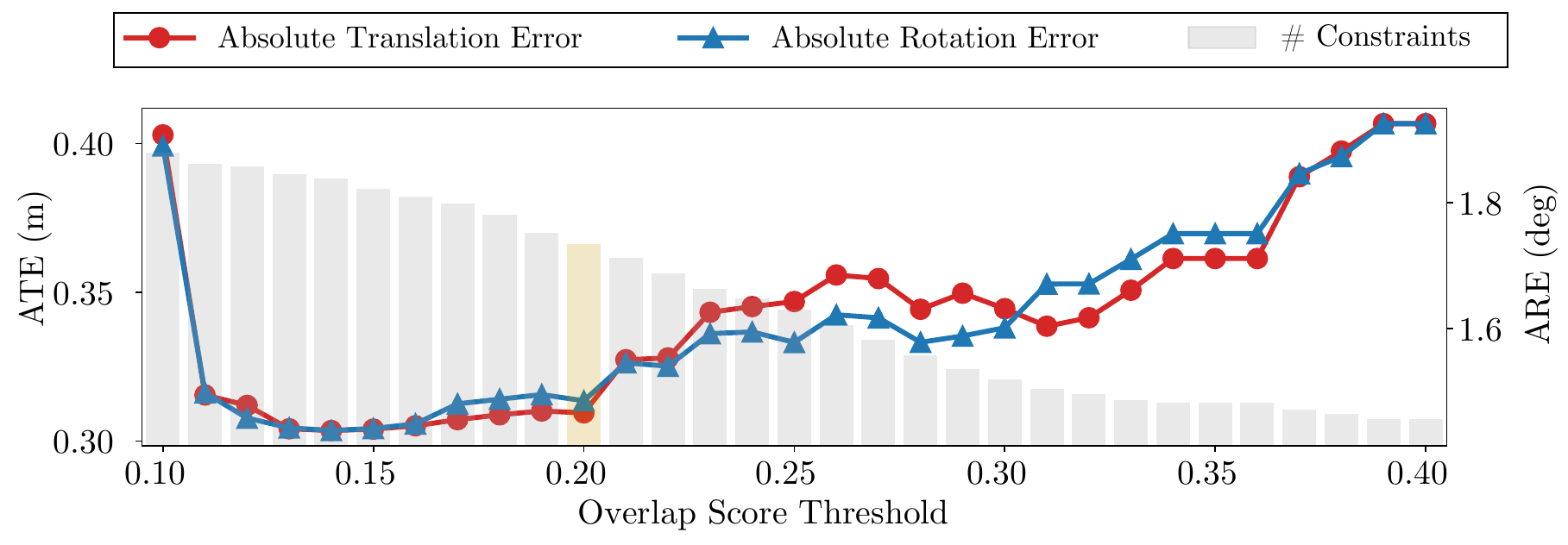}
        \vspace{-6mm}
        \caption{{Multi-session localization under varying overlap score thresholds.}}
        \vspace{-2mm}
        \label{fig:overlap_sensitivity2}
    \end{subfigure}
    \caption{{(a) Inter-session place recognition performance on \texttt{Venman} and \texttt{Karawatha} under varying overlap score thresholds. The shaded regions indicate the min-max ranges of pairwise F1-optimal thresholds, and the dotted line marks the fixed threshold of 0.2, chosen as a conservative operating point favoring high specificity. (b) Effect of the overlap score threshold on multi-session localization. The lines show ATE and ARE, and the bars show the number of accepted constraints. The threshold of 0.2 is highlighted. Over the range of 0.1-0.4, the localization result changed only slightly, with about {10} {cm} variation in ATE and 0.5$^\circ$ variation in ARE, indicating low practical sensitivity.}}
    \vspace{-3mm}
\end{figure}

\subsubsection{Effect of the Overlap Score Threshold}

\noindent \textbf{Evaluation Protocol:}
{We analyzed the sensitivity of TreeLoc++ to the fixed overlap score threshold used for match acceptance. The threshold was first estimated on Wild-Places and then applied uniformly to all datasets. For inter-session place recognition, we evaluated all 24 directed pairs from \texttt{Venman} and \texttt{Karawatha}, and reported the mean F1 score across pairs, together with recall and specificity over all query results, where specificity is defined as $\mathrm{TN}/(\mathrm{TN}+\mathrm{FP})$. For multi-session localization, we additionally reported ATE, ARE, and the number of accepted constraints.}

\noindent \textbf{Threshold Trade-off and Selection:}
{As shown in \figref{fig:overlap_sensitivity1}, the F1-optimal thresholds were generally concentrated around 0.1, consistent with the trend observed in \secref{sec:experiment_inter}. Thus, the fixed threshold of 0.2 was not chosen because it maximized place recognition performance on every dataset, but because it provided a conservative operating point for practical localization, where suppressing false positives is as important as achieving high recall. Correspondingly, recall at 0.2 was lower than near the F1-optimal region, whereas specificity was substantially higher.}

\noindent \textbf{Localization Robustness:}
{This choice was further supported by the multi-session localization results in \figref{fig:overlap_sensitivity2}. We used \texttt{Evo25:00} for this evaluation under the same setup as in \secref{sec:relocalization}. Although the F1-optimal region was generally near 0.1, it could produce relatively larger trajectory errors. By contrast, the selected threshold of 0.2 yielded consistently low ATE and ARE, even if it was not the absolute optimum for either metric. Moreover, sensitivity remained limited in practice: over the range of 0.1-0.4, the variation was only about \unit{10}{cm} in ATE and 0.5$^\circ$ in ARE. These results indicate that the overlap threshold was not highly sensitive and that $\mathcal{O}>0.2$ was a stable and practical choice.}

\begin{figure}[t]
    \centering
    \includegraphics[width=\columnwidth]{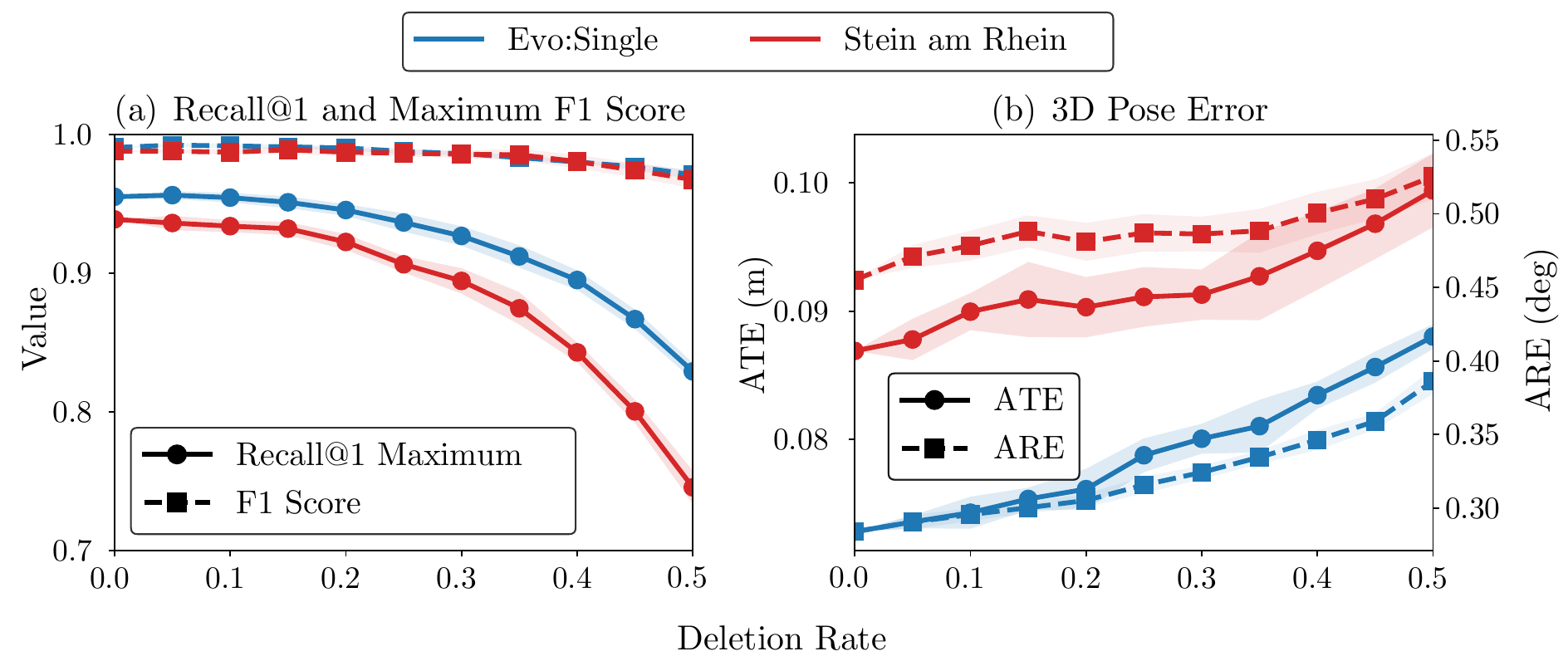}       
    \caption{{Sensitivity to missing tree detections in intra-session localization. The plots show the mean and standard deviation of Recall@1 (R@1), Maximum F1 score (MF1), and 3D pose errors under random deletion of input trees, averaged over 10 trials, on \texttt{Evo:Single} and \texttt{Stein am Rhein}. Although Recall@1 decreased as more trees were removed, MF1 and 3D pose errors remained comparatively stable, indicating that TreeLoc++ was relatively robust to missing tree detections.}}
    \label{fig:robust_delete}
    \vspace{-3mm}
\end{figure}

\subsubsection{Robustness to Missing Detections and DFI Noise}

\noindent \textbf{Evaluation Protocol:}
{To better separate localization performance from upstream DFI quality, we conducted two controlled robustness tests. For intra-session localization, we randomly deleted 0-50\% of the input trees in the DFI used for descriptor generation on \texttt{Evo:Single} and \texttt{Stein am Rhein}, and reported R@1, MF1, ATE, and ARE over 10 trials. For inter-session place recognition, we perturbed only the query-side DFI by adding Gaussian noise to tree centers, DBH, or both, with standard deviations of 1, 3, and \unit{5}{cm}, and evaluated all directed pairs in \texttt{Venman} and \texttt{Karawatha}. We again reported R@1 and MF1 over 10 trials.}

\noindent \textbf{Effect of Missing Tree Detections:}
{As shown in \figref{fig:robust_delete}, increasing tree deletion progressively reduced R@1 on both \texttt{Evo:Single} and \texttt{Stein am Rhein}, whereas the MF1 degraded more slowly. Importantly, the degradation remained moderate, with the reduction in R@1 staying within about 0.1 even when about 40\% of the input trees were removed. The 3D localization errors also increased only gradually, and even at 50\% deletion, ATE rose by only about 1--2\,cm and ARE by about 0.1$^\circ$. These results suggest that missing detections mainly weakened retrieval, while having a smaller effect on candidate discrimination and downstream pose estimation once a valid match was found.}

\begin{figure}[t]
    \centering
    \includegraphics[width=\columnwidth]{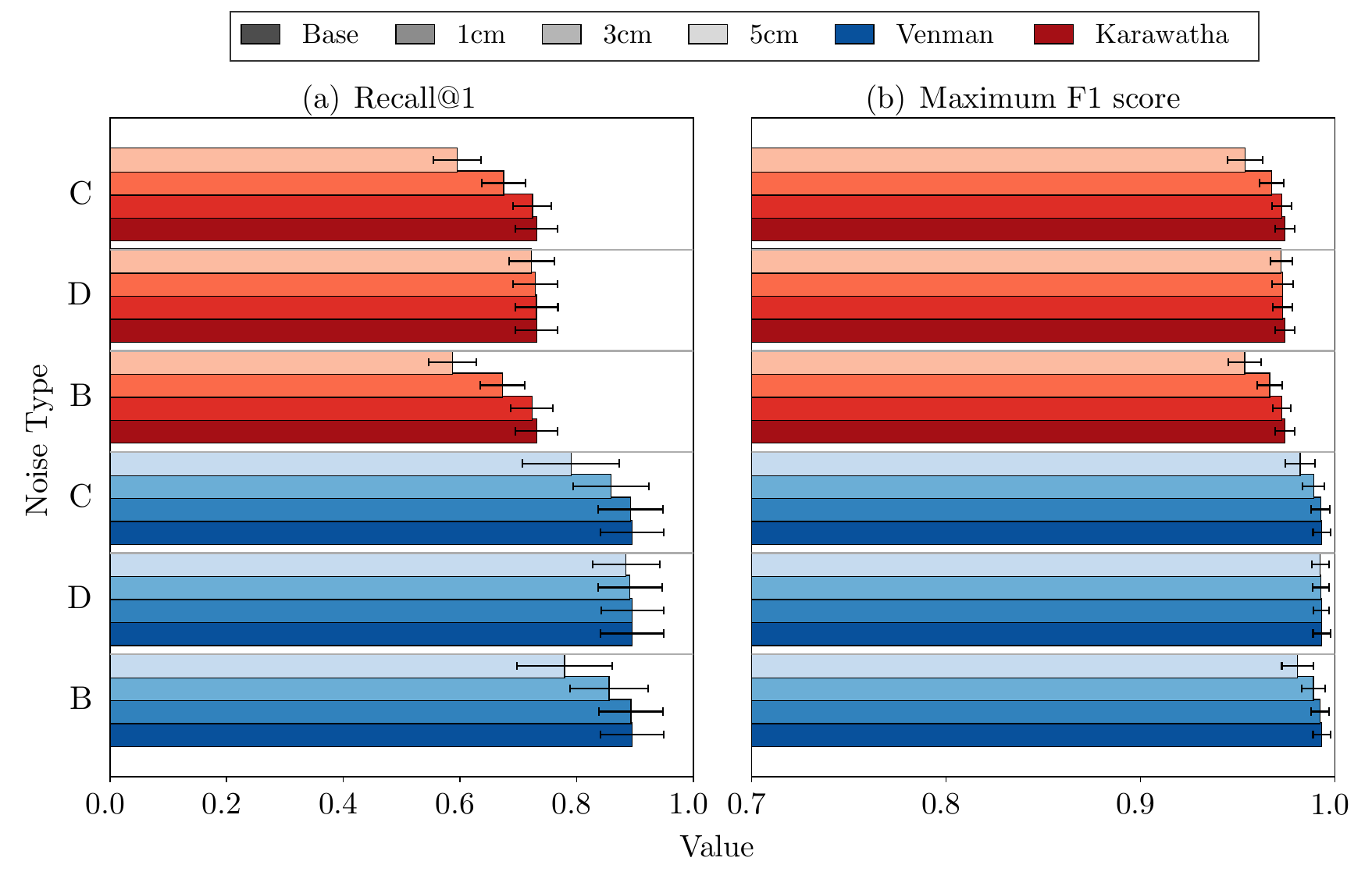}       
    \caption{{Sensitivity to upstream DFI perturbations in inter-session place recognition. The plots show the mean and standard deviation of Recall@1 (R@1) and Maximum F1 score (MF1) under random query-side perturbations of tree centers (C), DBH (D), and both (B), averaged over 10 trials. While R@1 was more sensitive to center perturbation than to DBH perturbation, MF1 remained relatively stable, indicating that TreeLoc++ could still reject outliers and identify correct matches once informative candidates were retrieved.}}
    \label{fig:robust_noise}
    \vspace{-3mm}
\end{figure}

\noindent \textbf{Effect of Upstream DFI Quality:}
{As shown in \figref{fig:robust_noise}, Recall@1 was substantially more sensitive to tree-center perturbation than to DBH perturbation, whereas the MF1 remained comparatively stable, similar to the trend observed under missing detections. Even under DFI perturbations, the retrieval metrics remained within 0.1-0.2 of the nominal case, indicating limited degradation. This suggests that degraded geometry primarily affected retrieval ranking, while TreeLoc++ could still reject outliers and identify correct correspondences once informative candidates were retrieved. The joint perturbation largely followed the center-noise trend, indicating that accurate tree positions were more critical than accurate DBH estimates for robust inter-session matching.}

\noindent \textbf{Conclusion:}
{These controlled experiments show that TreeLoc++ is more sensitive to missing detections and tree-position errors than to DBH noise. However, the degradation was gradual rather than abrupt, and the method remained comparatively stable in MF1 and localization accuracy under moderate perturbations. Overall, these results suggest that, although upstream DFI quality contributes to performance variation, the gains of TreeLoc++ are not solely attributable to a specific tree reconstruction pipeline such as RealtimeTrees.}

\subsection{TreeLoc vs. TreeLoc++: Statistical Evaluation}

\begin{figure}[!t]
    \centering
    \begin{subfigure}[t]{\columnwidth}
        \centering
        \includegraphics[width=\columnwidth]{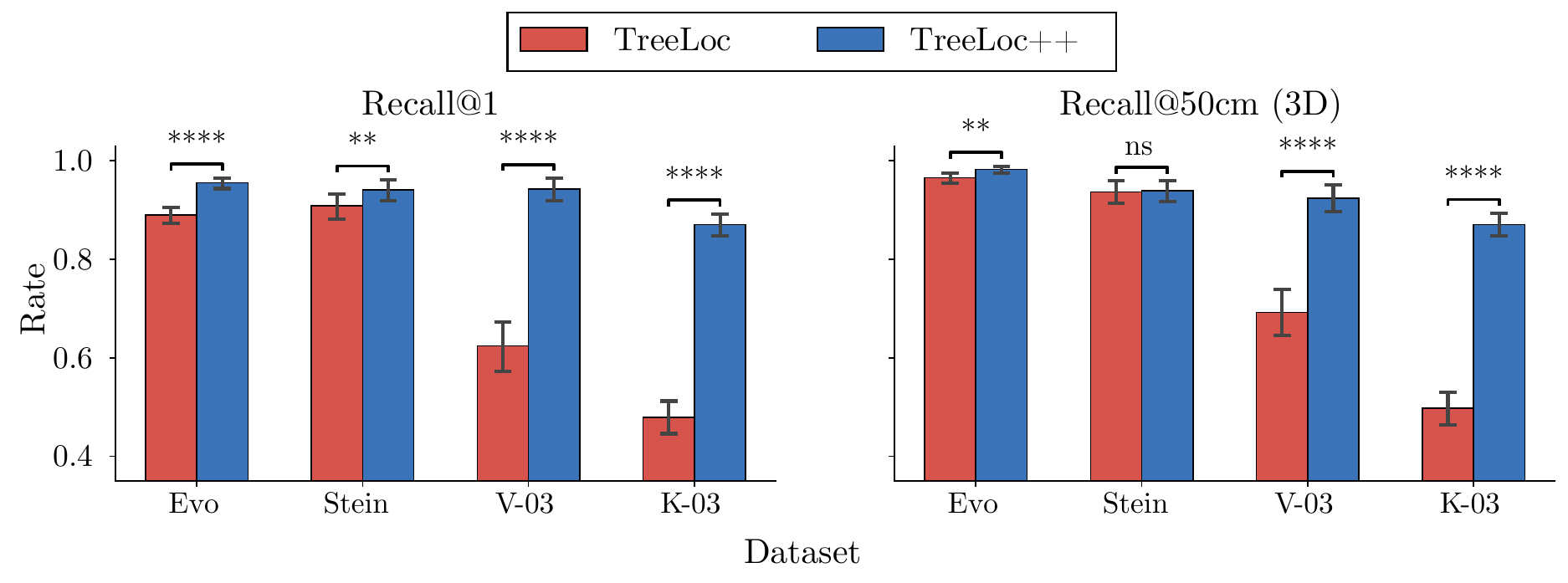}
        \vspace{-6mm}
        \caption{Binary retrieval and localization comparison}
        \label{fig:binary_test}
    \end{subfigure}
    \begin{subfigure}[t]{\columnwidth}
        \centering
        \includegraphics[width=\columnwidth]{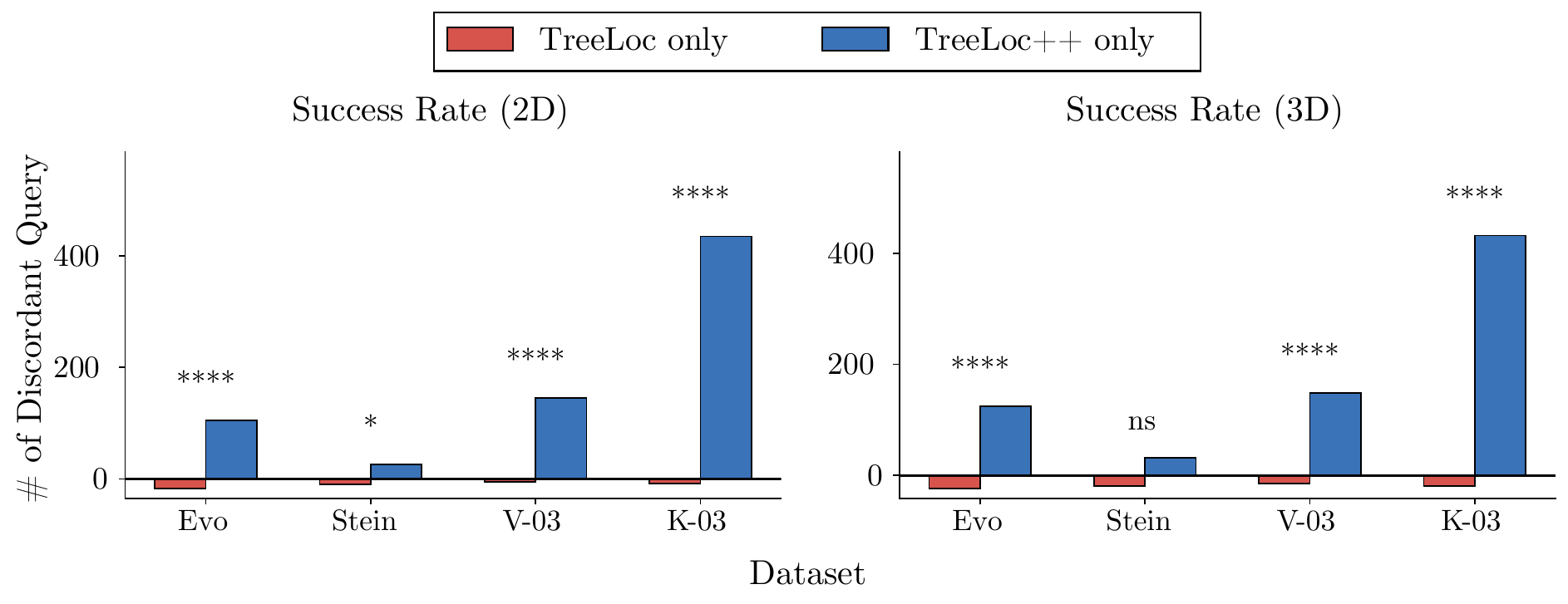}
        \vspace{-6mm}
        \caption{Discordant-query success comparison}
        \vspace{-2mm}
        \label{fig:discordant}
    \end{subfigure}
    \begin{subfigure}[t]{\columnwidth}
        \centering
        \includegraphics[width=\columnwidth]{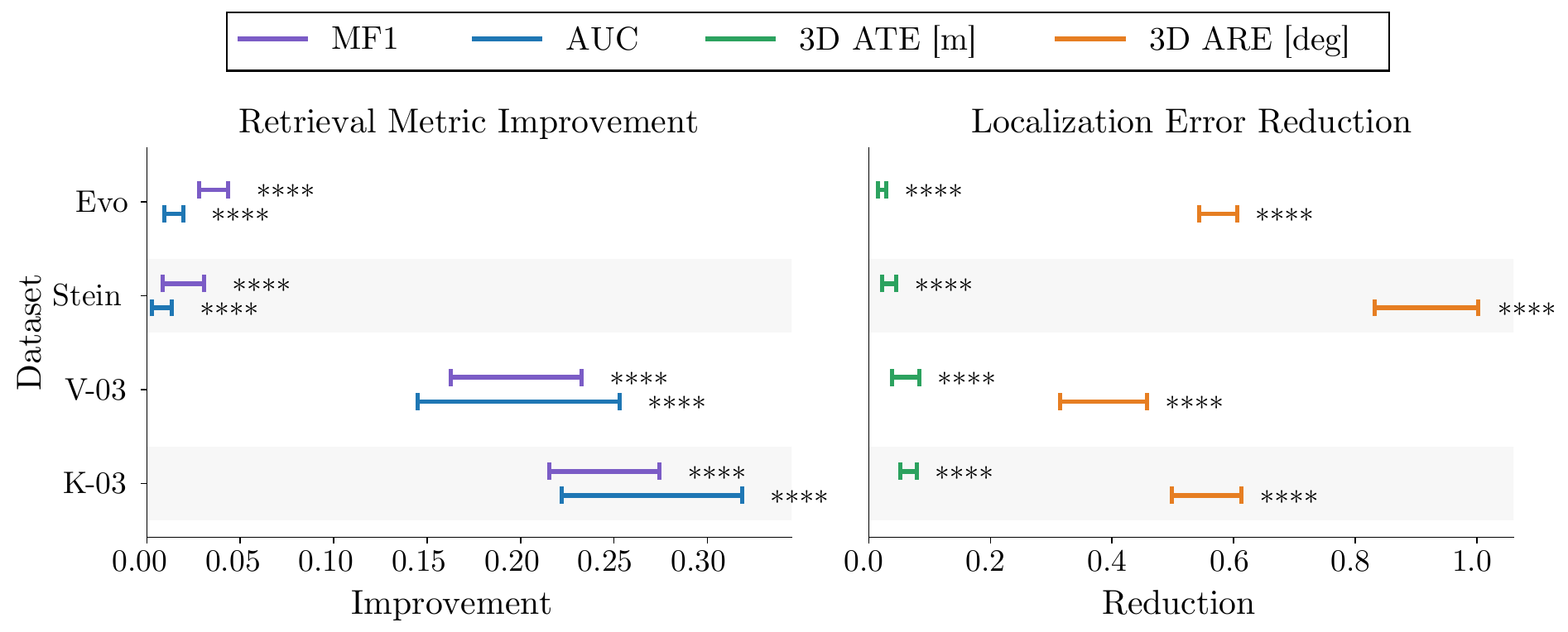}
        \vspace{-6mm}
        \caption{Comparison of retrieval quality and localization accuracy}
        \vspace{-2mm}
        \label{fig:forest}
    \end{subfigure}
    \caption{{TreeLoc vs. TreeLoc++ statistical comparison on four intra-session sequences. Error bars indicate 95\% bootstrap confidence intervals. In (a) and (b), asterisks denote significance levels from exact McNemar tests; in (c), they denote paired bootstrap significance for MF1 and AUC, and Wilcoxon signed-rank significance for ATE and ARE. (a) Binary comparison of Recall@1 and Recall@{50}{cm} (3D). (b) Discordant-query counts for Success Rate (2D and 3D), showing queries localized only by TreeLoc++ or only by TreeLoc. (c) Comparison of retrieval and localization gains, including improvements in MF1 and AUC and reductions in ATE and ARE.}}
    \vspace{-3mm}
\end{figure}

\noindent \textbf{Evaluation Protocol:}
{To assess whether the gains of TreeLoc++ over TreeLoc were statistically meaningful, we compared the two methods on four representative intra-session sequences: \texttt{Evo:Single}, \texttt{Stein am Rhein}, \texttt{Venman03}, and \texttt{Karawatha03}. These sequences span both the smaller Oxford Forest setting and the more challenging Wild-Places setting. We summarized the comparison using three complementary views: binary performance, discordant-query counts, and performance improvements, covering R@1, R@50 in 3D, SR in 2D and 3D, and MF1, AUC, ATE, and ARE. Exact McNemar tests were used for the query-level binary metrics underlying both the binary and discordant comparisons, and Wilcoxon signed-rank tests were used for ATE and ARE on the common-success subset. For all reported metrics, confidence intervals were estimated using paired bootstrap resampling over queries with 2500 samples; for MF1 and AUC, significance was also assessed using the same paired bootstrap procedure.}

\noindent \textbf{Binary and Discordant Comparisons:}
{As shown in \figref{fig:binary_test} and \figref{fig:discordant}, TreeLoc++ consistently outperformed TreeLoc in both binary and discordant-query comparisons. The gains were smaller on the Oxford Forest sequences, especially \texttt{Stein am Rhein}, where TreeLoc already performed strongly and the stricter 3D criteria were statistically similar. By contrast, on the more challenging Wild-Places sequences, \texttt{Venman03} and \texttt{Karawatha03}, TreeLoc++ achieved substantially larger and statistically significant improvements across all binary and discordant comparisons. This indicates that the advantage of TreeLoc++ was not driven by a few isolated cases, but by a systematic increase in the set of queries that could be localized successfully.}

\noindent \textbf{Magnitude of Performance Improvements:}
{As shown in \figref{fig:forest}, the gains of TreeLoc++ extended beyond binary success rates. TreeLoc++ improved both MF1 and AUC on all four sequences, with the largest gains again appearing on \texttt{Venman03} and \texttt{Karawatha03}. It also consistently reduced downstream localization errors. Averaged across the four sequences, TreeLoc++ reduced 3D ATE by about 31.8\% and 3D ARE by about 61.4\%. The corresponding confidence intervals indicate that the improvements were consistent in both retrieval quality and downstream pose accuracy.}

\noindent \textbf{Conclusion:}
{Overall, these statistical comparisons support the conclusion that the improved performance of TreeLoc++ over TreeLoc was systematic rather than incidental. The improvements appeared not only in query-level recall and success rates, but also in retrieval quality and final pose accuracy, with the largest gains observed in the more cluttered and structurally ambiguous forest scenes.}

%% file: tab/appendix_learning.tex

\begin{table}[!t]
\centering
\caption{Comparative analysis of intra-session place recognition in Oxford Forest using urban-trained versus forest-trained models}
\label{tab:intra_appendix_oxford}
\resizebox{\columnwidth}{!}{%
\begin{tabular}{ll|cccc|cccc}
\toprule
\multicolumn{2}{c|}{\multirow{2}{*}{\textbf{Dimension / Method}}} & 
\multicolumn{4}{c|}{\texttt{Stein am Rhein}} & 
\multicolumn{4}{c}{\texttt{Wytham}} \\
\multicolumn{2}{c|}{} & 
R@1 & MR & MF1 & AUC & 
R@1 & MR & MF1 & AUC \\ 
\midrule

\multirow{5}{*}{\rotatebox{90}{\shortstack{Learning-based\\(Urban)}}}
& TransLoc3D   & 0.187 & 0.000 & 0.316 & 0.198 & 0.010 & 0.000 & 0.020 & 0.011 \\
& LoGG3D-Net   & 0.175 & 0.002 & 0.298 & 0.197 & 0.041 & 0.000 & 0.078 & 0.044 \\
& MinkLoc3Dv2  & 0.426 & 0.037 & 0.398 & 0.233 & 0.041 & 0.000 & 0.078 & 0.025 \\
& BEVPlace++   & 0.727 & 0.375 & 0.865 & 0.941 & 0.235 & 0.000 & 0.383 & 0.220 \\
& HOTFormerLoc & 0.330 & 0.136 & 0.598 & 0.703 & 0.327 & 0.063 & 0.535 & 0.414 \\
\midrule

\multirow{5}{*}{\rotatebox{90}{\shortstack{Learning-based\\(Forest)}}}
& TransLoc3D   & 0.395 & 0.004 & 0.566 & 0.507 & 0.082 & 0.000 & 0.151 & 0.091 \\
& LoGG3D-Net   & 0.633 & 0.257 & 0.776 & 0.843 & 0.010 & 0.012 & 0.020 & 0.012 \\
& MinkLoc3Dv2  & 0.538 & 0.086 & 0.577 & 0.429 & 0.031 & 0.010 & 0.020 & 0.006 \\
& BEVPlace++   & \secondbest{0.776} & \secondbest{0.263} & \secondbest{0.900} & \secondbest{0.929} & \secondbest{0.592} & \secondbest{0.172} & \secondbest{0.795} & \secondbest{0.806} \\
& HOTFormerLoc & 0.570 & 0.107 & 0.768 & 0.811 & 0.541 & 0.113 & 0.718 & 0.756 \\
\midrule

\multirow{1}{*}{Ours}

& \cellcolor[HTML]{f3f7fc}\textbf{TreeLoc++}
               & \best{0.941} & \best{0.937} & \best{0.986} & \best{0.998}
               & \best{0.796} & \best{0.962} & \best{0.994} & \best{0.993} \\
\bottomrule
\end{tabular}%
}
\small
\vspace{-0.1mm}
{$\,$}

\scriptsize 
\raggedright 
$\cdot$ Recall@1 (R@1), Maximum Recall at 100\% Precision (MR), Maximum F1 score (MF1), and Area Under the Precision--Recall Curve (AUC).
\vspace{-3mm}
\end{table}

%% file: tab/appendix_learning_wild.tex
\begin{table}[!t]
\centering
\caption{Comparative analysis of intra-session place recognition in Wild-Places using in-domain-trained versus out-of-domain forest-trained models}
\label{tab:intra_appendix_wild}
\resizebox{\columnwidth}{!}{%
\begin{tabular}{ll|cccc|cccc}
\toprule
\multicolumn{2}{c|}{\multirow{2}{*}{\textbf{Dimension / Method}}} & 
\multicolumn{4}{c|}{\texttt{Venman04}} & 
\multicolumn{4}{c}{\texttt{Karawatha04}} \\
\multicolumn{2}{c|}{} & 
R@1 & MR & MF1 & AUC & 
R@1 & MR & MF1 & AUC \\ 
\midrule

\multirow{6}{*}{\rotatebox{90}{\shortstack{Learning-based\\(Oxford Forest)}}}
& TransLoc3D   & 0.469 & 0.000 & 0.638 & 0.496 & 0.520 & 0.000 & 0.684 & 0.618 \\
& LoGG3D-Net   & 0.584 & 0.042 & 0.737 & 0.745 & 0.745 & 0.332 & 0.854 & 0.921 \\
& MinkLoc3Dv2  & 0.647 & 0.029 & 0.785 & 0.774 & 0.593 & 0.134 & 0.744 & 0.776 \\
& BEVPlace++   & 0.694 & 0.015 & 0.819 & 0.666 & 0.860 & 0.088 & 0.929 & 0.951 \\
& ForestLPR      & 0.188 & 0.000 & 0.317 & 0.181 & 0.377 & 0.037 & 0.548 & 0.554 \\
& HOTFormerLoc & 0.613 & 0.013 & 0.765 & 0.717 & 0.587 & 0.031 & 0.749 & 0.742 \\
\midrule

\multirow{6}{*}{\rotatebox{90}{\shortstack{Learning-based\\(Wild-Places)}}}
& TransLoc3D   & 0.801 & 0.003 & 0.890 & 0.868 & 0.790 & 0.003 & 0.883 & 0.928 \\
& LoGG3D-Net   & 0.814 & 0.113 & 0.898 & 0.934 & 0.885 & 0.012 & 0.939 & 0.945 \\
& MinkLoc3Dv2  & 0.853 & 0.037 & 0.921 & 0.913 & 0.867 & 0.360 & 0.927 & 0.947 \\
& BEVPlace++   & 0.885 & \secondbest{0.234} & 0.940 & \secondbest{0.964} 
               & \best{0.945} & \secondbest{0.617} & \secondbest{0.975} & \secondbest{0.987} \\
& ForestLPR    & 0.743 & 0.045 & 0.853 & 0.839 & 0.599 & 0.271 & 0.749 & 0.865 \\
& HOTFormerLoc & \secondbest{0.911} & 0.037 & \secondbest{0.953} & 0.955 
               & \secondbest{0.924} & 0.359 & 0.966 & 0.978 \\
\midrule

\multirow{1}{*}{Ours}
& \cellcolor[HTML]{f3f7fc}\textbf{TreeLoc++}
               & \best{0.924} & \best{0.980} & \best{0.993} & \best{0.998}
               & 0.903 & \best{0.940} & \best{0.995} & \best{0.998} \\
\bottomrule
\end{tabular}%
}
\small
\vspace{-0.1mm}
{$\,$}

\scriptsize 
\raggedright 
$\cdot$ Recall@1 (R@1), Maximum Recall at 100\% Precision (MR), Maximum F1 score (MF1), and Area Under the Precision--Recall Curve (AUC).
\vspace{-3mm}
\end{table}

%% file: tab/intra_pose_estimation.tex
\definecolor{Best}{HTML}{def3e6}
\definecolor{Second}{HTML}{ecf8f1}

\begin{table*}[!t]
\centering
\caption{Pose estimation performance comparison with registration methods.}
\label{tab:intra_registration}
\resizebox{\textwidth}{!}{%
\begin{tabular}{ll|ccc|ccc|ccc|ccc|ccc|ccc}
\toprule
\multicolumn{2}{c|}{\multirow{3}{*}{\textbf{Dimension / Method}}} & 
\multicolumn{9}{c|}{\texttt{Evo:Single}} & 
\multicolumn{9}{c}{\texttt{Venman03}} \\ 
\multicolumn{2}{c|}{} & 
\multicolumn{3}{c}{1--\unit{4}{m} $(N=4670)$} & 
\multicolumn{3}{c}{4--\unit{7}{m} $(N=4668)$} & 
\multicolumn{3}{c|}{7--\unit{10}{m} $(N=4669)$} & 
\multicolumn{3}{c}{1--\unit{4}{m} $(N=4189)$} & 
\multicolumn{3}{c}{4--\unit{7}{m} $(N=4481)$} & 
\multicolumn{3}{c}{7--\unit{10}{m} $(N=4097)$} \\ \cline{3-20}
\multicolumn{2}{c|}{} & 
SR & ATE & ARE & 
SR & ATE & ARE & 
SR & ATE & ARE & 
SR & ATE & ARE & 
SR & ATE & ARE & 
SR & ATE & ARE \\ 
\midrule

\multirow{9}{*}{\rotatebox[origin=c]{90}{\shortstack{2D}}}

& Quatro (S)       & 0.948 & 0.108 & 0.269 & 0.856 & 0.161 & 0.446 & 0.794 & 0.198 & 0.623 & 0.547 & 0.311 & 0.239 & 0.551 & 0.309 & 0.259 & 0.458 & 0.291 & 0.359 \\
& Quatro (L)       & 0.818 & 0.208 & 0.727 & 0.479 & 0.265 & 1.077 & 0.242 & 0.290 & 1.250 & 0.381 & 0.315 & 0.448 & 0.322 & 0.316 & 0.625 & 0.261 & 0.312 & 0.809 \\
& Quatro (T)       & 0.881 & 0.164 & 0.199 & 0.745 & 0.199 & 0.314 & 0.698 & 0.212 & \secondbest{0.365} & \secondbest{0.848} & 0.163 & 0.408 & 0.836 & 0.170 & 0.437 & 0.706 & 0.183 & 0.521 \\
& TEASER++ (S)     & \best{1.000} & 0.059 & \secondbest{0.197} & \secondbest{0.997} & 0.094 & 0.318 & 0.979 & 0.133 & 0.463 & \best{1.000} & 0.046 & 0.081 & \best{1.000} & 0.061 & 0.110 & \best{1.000} & 0.083 & 0.157 \\
& TEASER++ (L)     & 0.848 & 0.223 & 0.844 & 0.506 & 0.278 & 1.263 & 0.247 & 0.296 & 1.306 & 0.815 & 0.262 & 0.562 & 0.620 & 0.290 & 0.694 & 0.433 & 0.299 & 0.862 \\
& TEASER++ (T)     & \best{1.000} & \secondbest{0.038} & \best{0.039} & \best{1.000} & \secondbest{0.047} & \secondbest{0.056} & \best{0.999} & \secondbest{0.052} & \best{0.068} & \best{1.000} & \secondbest{0.036} & \best{0.039} & \secondbest{0.999} & \secondbest{0.055} & \best{0.064} & \secondbest{0.991} & \secondbest{0.074} & \best{0.090} \\
& KISS-Matcher (S) & \secondbest{0.997} & 0.079 & 0.306 & 0.970 & 0.128 & 0.531 & 0.861 & 0.172 & 0.731 & \best{1.000} & 0.082 & 0.179 & \secondbest{0.999} & 0.114 & 0.268 & 0.974 & 0.158 & 0.391 \\
& KISS-Matcher (L) & 0.794 & 0.197 & 0.815 & 0.400 & 0.257 & 1.135 & 0.168 & 0.272 & 1.208 & 0.839 & 0.214 & 0.516 & 0.628 & 0.261 & 0.724 & 0.357 & 0.282 & 0.921 \\
&\cellcolor[HTML]{f3f7fc}\textbf{TreeLoc++}        & \best{1.000} & \best{0.017} & \best{0.039} & \best{1.000} & \best{0.025} & \best{0.054} & \secondbest{0.998} & \best{0.031} & \best{0.068} & \best{1.000} & \best{0.027} & \secondbest{0.042} & 0.991 & \best{0.046} & \secondbest{0.071} & 0.967 & \best{0.067} & \secondbest{0.105} \\
\midrule \midrule

\multirow{6}{*}{\rotatebox[origin=c]{90}{\shortstack{3D}}}
& TEASER++ (S)    & \best{1.000} & \secondbest{0.065} & \best{0.197} 
& \best{0.996} & {0.103} & {0.509} 
& \best{0.975} & {0.146} & 0.740 
& \best{1.000} & \best{0.055} & \best{0.176} 
& \best{1.000} & \best{0.073} & \best{0.241} 
& \best{1.000} & \best{0.098} & \best{0.338} \\

& TEASER++ (L)    & 0.833 & 0.236 & 1.277 & 0.468 & 0.298 & 1.936 & 0.199 & 0.322 & 2.297 & 0.778 & 0.290 & 0.871 & 0.555 & 0.324 & 1.126 & 0.357 & 0.340 & 1.446 \\

& TEASER++ (T)    & 0.990 & 0.074 & 0.263 & \secondbest{0.961} & \secondbest{0.100} & \best{0.417} & \secondbest{0.934} & \secondbest{0.116} & \best{0.526} & \secondbest{0.976} & 0.084 & \secondbest{0.208} & 0.920 & 0.112 & \secondbest{0.352} & 0.845 & \secondbest{0.131} & \secondbest{0.521} \\

& KISS-Matcher (S) & \secondbest{0.995} & 0.090 & 0.718 & 0.958 & 0.148 & 1.092 & 0.815 & 0.198 & 1.405 & \best{1.000} & 0.100 & 0.467 & \secondbest{0.997} & 0.137 & 0.661 & \secondbest{0.962} & 0.188 & 0.879 \\

& KISS-Matcher (L) & 0.749 & 0.225 & 1.721 & 0.315 & 0.290 & 2.237 & 0.114 & 0.310 & 2.389 & 0.788 & 0.246 & 1.082 & 0.530 & 0.295 & 1.413 & 0.267 & 0.320 & 1.706 \\

& \cellcolor[HTML]{f3f7fc}\textbf{TreeLoc++}       & 0.990 & \best{0.061} & \secondbest{0.261} & 0.960 & \best{0.090} & \secondbest{0.418} & 0.930 & \best{0.112} & \secondbest{0.531} & \secondbest{0.976} & \secondbest{0.080} & 0.211 & 0.910 & \secondbest{0.110} & 0.360 & 0.824 & 0.143 & 0.544 \\
\bottomrule
\end{tabular}%
}

\small
\vspace{-0.1mm}
{$\,$}

\scriptsize 
\raggedright 
$\cdot$ S and L use small and large voxel sizes with FPFH-based correspondences, while T uses tree centers and base heights with initial all-to-all correspondences.

$\cdot$ Success Rate (SR), Average Translation Error (ATE) [m], and Average Rotation Error (ARE) [$^{\circ}$].
\vspace{-5mm}
\end{table*}

%% file: tab/appendix_threshold.tex
\begin{table*}[!t]
\centering
\caption{Localization performance under varying true-positive distance thresholds. 
Dataset-level $N$ denotes the total number of queries, while threshold-level $N$ indicates queries with at least one true-positive candidate within the given distance.}

\label{tab:appendix_localization}
\resizebox{\textwidth}{!}{%
\begin{tabular}{ll|ccc|ccc|ccc|ccc|ccc|ccc}
\toprule
\multicolumn{2}{c|}{\multirow{3}{*}{\textbf{Dimension / Method}}} & 
\multicolumn{9}{c|}{\texttt{Evo:Single} (Queries: $N=2334$)} & 
\multicolumn{9}{c}{\texttt{Karawatha04} (Queries: $N=1547$)} \\

\multicolumn{2}{c|}{} & 
\multicolumn{3}{c}{{3 m} ($N_\text{TP}=1138$)} & 
\multicolumn{3}{c}{{5 m} ($N_\text{TP}=1361$)} & 
\multicolumn{3}{c|}{{10 m} ($N_\text{TP}=1715$)} & 
\multicolumn{3}{c}{{3 m} ($N_\text{TP}=328$)}  & 
\multicolumn{3}{c}{{5 m} ($N_\text{TP}=328$)} & 
\multicolumn{3}{c}{{10 m}($N_\text{TP}=342$) } \\ \cline{3-20}

\multicolumn{2}{c|}{} & 
MF1 & R@50 & ATE & 
MF1 & R@50 & ATE & 
MF1 & R@50 & ATE & 
MF1 & R@50 & ATE & 
MF1 & R@50 & ATE & 
MF1 & R@50 & ATE \\ 
\midrule

\multirow{8}{*}{\rotatebox[origin=c]{90}{\shortstack{2D}}}
& LoGG3D-Net   & 0.596 & 0.726 & 0.125 & 0.685 & 0.671 & 0.128 & 0.741 & 0.566 & 0.146 & 0.863 & \secondbest{0.892} & 0.125 & 0.939 & \secondbest{0.900} & 0.131 & 0.940 & \secondbest{0.878} & 0.119 \\
& MinkLoc3Dv2  & 0.537 & 0.040 & 0.324 & 0.640 & 0.035 & 0.310 & 0.702 & 0.026 & 0.297 & 0.813 & 0.087 & 0.327 & 0.928 & 0.088 & 0.283 & 0.979 & 0.073 & 0.305 \\
& BEVPlace++   & 0.889 & 0.990 & 0.122 & 0.933 & 0.979 & 0.134 & 0.939 & 0.920 & 0.151 & \secondbest{0.892} & 0.511 & 0.211 &\secondbest{0.975} & 0.502 & 0.217 & \best{0.997} & 0.499 & 0.225 \\
& RING++       & 0.268 & 0.538 & 0.217 & 0.223 & 0.555 & 0.222 & 0.182 & 0.467 & 0.225 & 0.839 & 0.461 & 0.266 & 0.834 & 0.453 & 0.266 & 0.841 & 0.434 & 0.272 \\
& BTC          & 0.725 & 0.951  & 0.142 & 0.761 & 0.929 & 0.145 & 0.843 & 0.860 & 0.167 & 0.506 & 0.531 & 0.163 & 0.621 & 0.527 & 0.158 & 0.817 & 0.529 & 0.156 \\
& MapClosure   & 0.475 & 0.940 & 0.145 & 0.528 & 0.935 & 0.167 & 0.555 & 0.923 & 0.196 & 0.821 & 0.643 & 0.145 & 0.875 & 0.638 & 0.147 & 0.965 & 0.636 & 0.160  \\
& TreeLoc      & \secondbest{0.904} & \best{1.000} & \best{0.022} & \secondbest{0.955} & \best{0.998} & \best{0.024} & \secondbest{0.981} & \best{0.993} & \best{0.028} & 0.690 & 0.656 & \secondbest{0.049} & 0.816 & 0.654 & \secondbest{0.055} & 0.922 & 0.644 & \secondbest{0.055} \\
& \cellcolor[HTML]{f3f7fc}\textbf{TreeLoc++}
               & \best{0.986} & \secondbest{0.999} & \secondbest{0.025} & \best{0.991} & \secondbest{0.995} & \secondbest{0.026} & \best{0.990} & \secondbest{0.992} & \secondbest{0.031} & \best{0.971} & \best{0.926} & \best{0.037} & \best{0.992} & \best{0.924} & \best{0.038} & \secondbest{0.995}  & \best{0.909} & \best{0.039} \\
\midrule \midrule

\multirow{6}{*}{\rotatebox[origin=c]{90}{\shortstack{3D}}}
& LoGG3D-Net   & 0.596 & 0.704 & 0.144 & 0.685 & 0.645 & 0.151 & 0.741 & 0.539 & 0.171 & \secondbest{0.863} & \secondbest{0.889} & 0.152 & \secondbest{0.939} & \secondbest{0.891} & 0.157 & 0.940 & \secondbest{0.872} & 0.148 \\
& MinkLoc3Dv2  & 0.537 & 0.030  & 0.338 & 0.640 & 0.026 & 0.334 & 0.702 & 0.019 & 0.307 & 0.813 & 0.028 & 0.302 & 0.928 & 0.046 & 0.327 & \secondbest{0.979} & 0.018 & 0.298 \\
& BTC          & 0.725 & 0.684 & 0.177 & 0.761 & 0.632 & 0.185 & 0.843 & 0.546 & 0.206 & 0.506 & 0.408 & 0.227 & 0.621 & 0.403 & 0.227 & 0.817 & 0.400 & 0.237 \\
& MapClosure   & 0.475 & 0.916 & 0.154 & 0.528 & 0.897 & 0.177 & 0.555 & 0.881 & 0.207 & 0.821 & 0.640 & 0.145 & 0.875 & 0.635 & 0.148 & 0.965 & 0.632 & 0.160  \\
& TreeLoc      & \secondbest{0.904} & \secondbest{0.976} & \secondbest{0.089} & \secondbest{0.955} & \secondbest{0.967} & \secondbest{0.095} & \secondbest{0.981} & \secondbest{0.939} & \secondbest{0.107} & 0.690 & 0.594 & \secondbest{0.128} & 0.816 & 0.593 & \secondbest{0.135} & 0.922 & 0.583 & \secondbest{0.144} \\
& \cellcolor[HTML]{f3f7fc}\textbf{TreeLoc++}
               & \best{0.986} & \best{0.992} & \best{0.070} & \best{0.991} & \best{0.982} & \best{0.073} & \best{0.990} & \best{0.955} & \best{0.080} & \best{0.971} & \best{0.919} & \best{0.061} & \best{0.992} & \best{0.924} & \best{0.068} & \best{0.995} & \best{0.909} & \best{0.074} \\
\bottomrule
\end{tabular}%
}
\vspace{-0.1mm}
{$\,$}

\scriptsize 
\raggedright 

$\cdot$ Maximum F1 score (MF1), Recall@\unit{50}{cm} (R@50), Average Translation Error (ATE) [m].
\vspace{-5mm}
\end{table*}